\documentclass{article}

\usepackage[preprint]{neurips_2026}
\usepackage{latexsym}
\usepackage{enumitem}
\usepackage[table]{xcolor}
\usepackage{float}
\usepackage[T1]{fontenc}
\usepackage[utf8]{inputenc}
\usepackage{microtype}
\usepackage{graphicx}
\graphicspath{{figures/}}
\usepackage{booktabs}
\usepackage{array}
\usepackage{multirow}
\usepackage[most]{tcolorbox}
\usepackage{amsmath}
\usepackage{subfigure}
\usepackage{amssymb}
\usepackage{amsfonts}
\usepackage{xcolor}
\usepackage{url}
\usepackage{hyperref}
\usepackage{subcaption}
\usepackage{algorithm}
\usepackage{algorithmic}
\usepackage{tikz}
\usepackage{wrapfig}
\definecolor{ourblue}{RGB}{31,119,180}
\definecolor{ourorange}{RGB}{255,127,14}
\definecolor{ourgreen}{RGB}{44,160,44}
\usepackage{geometry}
\usepackage[utf8]{inputenc} % allow utf-8 input
\usepackage[T1]{fontenc}    % use 8-bit T1 fonts
\usepackage{hyperref}       % hyperlinks
\usepackage{url}            % simple URL typesetting
\usepackage{booktabs}       % professional-quality tables
\usepackage{amsfonts}       % blackboard math symbols
\usepackage{nicefrac}       % compact symbols for 1/2, etc.
\usepackage{microtype}      % microtypography
\usepackage{xcolor}         % colors
\usepackage{pifont}
\title{One Token per Multimodal Evidence: \\Latent Memory for Resource-Constrained QA}

\author{%
  Zhi Zheng\thanks{Email: \texttt{zhi.zheng@u.nus.edu}} \quad
  Ziqiao Meng \quad
  Hao Luan \quad
  Wei Liu \quad
  Wee Sun Lee\\
  School of Computing, National University of Singapore\\
}

\begin{document}

\maketitle
\vspace{-5pt}
% ============================================================
\begin{abstract}
External memory effectively grounds large language models (LLMs) and vision-language models (VLMs)-based question answering (QA) in relevant multimodal evidence. However, existing memory paradigms represent each memory item in raw text and image forms, so retrieval-based systems must pass the retrieved text or images to the generation LLMs/VLMs, resulting in high token consumption and storage pressure, making it unaffordable for resource-constrained applications. We propose \textbf{Latent Memory}, a latent-space memory paradigm that replaces each raw text or image evidence item with \emph{a single high-dimensional latent token} produced by a small compressor LLM/VLM. Rather than retrieving raw evidence for generation, Latent Memory operates in a unified latent representation space: the query is embedded into this space to retrieve relevant latent tokens, and the retrieved latent tokens are directly prompted to a pretrained LLM or VLM for answer generation. To make each latent token simultaneously informative for reconstruction, retrieval, and generation, we train the compressor with reconstruction, contrastive, and distillation objectives in a unified end-to-end manner. Latent Memory is evaluated on seven text-only QA benchmarks (e.g., HotpotQA) and multimodal QA benchmarks, where it achieves competitive QA performance compared to advanced RAG baselines while consuming 3$\times$ to 10$\times$ fewer generator tokens. It can also deliver the strongest image-grounded QA performance on WebQA.\footnote{Code is available at \url{https://github.com/zz1358m/Latent-Memory-Master}}
\end{abstract}\vspace{-5pt}

% ============================================================
\section{Introduction}
% ============================================================

\begin{figure}[t]
\centering\vspace{-6pt}
\subfigure[The conventional RAG-based contextual generation]{\includegraphics[width=\linewidth]{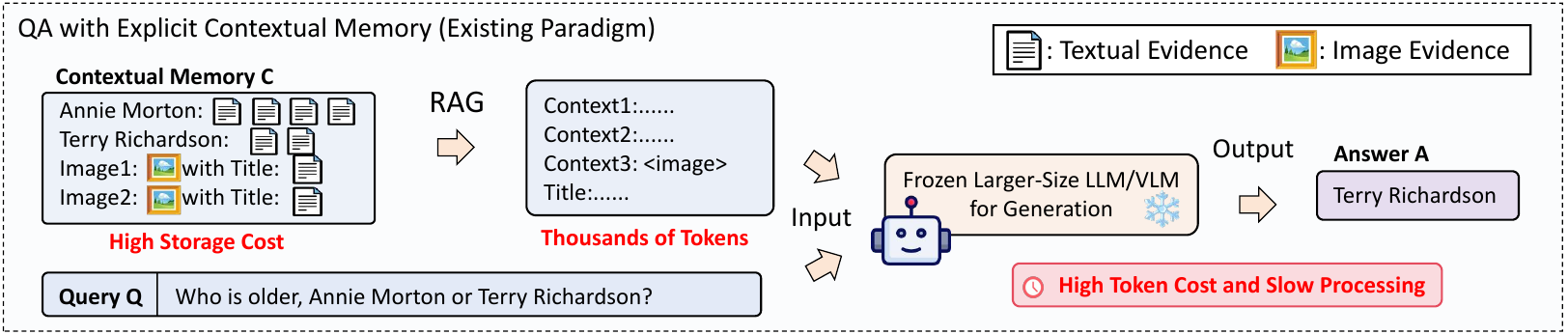}}
\subfigure[Generation with retrievable Latent Memory (Ours)]{\includegraphics[width=\linewidth]{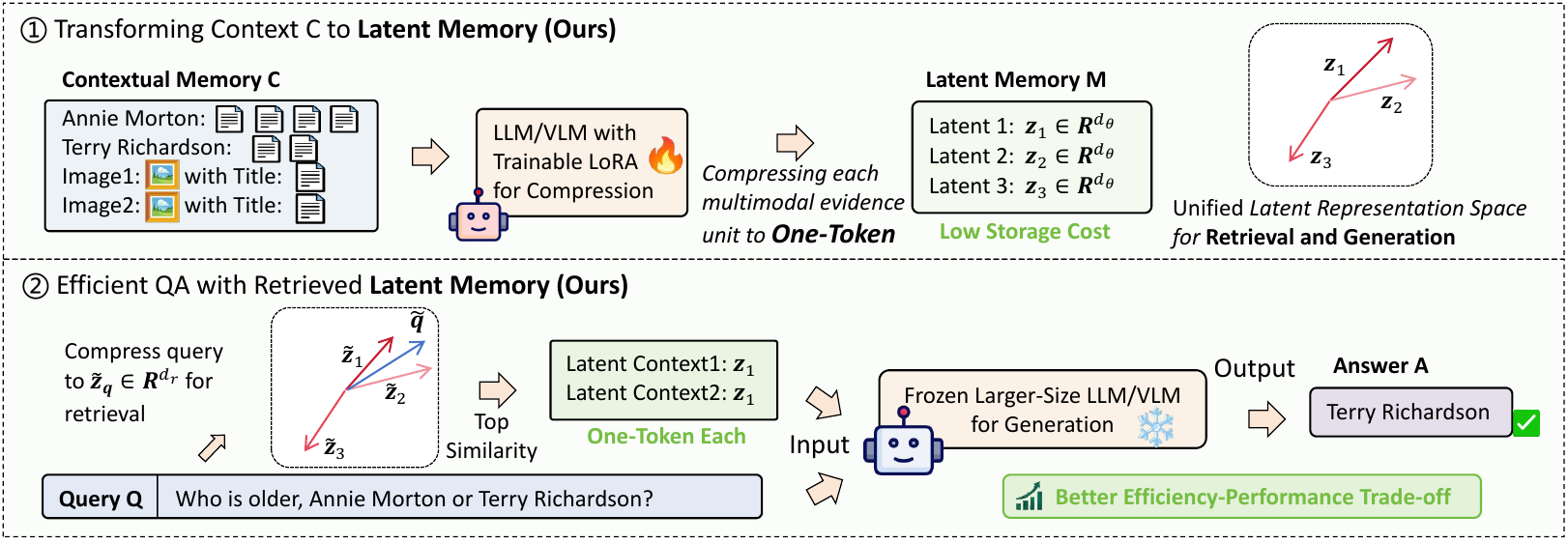}}
\caption{(a) shows the existing pipeline for memory-based generation. To improve storage efficiency and token efficiency, our Latent Memory (b) can compress each multimodal evidence into one latent token, which achieves better retrieval ability and competitive generation performances.}
\label{fig:overview}\vspace{-6pt}
\end{figure}
Large language models (LLMs) and vision-language models (VLMs) have demonstrated remarkable capabilities in complex reasoning and knowledge-intensive tasks \citep{yue2025survey,lin2025llm,dmonte2024claim,zheng2025reasoning}, especially when equipped with an external memory \citep{wu2025human} and then retrieving the relevant memory items \citep{wu2024retrieval} for faithful and reliable generation. This memory usually contains external knowledge evidence or dialogue history in a multimodal form, allowing models to ground their outputs in long-context tasks or multi-turn dialogues, thus becoming a core component of a wide range of applications, including long-context question answering (QA) \citep{daull2023complex,zheng2025reasoning}, coding agents \citep{dong2024survey,zheng2025monte}, and agentic AI assistants \citep{xu2025mem,jimenez2024swe}.

Although external memory can improve answer generation by providing relevant evidence, the prevailing memory paradigm remains computationally expensive. The main bottleneck lies in generation, where the evidence in the memory is prompted to the LLM/VLM generator in uncompressed form, incurring substantial token costs and latency overhead. This problem is further amplified in handling the multimodal evidence in the memory, where each image may require \textbf{megabytes of storage}, and will expand into \textbf{hundreds of visual tokens} during generation \citep{liu2023visual,Kamath2025Gemma3T}. Together, these costs limit the scalability of external memory for long-context and multi-turn interactions, and make it difficult to deploy memory-augmented systems under tight storage or latency constraints, such as on-device assistants and edge-AI applications \citep{park2025mobilerag,mutlu2025memory,yao2024minicpm}.

In pursuit of an efficient memory representation paradigm for applications with strict latency or storage constraints, this paper investigates whether multimodal contextual memory can be represented as highly compressed one-token latent representations capable of replacing raw evidence during generation. To this end, we propose \textbf{Latent Memory}, a framework in which each unit of evidence is compressed into a compact, retrievable, high-dimensional latent token that can be directly utilized by a frozen LLM/VLM generator without fine-tuning. As illustrated in Figure~\ref{fig:overview}(b), Latent Memory uses a small compression LLM/VLM to encode each text or image context into a single latent token, forming a unified multimodal latent representation space. At query time, the question is embedded into the same space to retrieve relevant latent tokens through similarity search. These latent tokens are then projected into the hidden dimension of a frozen LLM or VLM for answer generation, replacing the conventional token-based context. This design enables:
\begin{enumerate}[label=(\arabic*),leftmargin=0.06\textwidth]
    \item \textbf{Token Efficiency and Storage Efficiency}: The proposed Latent Memory leads to fewer token consumptions and lower storage pressure to handle contextual multimodal evidence.
    \item \textbf{Unified Representation and Retrieval Space}: The latent tokens are used for both contextual representation and retrieval, making it easier to obtain better low-budget embeddings.
\end{enumerate}
To obtain better latent tokens for multimodal evidence, we fine-tune a small LLM/VLM for tokens that preserve both the raw information and retrieval capability. So, the training loss combines: (i) a reconstruction loss to ensure latent tokens carry sufficient information of the original evidence; (ii) a contrastive loss to align queries with only their labeled supporting evidence in the latent space; and (iii) a distillation loss to align the generation behavior prompted on latent tokens with that of the behavior generator prompted with raw positive evidence. We evaluate Latent Memory on both text-only and multimodal benchmarks. On 2WikiMultihopQA and MuSiQue, one-token Latent Memory trained on HotpotQA attains competitive EM/F1 compared to raw-evidence-level retrieval augmented generation (RAG) baselines using only \textbf{3$\times$} less LLaMA-8B or Mistral-7B generator tokens. On multimodal QA benchmarks, Latent Memory yields \textbf{10$\times$} efficiency gains and the outstanding image-grounded QA results under the LLaVA-13B and Gemma3-12B generators. Our contributions are as follows:
\begin{itemize}[leftmargin=0.03\textwidth]
    \item We propose \textbf{Latent Memory}, an efficient multimodal memory paradigm for resource-constrained scenarios, which compiles each evidence item into a single high-dimensional latent token.
    \item We show that one-token Latent Memory can support both retrieval and answer generation through a training objective that combines reconstruction, contrastive learning, and distillation.
    \item Empirically, Latent Memory substantially improves the efficiency-performance trade-off across \textbf{eight} text-only and multimodal contextual QA benchmarks, and \textbf{four} generator LLM/VLMs.
\end{itemize}

% ============================================================
\section{Related Work}
% ============================================================

\textbf{External Memory and RAG.}
For faithful QA, LLM/VLM usually takes the provided multimodal context as memory and cooperates with the external memory via full-context prompting \citep{liu2025comprehensive} or RAG \citep{lewis2020retrieval,guu2020retrieval,izacard2023atlas}. Dense Retrieval \citep{karpukhin2020dense}, BM25 Retrieval \citep{robertson2009probabilistic}, and other recent RAG methods \citep{guo2024lightrag,arslan2024survey} encode the feature of queries and evidence into a shared space for the nearest-neighbor search. In multimodal settings, recent multimodal embedding methods \citep{li2026qwen3,jiang2026embedrl,he2026plume} retrieve images based on embedding models and prompt image tokens (576 in LLaVA VLMs \citep{liu2023visual} and 256 in Gemma VLMs \citep{Kamath2025Gemma3T}) to VLMs. However, all of these methods represent memory at the raw-evidence level; as the multimodal memory scales up, this representation becomes a storage and efficiency bottleneck. Latent Memory stores multimodal evidence as compact latent tokens for less storage space, and injects only latent vectors rather than raw evidence into the generation LLM for token efficiency.

\textbf{Evidence Compression.}
Besides the RAG method, there are also Evidence Compression methods proposed to address the over-lengthened large memory. LLMLingua \citep{jiang2023llmlingua} learns to drop unnecessary tokens, AutoCompressor \citep{chevalier2023adapting}, ICAE \citep{ge2023context}, and LCC \citep{li2026latent} learn to compress text documents into a small number of continuous tokens for efficient LLM-based QA generation, while xRAG \citep{cheng2024xrag}, and the concurrent work, CLaRa \citep{he2025clara} take the same latent for retrieval. However, these methods focus only on text-based situations, making them incompatible with recent multimodal memory settings. Moreover, as detailed in Appendix \ref{app:related_context}, considering that high-dimensional tokens are actually larger than the raw text evidence units, these algorithms cannot provide the storage saving as Latent Memory does in multimodal settings (discussed in Appendix \ref{app:space_complexity}).

\textbf{Latent-Space Reasoning.}
Latent reasoning methods \citep{hao2024training,shen2025codi,wei2025sim,zheng2026beyond,he2025clara,li2026latent,fu2026latentmem,zhang2025memgen,yu2025vismem} show that LLMs can reason and communicate through continuous latent states rather than discrete token sequences, while \citep{assran2023self,chen2025vl,nam2026causal,sun2026vla} tries to represent multimodal comprehension and generation in a unified embedding space. Latent Memory applies the idea principle to design a new memory paradigm, where multimodal evidence is encoded, retrieved, and prompted in a unified latent representation.

\begin{figure*}[t]
\centering\vspace{-12pt}
\includegraphics[width=\textwidth]{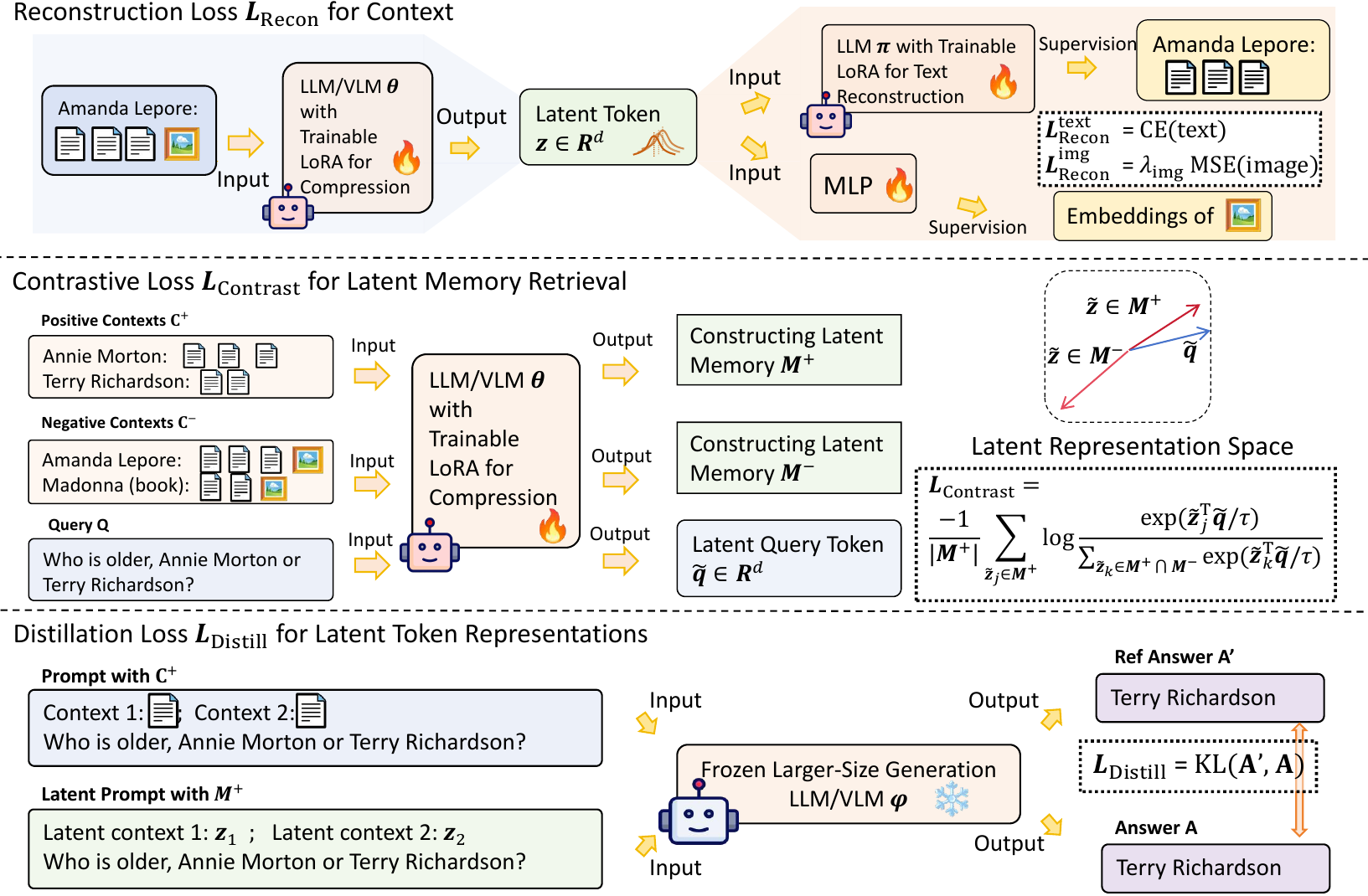}
\caption{The training process of the compressor and decoder consists of three losses. \textbf{Reconstruction Loss $\mathcal{L}_{\mathrm{Recon}}$} aims at recovering the raw text and image in a teacher-forcing way. \textbf{Contrastive Loss $\mathcal{L}_{\mathrm{Constrast}}$} aligns the query embedding to positive latent tokens. \textbf{Distillation Loss $\mathcal{L}_{\mathrm{Distill}}$}  aligns the generator output between prompting raw evidence and latent tokens.}\vspace{-5pt}
\label{fig:training}
\end{figure*}

% ============================================================
\section{Methodology: Latent Memory}
% ============================================================

To achieve efficient external memory, this paper proposes the Latent Memory paradigm. In this section, we first present how the one-token Latent Memory is built and used for QA generation. Then, we will describe how the compression LLM/VLM is trained to produce the Latent Memory.

% ------------------------------------------------------------
\subsection{Definition: Latent Memory for QA with Contextual Memory}
\label{sec:latent_memory}
% ------------------------------------------------------------

A QA problem seeks the answer $\boldsymbol{A}=(a_1,a_2,\ldots,a_{|\boldsymbol{A}|})$ of question $\boldsymbol{Q}=(q_1,q_2,\ldots,q_{|\boldsymbol{Q}|})$ with $N$ external contexts $\mathcal{C}=\{\boldsymbol{x}_i\}_{i=1}^N$. Prompting all the contexts $\mathcal{C}$ for generation will improve perplexity and may exceed the pre-trained context window in some cases. So RAG systems are usually employed to retrieve a subset of raw evidence $\mathcal{R}(q,\mathcal{C})$ and prompt them to a LLM/VLM generator as follows:
\begin{equation}
    P(\boldsymbol{A} \mid  \boldsymbol{Q}, \mathcal{C}, \phi) = \prod_{t=1}^{|\boldsymbol{A}|} P(a_t \mid a_{<t}, \boldsymbol{Q}, \mathcal{R}(\boldsymbol{Q},\mathcal{C}), \phi).
\end{equation}
As shown in Figure~\ref{fig:overview}(b), Latent Memory changes this interface by storing and passing raw evidence to a collection of compact latent tokens. Latent tokens serve as a retrieval representation for selecting relevant evidence, and then the retrieved tokens are directly prompted to the generator as a continuous evidence token. The full inference-time pipeline consists of four components.

\textbf{Memory Compression.}
For each evidence item $\boldsymbol{x}_i \in \mathcal{C}$, we use a small compressor LLM/VLM $\theta$ to produce a single Latent Memory token. Concretely, $\boldsymbol{x}_i$ is appended with a learnable embedding (noted \texttt{[MEM]}), and we take the final hidden state of this token as the latent token:
\begin{equation}
    \boldsymbol{z}_i = \theta(\boldsymbol{x}_i,\texttt{[MEM]}) \in \mathbb{R}^{d_\theta},\label{equation_compressor}
\end{equation}
where $d_\theta$ is the hidden dimension of the model $\theta$. The Latent Memory $\mathcal{M}=\{\boldsymbol{z}_i\}_{i=1}^N$ is then formed by collecting all these latent tokens while discarding the original raw evidence.

\textbf{Retrieving Latent Memory.}
Prompting generators with compressed contexts may be enough for faithful generation within the context window. However, with a large amount of context, the corresponding latent tokens still pose a significant processing difficulty. So, we design to make the Latent Memory token retrievable, projecting $\boldsymbol{z}_i$ into a $d_r$ dimensional retrieval space with an MLP projection, noted $\widetilde{\boldsymbol{z}}_i =\text{MLP}_r( \boldsymbol{z}_i), \ \text{MLP}_r : \mathbb{R}^{d_r}\rightarrow\mathbb{R}^{d_\theta}$. In retrieving the most relevant latent context for the query $q$, we compress the query into a query representation $\widetilde{\boldsymbol{q}}\in\mathbb{R}^{d_r}$ in this retrieval space. We then retrieve the top-$k$ memories by inner product:
\begin{equation}
    \mathcal{I} = \operatorname*{arg\,top\text{-}k}_{i} \; \widetilde{\boldsymbol{q}}^{\top}\widetilde{\boldsymbol{z}}_i.
\end{equation}
\textbf{Latent-Conditioned Generation.}
Retrieved latent tokens $\{\boldsymbol{z}_i : i \in \mathcal{I}\}$ are mapped into the higher-dimensional hidden space of the frozen LLM/VLM $\phi$ for generation. Let $\widehat{\boldsymbol{z}}_j$ denote the projected latent token corresponding to the $j$-th retrieved memory. We construct the input embeddings of the generator with another projector $\boldsymbol{W}_g \in \mathbb{R}^{d_\phi \times d_\theta}$ and prompt them to $\phi$ as follows:
\begin{equation}
    P(\boldsymbol{A} \mid  \boldsymbol{Q}, \mathcal{C}, \phi, \theta) = \prod_{t=1}^{|\boldsymbol{A}|} P(a_t \mid a_{<t}, \boldsymbol{Q}, \widehat{\boldsymbol{z}}_1,\ldots,\widehat{\boldsymbol{z}}_k, \phi) , \quad \text{where}\quad \widehat{\boldsymbol{z}}_i = \boldsymbol{W}_g\boldsymbol{z}_i.
\end{equation}

\textbf{Reconstruction (Optional).}
To preserve some interpretability, Latent Memory can be roughly reconstructed back to the raw multimodal context. Through a fine-tuned decoder $\pi$, we can recover the raw-text for text-evidence or the caption of an image evidence in an autoregressive manner as follows:
\begin{equation}
    P(\boldsymbol{x}_i \mid \boldsymbol{z}_i) = \prod_{t=1}^{|\boldsymbol{x}_i|} P(\boldsymbol{x}_{i,t} \mid \boldsymbol{x}_{i,<t}, \boldsymbol{z}_i,\pi).
\end{equation}
Latent Memory also allows image reconstruction for the latent tokens that are compressed from an image. We train a multi-layer perceptron (MLP) to predict the CLIP embedding of the image \citep{radford2021learning}. Then, the image can be roughly recovered with a pre-trained diffusion-based image generator unCLIP \citep{ramesh2022hierarchical}, conditioning on the recovered CLIP embedding.% Please refer to the Appendix \ref{app:case_recon} for more information on the effects and details of reconstruction.

% ------------------------------------------------------------
\subsection{Training the Compressor for Latent Memory}
\label{sec:training}
% ------------------------------------------------------------

Latent Memory carries retrieval, information-providing, and optional reconstruction roles. In this subsection, we will introduce the algorithm to fine-tune a powerful compressor $\theta$ so that one latent token can support all these roles simultaneously.

As illustrated in Figure~\ref{fig:training}, the training procedure optimizes the compressor with three complementary signals. \textbf{(1)} A reconstruction objective encourages each compressed latent token $\boldsymbol{z}_i$ to preserve the content of its original evidence. \textbf{(2)} A contrastive objective shapes the retrieval space by pulling queries close only to their supporting evidence. \textbf{(3)} A distillation objective aligns the behavior of the frozen generator conditioned on latent memories with the behavior of the same generator conditioned on raw evidence.
To avoid catastrophic forgetting, the large generator $\phi$ is kept frozen throughout training. We only optimize the LoRAs of compressor LLM/ LLM $\theta$, reconstruction decoder LLM $\pi$, and retrieve and generate projections $\boldsymbol{W}_r$ and $\boldsymbol{W}_g$.
The training process only requires supervision from positive samples, and does not need other supervision signals, e.g., labeled answers.

\textbf{Multimodal Reconstruction Loss.} A one-token memory should not collapse into a purely discriminative retrieval identifier. It should still preserve recoverable information about the evidence item it represents. Therefore, for a text evidence item $\boldsymbol{x}_i$ (or the caption of an image), we fine-tune an LLM decoder $\pi$ to reconstruct $\boldsymbol{x}_i$ in its Latent Memory $\boldsymbol{z}_i$ compressed with $\theta$ and a trainable LoRA form Eq. \eqref{equation_compressor} in a teach-forcing way as follows:
\begin{equation}
    \mathcal{L}_{\mathrm{Recon}}^{\mathrm{text}} = - \sum_{i} \sum_{t=1}^{|\boldsymbol{x}_i|}  \log  P_{\pi} \left(        x_{i,t} \mid   \boldsymbol{x}_{i,<t},        \boldsymbol{z}_i    \right).
\end{equation}
For image evidence, we do not reconstruct raw pixels. Instead, following the idea of unCLIP-style reconstruction \citep{ramesh2022hierarchical}, we reconstruct the CLIP image embedding of the original image. Given an image evidence item $\boldsymbol{x}_i^{\mathrm{img}}$ and its latent token $\boldsymbol{z}_i^{\mathrm{img}}$. We train a lightweight MLP to predict a CLIP-space image embedding $\boldsymbol{v}_i$ with the loss function as follows:
\begin{equation}
    \mathcal{L}_{\mathrm{Recon}}^{\mathrm{img}} = \sum_i    \left\|        \boldsymbol{v}_i -  \text{MLP}(\boldsymbol{z}_i) \right\|_2^2 .
\end{equation}
In multimodal training, the reconstruction term combines the available text-side and image-side reconstruction signals as $ \mathcal{L}_{\mathrm{Recon}}  = \mathcal{L}_{\mathrm{Recon}}^{\mathrm{text}}  + \lambda_{\mathrm{img}} \mathcal{L}_{\mathrm{Recon}}^{\mathrm{img}}$.

\textbf{Contrastive Retrieval Loss.} The retrieval projection in Section~\ref{sec:latent_memory} maps each Latent Memory $\boldsymbol{z}_i$ into a retrieval vector $\widetilde{\boldsymbol{z}}_i$. To make this space useful for evidence selection, we train another LoRA of compressor for query representation $\widetilde{\boldsymbol{q}}$ to be close to the latent memories of supporting evidence and far from irrelevant memories. For each question $\boldsymbol{Q}$, let $\mathcal{M}^+$ and $\mathcal{M}^-$ denote its positive and sampled negative latent evidence, respectively. We use a multi-positive contrastive objective, where each positive contributes one InfoNCE \citep{oord2018representation} term:
\begin{equation}
    \mathcal{L}_{\mathrm{Contrast}}
    =
    \frac{1}{|\mathcal{M}_i^+|}
    \sum_{\widetilde{\boldsymbol{z}}_j \in \mathcal{M}^+}
    -
    \log
    \frac{
        \exp
        \left(
            \widetilde{\boldsymbol{q}}^{\top}
            \widetilde{\boldsymbol{z}}_j
            / \tau
        \right)
    }{
        \sum_{\widetilde{\boldsymbol{z}}_k \in \mathcal{M}^+ \cup \mathcal{M}^-}
        \exp
        \left(
            \widetilde{\boldsymbol{q}}^{\top}
            \widetilde{\boldsymbol{z}}_k
            / \tau
        \right)
    },
\end{equation}
where $\tau$ is the temperature. Since both text and image evidence are projected into the same retrieval space, the same loss supports unified retrieval over mixed multimodal memory.

\textbf{Generation distillation loss.}
To ensure latent tokens have similar roles compared to raw evidence, we add a distillation objective. For each training example, the teacher distribution is obtained by autoregressively running the frozen generator $\phi$ with the raw supporting context $\mathcal{C}^+$. This produces a teacher sequence $\boldsymbol{A}^{\mathrm{tea}}=(a^{\mathrm{tea}}_1,\ldots, a^{\mathrm{tea}}_{|\boldsymbol{A}^{\mathrm{tea}}|})$. The student uses the same frozen generator $\phi$, but replaces the raw context with the projected latent memories $\widehat{\boldsymbol{z}}_1,\ldots,\widehat{\boldsymbol{z}}_k$. We then minimize the token-level KL divergence along the teacher-generated trajectory:
\begin{equation}
    \mathcal{L}_{\mathrm{Distill}} =
    \sum_{t=1}^{|\boldsymbol{A}^{\mathrm{tea}}|}
    \mathbb{KL}
    \left(
    P\left(
    \cdot \mid
    \boldsymbol{a}^{\mathrm{tea}}_{<t}, \boldsymbol{Q},
    \mathcal{C}^+), \phi
    \right).
    \,\Vert\,
    P\left(
    \cdot \mid
    \boldsymbol{a}^{\mathrm{tea}}_{<t}, \boldsymbol{Q},
    \widehat{\boldsymbol{z}}_1,\ldots,\widehat{\boldsymbol{z}}_k, \phi
    \right)
    \right).
\end{equation}
This term teaches the compressor and generator projection to produce latent
tokens that the frozen generator can interpret as evidence. In this way,
distillation loss connects the latent retrieval interface to the final QA objective
without fine-tuning the large LLM/VLM generator.

Finally, the overall training objective is as follows, where $\lambda_{\mathrm{Recon}}$, $\lambda_{\mathrm{Contrast}}$, and
$\lambda_{\mathrm{Distill}}$ control the relative weights of reconstruction,
retrieval, and distillation, respectively:
\begin{equation}
    \mathcal{L} =  \lambda_{\mathrm{Recon}} \mathcal{L}_{\mathrm{Recon}}    +    \lambda_{\mathrm{Contrast}} \mathcal{L}_{\mathrm{Contrast}}    +    \lambda_{\mathrm{Distill}} \mathcal{L}_{\mathrm{Distill}},
\end{equation}

\begin{table}[t]
\centering\vspace{-10pt}
\caption{Text-based QA results using \textit{Meta-Llama-3-8B-Instruct} as the generation LLM. All methods use a frozen \textit{Meta-Llama-3-8B-Instruct} generator. The Average columns report the out-of-domain average over 2WikiMultihopQA and MuSiQue. \textbf{Bold} indicates the best metric in each column, and underlining indicates the second-best one. R@$k$ = Recall@$k$.}
\label{tab:text_results}
\renewcommand\arraystretch{1 }
\setlength{\tabcolsep}{1.33mm}
\resizebox{\textwidth}{!}{
\begin{tabular}{lcccc| cccc cccc| cccc}
\toprule[0.5mm]
\multicolumn{17}{c}{\textbf{Generation LLM (fixed): Meta-Llama-3-8B-Instruct}} \\
\midrule
& \multicolumn{4}{c}{\textbf{In-Domain}} & \multicolumn{12}{c}{\textbf{Out-of-Domain}} \\
\cmidrule(lr){2-5}\cmidrule(lr){6-17}
Dataset & \multicolumn{4}{c}{\textbf{HotpotQA}} & \multicolumn{4}{c}{\textbf{2WikiMultihopQA}} & \multicolumn{4}{c}{\textbf{MuSiQue}} & \multicolumn{4}{c}{\textbf{Average}} \\
\cmidrule(lr){2-5}\cmidrule(lr){6-9}\cmidrule(lr){10-13}\cmidrule(lr){14-17}
Method & EM & F1 & R@$k$ & \#Tok & EM & F1 & R@$k$ & \#Tok & EM & F1 & R@$k$ & \#Tok & EM & F1 & R@$k$ & \#Tok \\
\midrule
Full Context
& \textbf{42.0} & \textbf{57.8} & -- & 1462
& 17.7 & \textbf{39.2} & -- & 1074
& 6.0 & 17.1 & -- & 2580
& 11.9 & 28.2 & -- & 1827 \\
\midrule
\multicolumn{17}{l}{\textit{Evidence Compression Baselines}} \\
LLMLingua (20\%)
& 31.8 & 44.8 & -- & 283
& 17.0 & 30.4 & -- & 199
& \underline{11.6} & \underline{21.8} & -- & 492
& 14.3 & 26.1 & -- & 346 \\

LLMLingua (10\%)
& 25.0 & 36.1 & -- & 154
& 14.8 & 24.7 & -- & 108
& 6.9 & 14.9 & -- & 259
& 10.9 & 19.8 & -- & 184 \\

LLMLingua (5\%)
& 20.9 & 30.2 & -- & 87
& 15.0 & 22.2 & -- & 63
& 4.3 & 10.6 & -- & 137
& 9.7 & 16.4 & -- & 100 \\
\midrule
\multicolumn{17}{l}{\textit{RAG Baselines}} \\
BM25 Retrieval ($k{=}1$)
& 32.5 & 44.8 & 30.2 & 68
& 17.2 & 27.2 & 18.3 & 60
& 6.2 & 14.2 & 7.0 & 69
& 11.7 & 20.7 & 12.7 & 65 \\

BM25 Retrieval ($k{=}2$)
& 36.8 & 49.9 & 45.5 & 106
& 16.3 & 28.3 & 29.2 & 94
& 8.0 & 16.7 & 12.2 & 106
& 12.2 & 22.5 & 20.7 & 100 \\

BM25 Retrieval ($k{=}5$)
& \underline{41.3} & \underline{55.3} & 65.5 & 224
& 16.3 & 32.7 & 46.9 & 196
& 9.5 & 19.6 & 22.4 & 221
& 12.9 & 26.2 & 34.7 & 209 \\

Dense Retrieval ($k{=}1$)
& 29.4 & 41.0 & 27.9 & 67
& 19.0 & 30.0 & 23.3 & 61
& 7.8 & 17.1 & 9.6 & 70
& 13.4 & 23.6 & 16.5 & 66 \\

Dense Retrieval ($k{=}2$)
& 32.6 & 45.4 & 42.2 & 104
& 18.1 & 31.7 & 37.6 & 95
& 9.9 & 19.9 & 17.0 & 106
& 14.0 & 25.8 & 27.3 & 101 \\

Dense Retrieval ($k{=}5$)
& 37.0 & 50.8 & 62.4 & 214
& 19.1 & 37.1 & 58.0 & 198
& \underline{12.8} & \underline{23.4} & \underline{31.3} & 218
& 16.0 & \underline{30.3} & 44.7 & 208 \\
Qwen3-Emb-0.6B ($k{=}1$)
& 30.9 & 43.6 & 33.2 & 70
& 18.1 & 30.9 & 32.5 & 66
& 8.1 & 18.2 & 10.3 & 73
& 13.1 & 24.6 & 21.4 & 70 \\

Qwen3-Emb-0.6B ($k{=}2$)
& 35.6 & 49.0 & 50.1 & 109
& 17.5 & 33.5 & 47.7 & 102
& 9.8 & 20.4 & 18.5 & 112
& 13.7 & 27.0 & 33.1 & 107 \\

Qwen3-Emb-0.6B ($k{=}5$)
& 40.0 & 54.5 & \underline{70.1} & 224
& 19.1 & \underline{38.6} & \underline{64.3} & 208
& \textbf{13.7} & \textbf{24.7} & \textbf{34.8} & 230
& \underline{16.4} & \textbf{31.7} & \underline{49.6} & 219 \\
\midrule
\multicolumn{17}{l}{\textit{Ours: Latent Memory}} \\
Latent Memory ($k{=}1$)
& 27.4 & 39.4 & 34.6 & 36
& 19.8 & 29.2 & 28.4 & 33
& 5.8 & 14.6 & 8.7 & 37
& 12.8 & 21.9 & 18.6 & 35 \\

Latent Memory ($k{=}2$)
& 31.6 & 45.2 & 62.6 & 45
& \underline{21.5} & 33.5 & 49.5 & 42
& 7.3 & 16.9 & 15.5 & 46
& 14.4 & 25.2 & 32.5 & 44 \\

Latent Memory ($k{=}5$)
& 34.8 & 48.9 & \textbf{87.1} & 72
& \textbf{24.3} & 36.7 & \textbf{74.2} & 69
& 8.7 & 19.2 & 30.1 & 73
& \textbf{16.5} & 28.0 & \textbf{52.2} & 71 \\
\bottomrule[0.5mm]
\end{tabular}}
\end{table}

\begin{table}[t]
\centering
\caption{Text-based QA results using \textit{Mistral-7B-Instruct} as the generation LLM. The Latent Memory rows use \textit{LLaMA-3.2-1B-Instruct} as both compressor/encoder and reconstruction decoder. xRAG and CLaRa use their pretrained Mistral-based checkpoints. The Average columns report the out-of-domain average over 2WikiMultihopQA and MuSiQue.}
\label{tab:mistral_text_results}
\renewcommand\arraystretch{1 }
\setlength{\tabcolsep}{1.1mm}
\resizebox{\textwidth}{!}{
\begin{tabular}{lcccc| cccc cccc| cccc}
\toprule[0.5mm]
\multicolumn{17}{c}{\textbf{Generation LLM (fixed): Mistral-7B-Instruct}} \\
\midrule
& \multicolumn{4}{c}{\textbf{In-Domain}} & \multicolumn{12}{c}{\textbf{Out-of-Domain}} \\
\cmidrule(lr){2-5}\cmidrule(lr){6-17}
Dataset & \multicolumn{4}{c}{\textbf{HotpotQA}} & \multicolumn{4}{c}{\textbf{2WikiMultihopQA}} & \multicolumn{4}{c}{\textbf{MuSiQue}} & \multicolumn{4}{c}{\textbf{Average}} \\
\cmidrule(lr){2-5}\cmidrule(lr){6-9}\cmidrule(lr){10-13}\cmidrule(lr){14-17}
Method & EM & F1 & R@$k$ & \#Tok & EM & F1 & R@$k$ & \#Tok & EM & F1 & R@$k$ & \#Tok & EM & F1 & R@$k$ & \#Tok \\
\midrule
Full Context
& \textbf{21.0} & \textbf{44.9} & -- & 1701
& \underline{3.8} & \textbf{27.1} & -- & 1244
& \underline{2.8} & \textbf{14.9} & -- & 3012
& \textbf{3.3} & \textbf{21.0} & -- & 2128 \\
\midrule
\multicolumn{17}{l}{\textit{Evidence Compression Baselines}} \\
LLMLingua (20\%)
& 8.9 & 27.6 & -- & 406
& 1.4 & 18.4 & -- & 286
& 1.3 & 10.3 & -- & 708
& 1.3 & 14.4 & -- & 497 \\
LLMLingua (10\%)
& 6.8 & 22.6 & -- & 217
& 0.6 & 15.1 & -- & 149
& 0.9 & 7.4 & -- & 371
& 0.8 & 11.2 & -- & 260 \\
LLMLingua (5\%)
& 4.7 & 17.9 & -- & 117
& 0.4 & 13.0 & -- & 81
& 0.5 & 5.5 & -- & 191
& 0.5 & 9.3 & -- & 136 \\
xRAG
& 12.1 & 25.3 & -- & 42
& 3.5 & 17.6 & -- & 38
& 1.4 & 9.5 & -- & 42
& 2.5 & 13.6 & -- & 40 \\
CLaRa-16x ($k{=}5$)
& 5.2 & 15.3 & 55.8 & 119
& \textbf{5.1} & 16.2 & 26.5 & 115
& 0.7 & 6.0 & 22.3 & 119
& 2.9 & 11.1 & 24.4 & 117 \\
\midrule
\multicolumn{17}{l}{\textit{RAG and Latent-context Baselines}} \\
BM25 Retrieval ($k{=}1$)
& 11.0 & 29.6 & 30.2 & 76 & 0.4 & 14.4 & 18.3 & 66 & 0.9 & 8.1 & 6.9 & 75 & 0.6 & 11.3 & 12.6 & 71 \\
BM25 Retrieval ($k{=}2$)
& 13.8 & 33.8 & 45.5 & 120 & 0.7 & 15.8 & 29.2 & 104 & 1.7 & 9.5 & 12.1 & 119 & 1.2 & 12.6 & 20.7 & 111 \\
BM25 Retrieval ($k{=}5$)
& 15.9 & 37.8 & 65.5 & 255 & 1.0 & 18.6 & 46.9 & 221 & 1.7 & 10.7 & 22.4 & 250 & 1.3 & 14.7 & 34.6 & 236 \\
Dense Retrieval ($k{=}1$)
& 9.0 & 26.4 & 27.9 & 75 & 0.6 & 15.8 & 23.3 & 67 & 1.0 & 10.2 & 9.5 & 77 & 0.8 & 13.0 & 16.4 & 72 \\
Dense Retrieval ($k{=}2$)
& 11.3 & 30.2 & 42.2 & 117 & 1.0 & 17.9 & 37.6 & 107 & 1.2 & 11.1 & 16.9 & 119 & 1.1 & 14.5 & 27.3 & 113 \\
Dense Retrieval ($k{=}5$)
& 13.9 & 34.0 & 62.4 & 244 & 2.7 & 22.5 & 58.0 & 225 & 1.6 & 12.8 & \underline{31.3} & 247 & 2.2 & 17.6 & 44.7 & 236 \\
Qwen3-Emb-0.6B ($k{=}1$)
& 10.3 & 28.4 & 33.2 & 79 & 0.7 & 17.1 & 32.5 & 74 & 1.0 & 10.6 & 10.3 & 81 & 0.9 & 13.8 & 21.4 & 77 \\
Qwen3-Emb-0.6B ($k{=}2$)
& 12.6 & 32.7 & 50.1 & 124 & 1.2 & 19.8 & 47.7 & 115 & 1.5 & 11.2 & 18.5 & 126 & 1.3 & 15.5 & 33.1 & 121 \\
Qwen3-Emb-0.6B ($k{=}5$)
& 15.1 & 36.6 & 70.1 & 256 & 3.0 & 24.0 & 64.3 & 236 & 1.6 & 13.4 & \textbf{34.8} & 262 & 2.3 & 18.7 & 49.6 & 249 \\
Qwen3-Emb-0.6B-ft ($k{=}1$)
& 12.3 & 32.2 & 37.9 & 80 & 0.9 & 17.0 & 35.5 & 74 & 1.3 & 11.4 & 9.8 & 83 & 1.1 & 14.2 & 22.6 & 79 \\
Qwen3-Emb-0.6B-ft ($k{=}2$)
& 16.3 & 38.4 & 59.7 & 127 & 1.7 & 21.5 & 54.4 & 119 & 1.4 & 12.1 & 16.5 & 133 & 1.6 & 16.8 & 35.4 & 126 \\
Qwen3-Emb-0.6B-ft ($k{=}5$)
& \underline{18.6} & \underline{42.5} & \underline{80.5} & 267 & \underline{3.8} & \underline{25.9} & \underline{71.3} & 247 & 2.0 & 14.3 & 30.8 & 280 & 2.9 & \underline{20.1} & \underline{51.0} & 264 \\
\midrule
\multicolumn{17}{l}{\textit{Ours: Latent Memory}} \\
Latent Memory ($k{=}1$)
& 12.3 & 30.4 & 34.8 & 36 & 1.8 & 20.6 & 27.6 & 32 & 2.3 & 11.8 & 8.4 & 36 & 2.1 & 16.2 & 18.0 & 34 \\
Latent Memory ($k{=}2$)
& 14.8 & 34.8 & 62.2 & 46 & 2.0 & 22.8 & 49.4 & 42 & 2.2 & 12.8 & 14.9 & 46 & 2.1 & 17.8 & 32.2 & 44 \\
Latent Memory ($k{=}5$)
& 17.5 & 37.8 & \textbf{86.6} & 76 & 2.9 & 24.1 & \textbf{75.9} & 72 & \textbf{3.6} & \underline{14.7} & 28.8 & 76 & \underline{3.2} & 19.4 & \textbf{52.3} & 74 \\
\bottomrule[0.5mm]
\end{tabular}}
\end{table}

\vspace{-2pt}
% ============================================================
\section{Experiments}
% ============================================================
\vspace{-2pt}

We evaluate Latent Memory in two settings: \textbf{(1)} text-only contextual QA and \textbf{(2)} multimodal contextual QA. In the main text-only setting, we use a frozen \textit{Meta-Llama-3-8B-Instruct} \citep{touvron2023llama} generator $\phi$ and fine-tune \textit{LLaMA-3.2-1B-Instruct} \citep{touvron2023llama} with LoRA adapter as the compressor $\theta$. To compare against text-only latent-context baselines that are built around Mistral, we additionally report a frozen \textit{Mistral-7B-Instruct} \citep{jiang2023mistral} generator setting, where both the Latent Memory compressor/encoder and the reconstruction decoder are \textit{LLaMA-3.2-1B-Instruct}. In the multimodal setting, we use a frozen \textit{LLaVA-1.5-13B} \citep{liu2023visual} generator and fine-tune \textit{LLaVA-1.5-7B} \citep{liu2023visual} with LoRA adapters as the compressor. We also include another setting with \textit{Gemma-3-12B-Instruct} \citep{Kamath2025Gemma3T} generator and \textit{Gemma-3-4B} \citep{Kamath2025Gemma3T} compressor for multimodal QA in Appendix~\ref{app:gemma_mm_results}. For all reported settings, the reconstruction decoder $\pi$ is fine-tuned from \textit{LLaMA-3.2-1B-Instruct}. Experiments are done on a Nvidia H200 141GB GPU. Full training and evaluation details are deferred to Appendix~\ref{app:implementation_details}.

\textbf{Dataset.}
In the text-only setting, the Latent Memory is trained on \textit{HotpotQA} training dataset and evaluated on the validation set of in-domain \textit{HotpotQA} and out-of-domain \textit{2WikiMultihopQA}, \textit{MuSiQue}. To investigate a larger transfer, Appendix~\ref{app:text_super_ood} adds a \emph{generalization-on-more-domains} suite that spans open-domain factoid QA and scientific-document QA. \textit{WebQA} \citep{chang2022webqa} is used for multimodal training and evaluation. In testing on its validation set, we report image-grounded ($n{=}2{,}511$) and text-grounded ($n{=}2{,}455$) subsets separately, while retrieval itself remains unified over a mixed text-image candidate pool. We also consider multimodal domain transfer on SlideQA in Appendix~\ref{app:multimodal_super_ood}. For the Text-only setting, all evidence is processed in the ``Title: Sentence`` form. We process evidence as ``Title: Evidence`` and ``Caption: Image`` forms for the multimodal setting.

\textbf{Baselines.}
We include two categories of baselines. \textbf{(1)} \textbf{Context-based baselines}, including generation with full-context and evidence-compression baselines \textit{LLMLingua} \citep{jiang2023llmlingua}, xRAG \citep{cheng2024xrag}, and CLaRa \citep{he2025clara}. \textbf{(2)} \textbf{RAG baselines}, including BM25 Retrieval \citep{robertson2009probabilistic}, Dense Retrieval \citep{karpukhin2020dense}, Retrieval with practical off-the-shelf Qwen-3-Embedding \citep{zhang2025qwen3} (and its variant fine-tuned on in-domain), Nemo Retriever \citep{xu2025llama}, and Qwen-3-VL-Embedding \citep{li2026qwen3}. All baselines and the proposed Latent Memory-based QA use the same pre-trained generator. Baseline details are in Appendix~\ref{app:assets}.

\textbf{Metrics.}
For text-only QA, we report Exact Match (EM), Token F1, Recall@$k$, and average generator input tokens (\#Tok). For WebQA, we report F1, answer accuracy (Acc), Recall@$k$, and \#Tok under the same unified retrieval setting. Acc follows the official WebQA evaluation protocol \citep{chang2022webqa}. In all tables, \#Tok reports the task-relevant generator-side prompt budget (including the context length and the query length) and excludes fixed chat-template scaffolding; in multimodal settings, it still reflects the effective tokenized cost after visual expansion.

\begin{figure*}[t]
\centering

\subfigure[LLaMA-8B]{\includegraphics[width=0.45\linewidth]{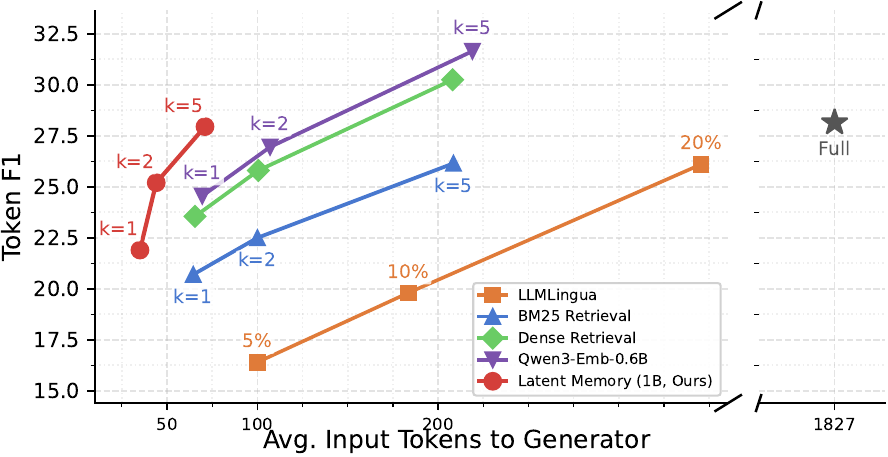}}\qquad
\subfigure[Mistral-7B-Instruct]{\includegraphics[width=0.45\linewidth]{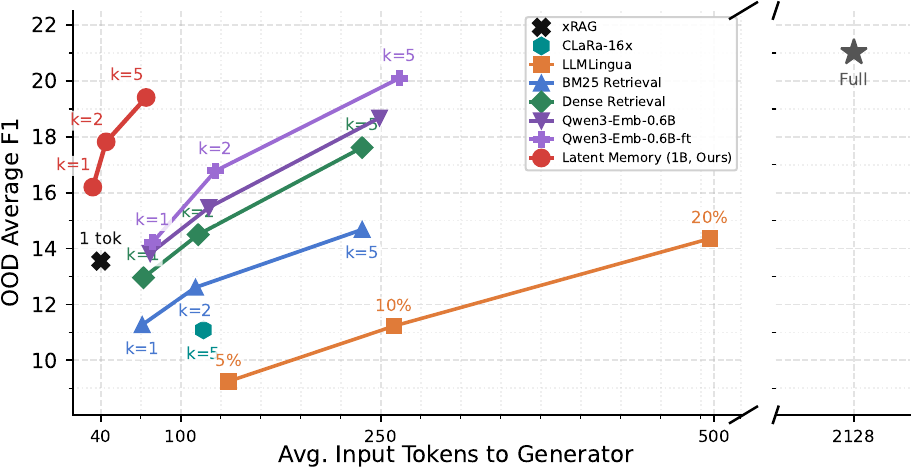}}
\vspace{-3pt}
\caption{Text-only QA trade-off curves on the \textbf{out-of-domain average F1} over 2WikiMultihopQA and MuSiQue.}
\label{fig:curve}\vspace{-8pt}
\end{figure*}

% ============================================================
\subsection{Text-only Setting}
% ============================================================

Table~\ref{tab:text_results} presents metrics of the proposed Latent Memory, context-based, and RAG baselines under a fixed 8B LLaMA generator. Qwen3-Emb-0.6B means using the Qwen3-Emedding-0.6B \citep{zhang2025qwen3} for retrieval, which is a size similar to our 1B compressor. One-token Latent Memory shows the strongest out-of-domain average Recall@$k$ among the reported text-only retrieval methods. As shown in Figure~\ref{fig:curve}, Latent Memory achieves competitive EM/F1 performance with much fewer tokens. At $k{=}5$, the 1B model uses only 71 tokens on the out-of-domain average, about a third versus 209 for BM25 and 208 for Dense. We discuss the time complexity in Appendix \ref{app:time_complexity}. These results indicate the superiority of using the same latent representation space for both retrieval and generation evidence.

Table~\ref{tab:mistral_text_results} further evaluates the same text-only setting under a frozen Mistral-7B-Instruct generator, which allows direct comparison with pretrained Mistral-based latent-context baselines such as xRAG and CLaRa. In this setting, our Latent Memory still uses the LLaMA-3.2-1B encoder/compressor and decoder. As shown in Figure~\ref{fig:curve}, Ours-1B at $k{=}5$ reaches the strongest out-of-domain Recall@$k$ and competitive F1 while using 74 generator tokens, compared with 264 tokens for Qwen3-Emb-0.6B \textbf{fine-tuned on HotpotQA} at $k{=}5$.

Importantly, the retrieval behavior is not only an in-domain effect: the same HotpotQA-trained Latent Memory is evaluated without additional tuning on \textbf{out-of-domain} 2WikiMultihopQA and MuSiQue. Thus, the comparison tests the transfer of the learned latent interface rather than dataset-specific memorization. In Appendix \ref{app:text_super_ood}, we generalize the Latent Memory with a compressor trained on HotpotQA to four more datasets, where Latent Memory can still demonstrate a strong performance-efficiency trade-off. This competitive performance may be partly explained by its stronger retrieval behavior. Latent Memory reaches an out-of-domain average Recall@k of 52.2 at $k=5$.

\begin{table}[t]
\centering
\caption{Multimodal QA (WebQA) with unified per-sample retrieval and unified generation over both images and text facts using a frozen \textit{LLaVA-1.5-13B} generator.}
\label{tab:mm_results}
\small
\renewcommand\arraystretch{1.03}
\setlength{\tabcolsep}{6pt}
\resizebox{\textwidth}{!}{
\begin{tabular}{lcccc cccc |cccc}
\toprule[0.5mm]
\multicolumn{13}{c}{\textbf{Generation VLM (fixed): LLaVA-1.5-13B}} \\
\midrule
Method & \multicolumn{4}{c}{\textbf{WebQA-Image}} & \multicolumn{4}{c}{\textbf{WebQA-Text}} & \multicolumn{4}{c}{\textbf{Average}} \\
\cmidrule(lr){2-5}\cmidrule(lr){6-9}\cmidrule(lr){10-13}
 & F1 & Acc & R@$k$ & \#Tok & F1 & Acc & R@$k$ & \#Tok & F1 & Acc & R@$k$ & \#Tok \\
\midrule
Full Context
& 0.0 & 0.0 & - & 11655
& 6.0 & 10.0 & - & 8371
& 3.0 & 5.0 & -- & 10013 \\
\midrule
\multicolumn{13}{l}{\textit{RAG Baselines: Text-only Retrieval Baselines}} \\
BM25 Retrieval ($k{=}1$) & 14.5 & 21.5 & 21.8 & 295 & 39.9 & 42.7 & 31.8 & 133 & 27.2 & 32.1 & 26.8 & 214 \\
BM25 Retrieval ($k{=}2$) & 19.2 & 23.4 & 28.3 & 476 & 43.9 & 48.4 & 51.0 & 234 & 31.6 & 35.9 & 39.6 & 355 \\
BM25 Retrieval ($k{=}5$) & 32.6 & 29.5 & 37.9 & 932 & 46.7 & 54.3 & 73.0 & 552 & 39.6 & 41.9 & 55.5 & 742 \\
Dense Retrieval ($k{=}1$) & 18.9 & 24.5 & 39.7 & 476 & 34.9 & 40.4 & 25.5 & 264 & 26.9 & 32.4 & 32.6 & 370 \\
Dense Retrieval ($k{=}2$) & 30.0 & 30.1 & 53.2 & 814 & 38.5 & 49.3 & 40.1 & 538 & 34.2 & 39.7 & 46.6 & 676 \\
Dense Retrieval ($k{=}5$) & 49.8 & 36.6 & 67.1 & 1645 & 37.6 & \textbf{59.5} & 62.0 & 1396 & 43.7 & \textbf{48.1} & 64.6 & 1520 \\
\midrule
\multicolumn{13}{l}{\textit{RAG Baselines: Multimodal Retrieval Baselines}} \\
Nemo-Emb-1B ($k{=}1$) & 19.9 & 25.2 & 48.7 & 507 & 41.1 & 44.0 & 40.4 & 130 & 30.5 & 34.6 & 44.6 & 319 \\
Nemo-Emb-1B ($k{=}2$) & 32.9 & 30.6 & 64.8 & 892 & 46.8 & 50.9 & 66.6 & 233 & 39.9 & 40.8 & 65.7 & 563 \\
Nemo-Emb-1B ($k{=}5$) & 53.0 & 37.2 & \underline{81.4} & 1885 & \textbf{48.6} & \underline{57.9} & \underline{87.1} & 629 & \textbf{50.8} & \underline{47.6} & \textbf{84.3} & 1257 \\
Qwen3-VL-Emb-8B ($k{=}1$) & 15.1 & 22.0 & 24.1 & 284 & 40.8 & 43.6 & 40.7 & 131 & 28.0 & 32.8 & 32.4 & 208 \\
Qwen3-VL-Emb-8B ($k{=}2$) & 20.1 & 24.5 & 32.9 & 465 & 46.1 & 50.2 & 66.4 & 235 & 33.1 & 37.3 & 49.6 & 350 \\
Qwen3-VL-Emb-8B ($k{=}5$) & 34.1 & 30.3 & 49.7 & 957 & \underline{47.8} & 57.2 & \textbf{87.2} & 612 & 41.0 & 43.8 & 68.5 & 784 \\
\midrule
\multicolumn{13}{l}{\textit{Ours: Generation with retrieving Latent Memory}} \\
Latent Memory ($k{=}1$) & 32.0 & 28.7 & 56.6 & 42 & 28.8 & 30.8 & 24.0 & 44 & 30.4 & 29.7 & 40.3 & 43 \\
Latent Memory ($k{=}2$) & \underline{56.5} & \underline{39.5} & 74.7 & 52 & 30.0 & 32.8 & 41.1 & 54 & 43.2 & 36.2 & 57.9 & 53 \\
Latent Memory ($k{=}5$) & \textbf{69.4} & \textbf{44.2} & \textbf{91.2} & 82 & 30.7 & 34.3 & 70.5 & 84 & \underline{50.0} & 39.2 & \underline{80.8} & 83 \\
\bottomrule[0.5mm]
\end{tabular}}
\end{table}

\begin{wrapfigure}{r}{0.48\textwidth}
\centering\vspace{-10pt}
\includegraphics[width=0.96\linewidth]{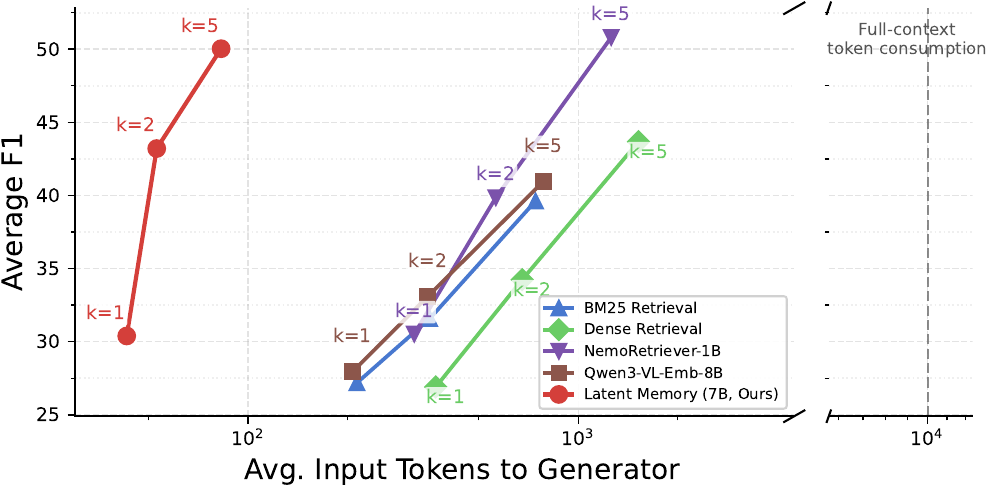}
\caption{LLaVA-based multimodal WebQA trade-off curves across $k{\in}\{1,2,5\}$.}
\label{fig:efficiency}\vspace{-12pt}
\end{wrapfigure}

\vspace{-2pt}
% ============================================================
\subsection{Multimodal Setting}
% ============================================================
\vspace{-2pt}

We report the Latent Memory and baselines on the WebQA benchmark, which requires unified retrieval and multi-hop reasoning over a multimodal candidate pool. For BM25 and Dense retrieval, we retrieve only the caption for the image evidence. Table~\ref{tab:mm_results} shows the image-grounded (\(n{=}2{,}511\)) and text-grounded (\(n{=}2{,}455\)) WebQA subsets separately. One-token Latent Memory is strongest on the image-grounded subset while using far fewer generator tokens than raw-evidence retrieval baselines: at \(k{=}5\), it reaches 69.4 image F1 with only 82 tokens, compared with 53.0 for Nemo-Emb at 1885 tokens. On the full benchmark, Latent Memory gives a competitive average F1 with a much smaller token budget, while Nemo-Emb gives the best average F1/Acc at substantially higher cost.

Similar to the text-only setting, this behavior can be attributed to the fact that the unified representation of Latent Memory leads to a better Recall@K (10\%+ higher compared to Dense Retrieval) on text-grounded and especially image-grounded questions. Moreover, as another reason, Raw-image prompting may exceed the pretrained context window for the generator, leading to poor-quality output (full-context often outputs meaningless content or blanks). To intuitively reflect the generation process augmented with Latent Memory, we provide a case study in Section~\ref{case_study_section}.

\vspace{-5pt}
% ============================================================
\section{Discussion}
% ============================================================
\vspace{-3pt}

\textbf{Ablations and Analysis.} Table~\ref{tab:main_discussion_recon_ablation} conducts an ablation study over the core reconstruction loss $\mathcal{L}_\text{Recon}$, where the results are averaged over HotpotQA, 2WikiMultihopQA, and MuSiQue. Removing reconstruction lowers both answer quality and retrieval accuracy, with a larger drop in EM/F1 than in Recall@$k$. Removing negative evidence from reconstruction also hurts retrieval and generation, supporting the view that negative evidence helps anchor the unified latent representation space. The full per-dataset breakdown and more ablation variants are reported in Appendix~\ref{app:ablation_recon} to ~\ref{app:generator_transfer}.

\begin{table}[H]
\centering
\caption{Ablation on reconstruction. Colored subscripts indicate the gap to the default model.}
\label{tab:main_discussion_recon_ablation}
\setlength{\tabcolsep}{10pt}
\resizebox{\linewidth}{!}{
\begin{tabular}{lccc ccc ccc}
\toprule[0.5mm]
\multirow{2}{*}{\textbf{$k$}} & \multicolumn{3}{c}{\shortstack{Original Latent Memory}} & \multicolumn{3}{c}{\shortstack{w/o Reconstruction Loss $\mathcal{L}_\text{Recon}$}} & \multicolumn{3}{c}{\shortstack{w/o $\mathcal{L}_\text{Recon}$ on Negative Evidence $\mathcal{M}^{-}$}} \\
\cmidrule(lr){2-4}\cmidrule(lr){5-7}\cmidrule(lr){8-10}
 & \textbf{EM} & \textbf{F1} & \textbf{R@$k$} & \textbf{EM} & \textbf{F1} & \textbf{R@$k$} & \textbf{EM} & \textbf{F1} & \textbf{R@$k$} \\
\midrule
\textbf{$k{=}1$} & 17.7 & 27.7 & 23.9 & 16.5$_{\textcolor{red}{-1.1}}$ & 27.2$_{\textcolor{red}{-0.5}}$ & 23.8$_{\textcolor{red}{-0.1}}$ & 16.3$_{\textcolor{red}{-1.3}}$ & 28.1$_{\textcolor{ourgreen}{+0.3}}$ & 23.2$_{\textcolor{red}{-0.6}}$ \\
\textbf{$k{=}2$} & 20.1 & 31.9 & 42.6 & 19.9$_{\textcolor{red}{-0.2}}$ & 31.0$_{\textcolor{red}{-0.9}}$ & 41.8$_{\textcolor{red}{-0.7}}$ & 18.4$_{\textcolor{red}{-1.7}}$ & 32.0$_{\textcolor{ourgreen}{+0.2}}$ & 39.6$_{\textcolor{red}{-3.0}}$ \\
\textbf{$k{=}5$} & 22.6 & 34.9 & 63.8 & 20.9$_{\textcolor{red}{-1.7}}$ & 32.7$_{\textcolor{red}{-2.2}}$ & 62.9$_{\textcolor{red}{-0.9}}$ & 20.8$_{\textcolor{red}{-1.9}}$ & 33.8$_{\textcolor{red}{-1.2}}$ & 62.3$_{\textcolor{red}{-1.5}}$ \\
\bottomrule[0.5mm]
\end{tabular}}
\end{table}
\textbf{Better Latent Memory Capability with more Token Budget.}
One-token Latent Memory already gives a strong efficiency--quality trade-off, and allocating more latent tokens per evidence item can improve the quality. As summarized in Table~\ref{tab:main_discussion_token_effect}, on the out-of-domain average over 2WikiMultihopQA and MuSiQue, upgrading to 8-token Latent Memory improves EM/F1 enough to surpass the strongest text-only RAG baseline (i.e., Qwen-3-Emb-0.6B) at each $k$, while still using fewer generator tokens. The gain mainly comes from stronger generation rather than retrieval, since Recall@$k$ changes only modestly. Full in-domain and out-of-domain results are in Appendix~\ref{app:ablation}.
\begin{table}[H]
\centering
\caption{Token-count ablation summary. We show the average Out-of-domain (2WikiMultihopQA and MuSiQue results). RAG* denotes the strongest RAG baseline at the same $k$.}
\label{tab:main_discussion_token_effect}
\setlength{\tabcolsep}{1.8pt}
\resizebox{\textwidth}{!}{
\begin{tabular}{ccccc|cccccccc|cccc}
\toprule[0.5mm]
\multirow{2}{*}{\textbf{$k$}} & \multicolumn{4}{c}{\textbf{RAG*}} & \multicolumn{4}{c}{\textbf{1-token Latent Memory}} & \multicolumn{4}{c}{\textbf{8-token Latent Memory}} & \multicolumn{4}{c}{\textbf{$\Delta$ (8-token $-$ 1-token)}} \\
\cmidrule(lr){2-5}\cmidrule(lr){6-9}\cmidrule(lr){10-13}\cmidrule(lr){14-17}
 & \textbf{EM} & \textbf{F1} & \textbf{R@$k$} & \textbf{\#Tok} & \textbf{EM} & \textbf{F1} & \textbf{R@$k$} & \textbf{\#Tok} & \textbf{EM} & \textbf{F1} & \textbf{R@$k$} & \textbf{\#Tok} & \textbf{EM} & \textbf{F1} & \textbf{R@$k$} & \textbf{\#Tok} \\
\midrule
$k{=}1$ & 13.1 & 24.6 & 21.4 & 70 & 12.8$_{\textcolor{red}{-0.3}}$ & 21.9$_{\textcolor{red}{-2.7}}$ & 18.6 & 35$_{\textcolor{ourgreen}{-35}}$ & \textbf{14.4}$_{\textcolor{ourgreen}{+1.3}}$ & \textbf{24.7}$_{\textcolor{ourgreen}{+0.1}}$ & 19.9 & 42$_{\textcolor{ourgreen}{-28}}$ & +1.6 & +2.8 & +1.3 & +7 \\
$k{=}2$ & 13.7 & 27.0 & 33.1 & 107 & 14.4$_{\textcolor{ourgreen}{+0.7}}$ & 25.2$_{\textcolor{red}{-1.8}}$ & 32.5 & 44$_{\textcolor{ourgreen}{-63}}$ & \textbf{17.3}$_{\textcolor{ourgreen}{+3.6}}$ & \textbf{29.0}$_{\textcolor{ourgreen}{+2.0}}$ & 34.5 & 58$_{\textcolor{ourgreen}{-49}}$ & +2.9 & +3.8 & +2.0 & +14 \\
$k{=}5$ & 16.4 & 31.7 & 49.6 & 219 & 16.5$_{\textcolor{ourgreen}{+0.1}}$ & 28.0$_{\textcolor{red}{-3.7}}$ & 52.2 & 71$_{\textcolor{ourgreen}{-148}}$ & \textbf{19.9}$_{\textcolor{ourgreen}{+3.5}}$ & \textbf{32.5}$_{\textcolor{ourgreen}{+0.9}}$ & 53.7 & 106$_{\textcolor{ourgreen}{-113}}$ & +3.4 & +4.6 & +1.5 & +35 \\
\bottomrule[0.5mm]
\end{tabular}}
\end{table}

\begin{figure}[t]
\centering\vspace{-5pt}
\includegraphics[width=\textwidth]{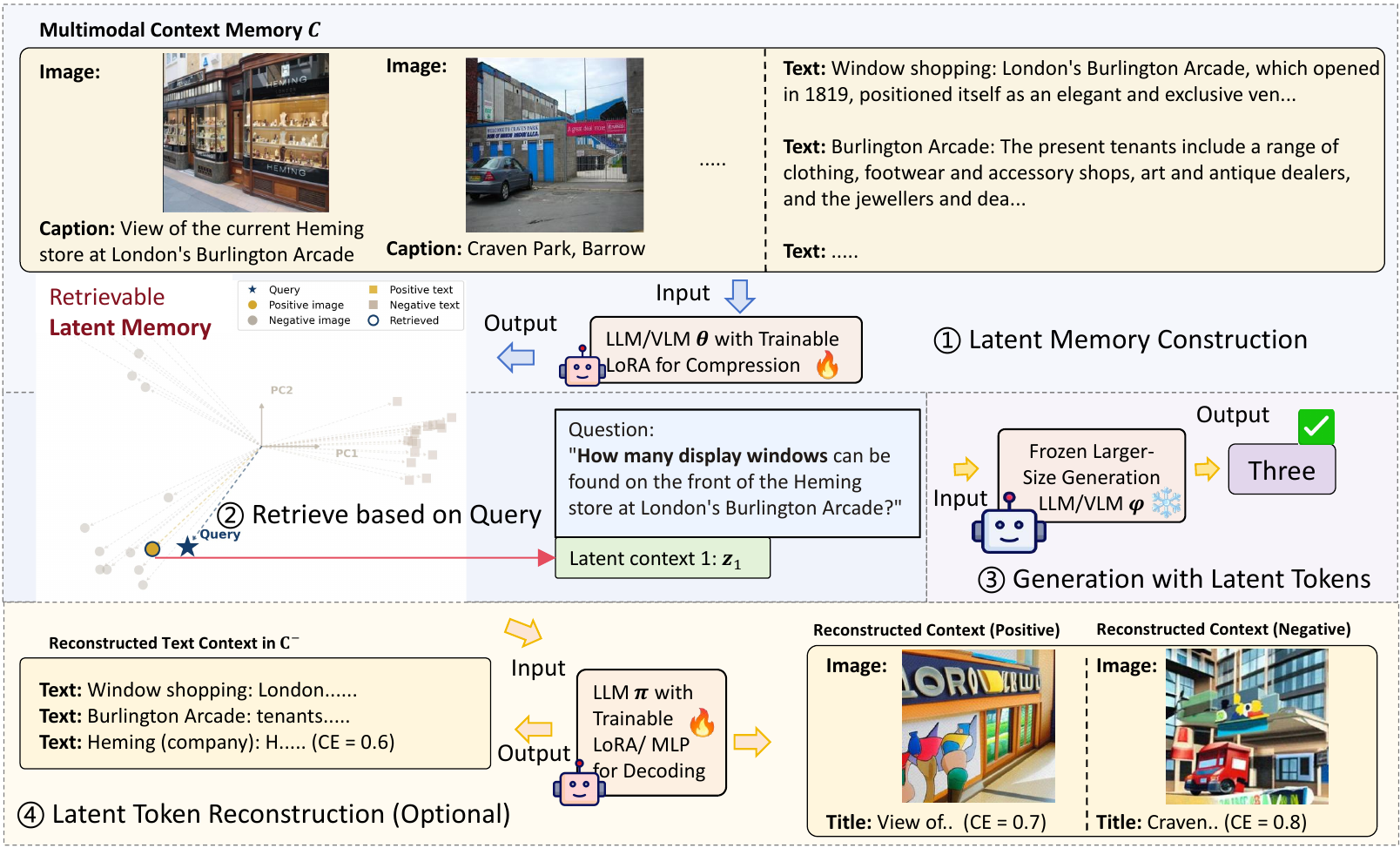}
\caption{LLaVA-1.5-based Latent Memory on an image-grounded WebQA question. The figure consists of four parts. \textbf{\ding{172}} Compressing multimodal evidence forms a unified Latent Memory. \textbf{\ding{173}} The retrieval process aligns the query embedding with the latent token with the positive image, while irrelevant candidates are pushed away in the unified latent space. \textbf{\ding{174}} The Latent Memory grounded QA preserves the counting ability. \textbf{\ding{175}} The optional reconstruction for both image and text evidence.}
\label{fig:case_study_main}\vspace{-3pt}
\end{figure}

\textbf{A Representative Case Study.}\label{case_study_section}
Figure~\ref{fig:case_study_main} illustrates how Latent Memory behaves on an image-grounded WebQA example. The retrieved latent evidence supports the correct answer and preserves the counting information required by the question. The optional reconstruction further shows that the latent token retains interpretable semantic content for both text and image evidence, rather than acting only as an opaque retrieval identifier.

\textbf{Limitation on Current Modality Coverage and Future Directions.}\label{Limitation_and_future_work}
The current design assumes that evidence can be decomposed into atomic text or image units, and each unit can be compressed and retrieved independently. This assumption is reasonable for WebQA-style mixed evidence, where answers are often grounded in a small number of facts or images. However, it becomes limiting when the input meaning depends on the global structure. Complex tables require row-column relations and layout information; long videos require temporal ordering; and document pages may require spatial relations between captions, figures, and surrounding text. Compressing such inputs into isolated latent tokens may preserve local semantics but lose these structural dependencies. A natural next step is therefore to augment Latent Memory with structural axes such as position, layout, and time, so that retrieval and generation can operate over both local evidence semantics and global organization.

% ============================================================
\section{Conclusion}
% ============================================================

We introduced Latent Memory, a novel memory paradigm that compiles each evidence item into one latent token, retrieves these latent memories by compressing the query, and feeds them directly to a frozen generator for QA. On text QA, it achieves competitive EM/F1 performance compared to RAG baselines with only 3$\times$ fewer tokens. On multimodal WebQA, it is especially effective for image-grounded questions and delivers the strongest average-F1 trade-off while sharply reducing generator cost by up to  10$\times$. Latent Memory improves the Recall@$k$ by employing a unified latent representation and prevents the context from exceeding the generator's context window length.

\paragraph{Future Work}
Overall, Latent Memory provides an efficient alternative to the current token-level memory paradigm. It is promising for use in scenarios that require fast response and low storage pressure, such as edge devices and other resource-intensive scenarios. In the future, as discussed in Section \ref{Limitation_and_future_work}, we will extend the Latent Memory to more modalities, including complex tables and complex videos. This paper focuses on external evidence, so we do not consider Agentic Memory \citep{xu2025mem,wang2025mirix,chhikara2025mem0}, which are generated by models themselves. Future works also include extending the Latent Memory to resource-contained agents scenarios for better Agentic Memory comprehension.

% ============================================================

{
\small
\bibliographystyle{unsrt}
\bibliography{ref}

@article{ramesh2022hierarchical,
  title={Hierarchical text-conditional image generation with clip latents},
  author={Ramesh, Aditya and Dhariwal, Prafulla and Nichol, Alex and Chu, Casey and Chen, Mark},
  journal={arXiv preprint arXiv:2204.06125},
  volume={1},
  number={2},
  pages={3},
  year={2022}
}

@article{chhikara2025mem0,
  title={Mem0: Building production-ready ai agents with scalable long-term memory},
  author={Chhikara, Prateek and Khant, Dev and Aryan, Saket and Singh, Taranjeet and Yadav, Deshraj},
  journal={arXiv preprint arXiv:2504.19413},
  year={2025}
}

@article{wang2025mirix,
  title={Mirix: Multi-agent memory system for llm-based agents},
  author={Wang, Yu and Chen, Xi},
  journal={arXiv preprint arXiv:2507.07957},
  year={2025}
}

@inproceedings{jimenez2024swe,
  title={Swe-bench: Can language models resolve real-world github issues?},
  author={Jimenez, Carlos E and Yang, John and Wettig, Alexander and Yao, Shunyu and Pei, Kexin and Press, Ofir and Narasimhan, Karthik},
  booktitle={International Conference on Learning Representations},
  volume={2024},
  pages={54107--54157},
  year={2024}
}

@article{jiang2023mistral,
  title={6G non-terrestrial networks enabled low-altitude economy: Opportunities and challenges},
  author={Jiang, Yihang and Li, Xiaoyang and Zhu, Guangxu and Li, Hang and Deng, Jing and Han, Kaifeng and Shen, Chao and Shi, Qingjiang and Zhang, Rui},
  journal={arXiv preprint arXiv:2311.09047},
  year={2023}
}

@article{jiang2024longrag,
  title={Longrag: Enhancing retrieval-augmented generation with long-context llms},
  author={Jiang, Ziyan and Ma, Xueguang and Chen, Wenhu},
  journal={arXiv preprint arXiv:2406.15319},
  year={2024}
}

@inproceedings{dasigi2021qasper,
  title={A dataset of information-seeking questions and answers anchored in research papers},
  author={Dasigi, Pradeep and Lo, Kyle and Beltagy, Iz and Cohan, Arman and Smith, Noah A and Gardner, Matt},
  booktitle={Proceedings of the 2021 Conference of the North American Chapter of the Association for Computational Linguistics: Human Language Technologies},
  pages={4599--4610},
  year={2021}
}

@article{Kamath2025Gemma3T,
  title={Gemma 3 Technical Report},
  author={Gemma Team},
  journal={ArXiv},
  year={2025},
  volume={abs/2503.19786},
  url={https://api.semanticscholar.org/CorpusID:277313563}
}

@article{yao2024minicpm,
  title={Minicpm-v: A gpt-4v level mllm on your phone},
  author={Yao, Yuan and Yu, Tianyu and Zhang, Ao and Wang, Chongyi and Cui, Junbo and Zhu, Hongji and Cai, Tianchi and Li, Haoyu and Zhao, Weilin and He, Zhihui and others},
  journal={arXiv preprint arXiv:2408.01800},
  year={2024}
}

@article{kwiatkowski2019natural,
  title={Natural questions: a benchmark for question answering research},
  author={Kwiatkowski, Tom and Palomaki, Jennimaria and Redfield, Olivia and Collins, Michael and Parikh, Ankur and Alberti, Chris and Epstein, Danielle and Polosukhin, Illia and Devlin, Jacob and Lee, Kenton and others},
  journal={Transactions of the Association for Computational Linguistics},
  volume={7},
  pages={453--466},
  year={2019},
  publisher={MIT Press One Rogers Street, Cambridge, MA 02142-1209, USA journals-info~…}
}

@inproceedings{joshi2017triviaqa,
  title={Triviaqa: A large scale distantly supervised challenge dataset for reading comprehension},
  author={Joshi, Mandar and Choi, Eunsol and Weld, Daniel S and Zettlemoyer, Luke},
  booktitle={Proceedings of the 55th Annual Meeting of the Association for Computational Linguistics (Volume 1: Long Papers)},
  pages={1601--1611},
  year={2017}
}

@article{zhang2025memgen,
  title={Memgen: Weaving generative latent memory for self-evolving agents},
  author={Zhang, Guibin and Fu, Muxin and Yan, Shuicheng},
  journal={arXiv preprint arXiv:2509.24704},
  year={2025}
}

@article{fu2026latentmem,
  title={LatentMem: Customizing Latent Memory for Multi-Agent Systems},
  author={Fu, Muxin and Xue, Xiangyuan and Li, Yafu and He, Zefeng and Huang, Siyuan and Qu, Xiaoye and Cheng, Yu and Yang, Yang},
  journal={arXiv preprint arXiv:2602.03036},
  year={2026}
}

@article{yu2026latent,
  title={The Latent Space: Foundation, Evolution, Mechanism, Ability, and Outlook},
  author={Yu, Xinlei and Chen, Zhangquan and He, Yongbo and Fu, Tianyu and Yang, Cheng and Xu, Chengming and Ma, Yue and Hu, Xiaobin and Cao, Zhe and Xu, Jie and others},
  journal={arXiv preprint arXiv:2604.02029},
  year={2026}
}

@inproceedings{radford2021learning,
  title={Learning transferable visual models from natural language supervision},
  author={Radford, Alec and Kim, Jong Wook and Hallacy, Chris and Ramesh, Aditya and Goh, Gabriel and Agarwal, Sandhini and Sastry, Girish and Askell, Amanda and Mishkin, Pamela and Clark, Jack and others},
  booktitle={International conference on machine learning},
  pages={8748--8763},
  year={2021},
  organization={PmLR}
}

@article{touvron2023llama,
  title={Llama: Open and efficient foundation language models},
  author={Touvron, Hugo and Lavril, Thibaut and Izacard, Gautier and Martinet, Xavier and Lachaux, Marie-Anne and Lacroix, Timoth{\'e}e and Rozi{\`e}re, Baptiste and Goyal, Naman and Hambro, Eric and Azhar, Faisal and others},
  journal={arXiv preprint arXiv:2302.13971},
  year={2023}
}

@article{chen2025vl,
  title={Vl-jepa: Joint embedding predictive architecture for vision-language},
  author={Chen, Delong and Shukor, Mustafa and Moutakanni, Theo and Chung, Willy and Yu, Jade and Kasarla, Tejaswi and Bang, Yejin and Bolourchi, Allen and LeCun, Yann and Fung, Pascale},
  journal={arXiv preprint arXiv:2512.10942},
  year={2025}
}

@article{sun2026vla,
  title={Vla-jepa: Enhancing vision-language-action model with latent world model},
  author={Sun, Jingwen and Zhang, Wenyao and Qi, Zekun and Ren, Shaojie and Liu, Zezhi and Zhu, Hanxin and Sun, Guangzhong and Jin, Xin and Chen, Zhibo},
  journal={arXiv preprint arXiv:2602.10098},
  year={2026}
}

@article{nam2026causal,
  title={Causal-JEPA: Learning World Models through Object-Level Latent Interventions},
  author={Nam, Heejeong and Lidec, Quentin Le and Maes, Lucas and LeCun, Yann and Balestriero, Randall},
  journal={arXiv preprint arXiv:2602.11389},
  year={2026}
}

@article{zheng2025soft,
  title={SofT-GRPO: Surpassing Discrete-Token LLM Reinforcement Learning via Gumbel-Reparameterized Soft-Thinking Policy Optimization},
  author={Zheng, Zhi and Lee, Wee Sun},
  journal={arXiv preprint arXiv:2511.06411},
  year={2025}
}

@article{zhu2025survey,
  title={A survey on latent reasoning},
  author={Zhu, Rui-Jie and Peng, Tianhao and Cheng, Tianhao and Qu, Xingwei and Huang, Jinfa and Zhu, Dawei and Wang, Hao and Xue, Kaiwen and Zhang, Xuanliang and Shan, Yong and others},
  journal={arXiv preprint arXiv:2507.06203},
  year={2025}
}

@article{xu2025llama,
  title={Llama nemoretriever colembed: Top-performing text-image retrieval model},
  author={Xu, Mengyao and Moreira, Gabriel and Ak, Ronay and Osmulski, Radek and Babakhin, Yauhen and Yu, Zhiding and Schifferer, Benedikt and Oldridge, Even},
  journal={arXiv preprint arXiv:2507.05513},
  year={2025}
}

@article{li2026qwen3,
  title={Qwen3-VL-Embedding and Qwen3-VL-Reranker: A Unified Framework for State-of-the-Art Multimodal Retrieval and Ranking},
  author={Li, Mingxin and Zhang, Yanzhao and Long, Dingkun and Chen, Keqin and Song, Sibo and Bai, Shuai and Yang, Zhibo and Xie, Pengjun and Yang, An and Liu, Dayiheng and others},
  journal={arXiv preprint arXiv:2601.04720},
  year={2026}
}

@article{zhang2025qwen3,
  title={Qwen3 embedding: Advancing text embedding and reranking through foundation models},
  author={Zhang, Yanzhao and Li, Mingxin and Long, Dingkun and Zhang, Xin and Lin, Huan and Yang, Baosong and Xie, Pengjun and Yang, An and Liu, Dayiheng and Lin, Junyang and others},
  journal={arXiv preprint arXiv:2506.05176},
  year={2025}
}

@article{zheng2025monte,
  title={Monte carlo tree search for comprehensive exploration in llm-based automatic heuristic design},
  author={Zheng, Zhi and Xie, Zhuoliang and Wang, Zhenkun and Hooi, Bryan},
  journal={arXiv preprint arXiv:2501.08603},
  year={2025}
}

@inproceedings{louis2025pisco,
  title={Pisco: Pretty simple compression for retrieval-augmented generation},
  author={Louis, Maxime and D{\'e}jean, Herv{\'e} and Clinchant, St{\'e}phane},
  booktitle={Findings of the Association for Computational Linguistics: ACL 2025},
  pages={15506--15521},
  year={2025}
}

@article{oord2018representation,
  title={Representation learning with contrastive predictive coding},
  author={Oord, Aaron van den and Li, Yazhe and Vinyals, Oriol},
  journal={arXiv preprint arXiv:1807.03748},
  year={2018}
}

@inproceedings{yang2018hotpotqa,
  title={HotpotQA: A dataset for diverse, explainable multi-hop question answering},
  author={Yang, Zhilin and Qi, Peng and Zhang, Saizheng and Bengio, Yoshua and Cohen, William and Salakhutdinov, Ruslan and Manning, Christopher D},
  booktitle={Proceedings of the 2018 conference on empirical methods in natural language processing},
  pages={2369--2380},
  year={2018}
}

@article{liu2025comprehensive,
  title={A comprehensive survey on long context language modeling},
  author={Liu, Jiaheng and Zhu, Dawei and Bai, Zhiqi and He, Yancheng and Liao, Huanxuan and Que, Haoran and Wang, Zekun and Zhang, Chenchen and Zhang, Ge and Zhang, Jiebin and others},
  journal={arXiv preprint arXiv:2503.17407},
  year={2025}
}

@article{trivedi2022musique,
  title={MuSiQue: Multihop Questions via Single-hop Question Composition},
  author={Trivedi, Harsh and Balasubramanian, Niranjan and Khot, Tushar and Sabharwal, Ashish},
  journal={Transactions of the Association for Computational Linguistics},
  volume={10},
  pages={539--554},
  year={2022},
  publisher={MIT Press One Broadway, 12th Floor, Cambridge, Massachusetts 02142, USA~…}
}

@inproceedings{chang2022webqa,
  title={Webqa: Multihop and multimodal qa},
  author={Chang, Yingshan and Narang, Mridu and Suzuki, Hisami and Cao, Guihong and Gao, Jianfeng and Bisk, Yonatan},
  booktitle={Proceedings of the IEEE/CVF conference on computer vision and pattern recognition},
  pages={16495--16504},
  year={2022}
}

@inproceedings{ho20202wikimultihop,
  title={Constructing a multi-hop qa dataset for comprehensive evaluation of reasoning steps},
  author={Ho, Xanh and Nguyen, Anh-Khoa Duong and Sugawara, Saku and Aizawa, Akiko},
  booktitle={Proceedings of the 28th International Conference on Computational Linguistics},
  pages={6609--6625},
  year={2020}
}

@article{kamoi2023wice,
  title={Wice: Real-world entailment for claims in wikipedia},
  author={Kamoi, Ryo and Goyal, Tanya and Rodriguez, Juan Diego and Durrett, Greg},
  journal={arXiv preprint arXiv:2303.01432},
  year={2023}
}

@article{dmonte2024claim,
  title={Claim verification in the age of large language models: A survey},
  author={Dmonte, Alphaeus and Oruche, Roland and Zampieri, Marcos and Calyam, Prasad and Augenstein, Isabelle},
  journal={arXiv preprint arXiv:2408.14317},
  year={2024}
}

@article{yu2025vismem,
  title={Vismem: Latent vision memory unlocks potential of vision-language models},
  author={Yu, Xinlei and Xu, Chengming and Zhang, Guibin and Chen, Zhangquan and Zhang, Yudong and He, Yongbo and Jiang, Peng-Tao and Zhang, Jiangning and Hu, Xiaobin and Yan, Shuicheng},
  journal={arXiv preprint arXiv:2511.11007},
  year={2025}
}

@article{wang2025monet,
  title={Monet: Reasoning in latent visual space beyond images and language},
  author={Wang, Qixun and Shi, Yang and Wang, Yifei and Zhang, Yuanxing and Wan, Pengfei and Gai, Kun and Ying, Xianghua and Wang, Yisen},
  journal={arXiv preprint arXiv:2511.21395},
  year={2025}
}

@article{liu2025efficient,
  title={Efficient inference for large reasoning models: A survey},
  author={Liu, Yue and Wu, Jiaying and He, Yufei and Gong, Ruihan and Xia, Jun and Li, Liang and Gao, Hongcheng and Chen, Hongyu and Bi, Baolong and Zhang, Jiaheng and others},
  journal={arXiv preprint arXiv:2503.23077},
  year={2025}
}

@article{zheng2025reasoning,
  title={Reasoning-CV: Fine-tuning Powerful Reasoning LLMs for Knowledge-Assisted Claim Verification},
  author={Zheng, Zhi and Lee, Wee Sun},
  journal={arXiv preprint arXiv:2505.12348},
  year={2025}
}

@article{wu2024retrieval,
  title={Retrieval-augmented generation for natural language processing: A survey},
  author={Wu, Shangyu and Xiong, Ying and Cui, Yufei and Wu, Haolun and Chen, Can and Yuan, Ye and Huang, Lianming and Liu, Xue and Kuo, Tei-Wei and Guan, Nan and others},
  journal={arXiv preprint arXiv:2407.13193},
  year={2024}
}

@article{guo2024lightrag,
  title={LightRAG: Simple and Fast Retrieval-Augmented Generation},
  author={Guo, Zirui and Xia, Lianghao and Yu, Yanhua and Ao, Tu and Huang, Chao},
  year={2024}
}

@article{wu2025llms,
  title={LLMs are Single-threaded Reasoners: Demystifying the Working Mechanism of Soft Thinking},
  author={Wu, Ch{\"u}nhung and Lu, Jinliang and Ren, Zixuan and Hu, Gangqiang and Wu, Zhi and Dai, Dai and Wu, Hua},
  journal={arXiv preprint arXiv:2508.03440},
  year={2025}
}

@article{butt2025soft,
  title={Soft Tokens, Hard Truths},
  author={Butt, Natasha and Kwiatkowski, Ariel and Labiad, Ismail and Kempe, Julia and Ollivier, Yann},
  journal={arXiv preprint arXiv:2509.19170},
  year={2025}
}

@article{zhang2025soft,
  title={Soft thinking: Unlocking the reasoning potential of llms in continuous concept space},
  author={Zhang, Zhen and He, Xuehai and Yan, Weixiang and Shen, Ao and Zhao, Chenyang and Wang, Shuohang and Shen, Yelong and Wang, Xin Eric},
  journal={arXiv preprint arXiv:2505.15778},
  year={2025}
}

@article{wei2025sim,
  title={SIM-CoT: Supervised Implicit Chain-of-Thought},
  author={Wei, Xilin and Liu, Xiaoran and Zang, Yuhang and Dong, Xiaoyi and Cao, Yuhang and Wang, Jiaqi and Qiu, Xipeng and Lin, Dahua},
  journal={arXiv preprint arXiv:2509.20317},
  year={2025}
}

@article{chen2025reasoning,
  title={Reasoning Beyond Language: A Comprehensive Survey on Latent Chain-of-Thought Reasoning},
  author={Chen, Xinghao and Zhao, Anhao and Xia, Heming and Lu, Xuan and Wang, Hanlin and Chen, Yanjun and Zhang, Wei and Wang, Jian and Li, Wenjie and Shen, Xiaoyu},
  journal={arXiv preprint arXiv:2505.16782},
  year={2025}
}

@article{shen2025codi,
  title={Codi: Compressing chain-of-thought into continuous space via self-distillation},
  author={Shen, Zhenyi and Yan, Hanqi and Zhang, Linhai and Hu, Zhanghao and Du, Yali and He, Yulan},
  journal={arXiv preprint arXiv:2502.21074},
  year={2025}
}

@article{hao2024training,
  title={Training large language models to reason in a continuous latent space},
  author={Hao, Shibo and Sukhbaatar, Sainbayar and Su, DiJia and Li, Xian and Hu, Zhiting and Weston, Jason and Tian, Yuandong},
  journal={arXiv preprint arXiv:2412.06769},
  year={2024}
}

@article{li2026latent,
  title={Latent Context Compilation: Distilling Long Context into Compact Portable Memory},
  author={Li, Zeju and Zhou, Yizhou and Xu, Qiang},
  journal={arXiv preprint arXiv:2602.21221},
  year={2026}
}

@inproceedings{assran2023self,
  title={Self-supervised learning from images with a joint-embedding predictive architecture},
  author={Assran, Mahmoud and Duval, Quentin and Misra, Ishan and Bojanowski, Piotr and Vincent, Pascal and Rabbat, Michael and LeCun, Yann and Ballas, Nicolas},
  booktitle={Proceedings of the IEEE/CVF conference on computer vision and pattern recognition},
  pages={15619--15629},
  year={2023}
}

@article{he2025clara,
  title={CLaRa: Bridging Retrieval and Generation with Continuous Latent Reasoning},
  author={He, Jie and Bai, Richard He and Williamson, Sinead and Pan, Jeff Z and Jaitly, Navdeep and Zhang, Yizhe},
  journal={arXiv preprint arXiv:2511.18659},
  year={2025}
}

@article{cheng2024xrag,
  title={xrag: Extreme context compression for retrieval-augmented generation with one token},
  author={Cheng, Xin and Wang, Xun and Zhang, Xingxing and Ge, Tao and Chen, Si-Qing and Wei, Furu and Zhang, Huishuai and Zhao, Dongyan},
  journal={Advances in Neural Information Processing Systems},
  volume={37},
  pages={109487--109516},
  year={2024}
}

@inproceedings{jiang2023llmlingua,
  title={Llmlingua: Compressing prompts for accelerated inference of large language models},
  author={Jiang, Huiqiang and Wu, Qianhui and Lin, Chin-Yew and Yang, Yuqing and Qiu, Lili},
  booktitle={Proceedings of the 2023 conference on empirical methods in natural language processing},
  pages={13358--13376},
  year={2023}
}

@article{arslan2024survey,
  title={A Survey on RAG with LLMs},
  author={Arslan, Muhammad and Ghanem, Hussam and Munawar, Saba and Cruz, Christophe},
  journal={Procedia computer science},
  volume={246},
  pages={3781--3790},
  year={2024},
  publisher={Elsevier}
}

@article{ge2023context,
  title={In-context autoencoder for context compression in a large language model},
  author={Ge, Tao and Hu, Jing and Wang, Lei and Wang, Xun and Chen, Si-Qing and Wei, Furu},
  journal={arXiv preprint arXiv:2307.06945},
  year={2023}
}

@inproceedings{chevalier2023adapting,
  title={Adapting language models to compress contexts},
  author={Chevalier, Alexis and Wettig, Alexander and Ajith, Anirudh and Chen, Danqi},
  booktitle={Proceedings of the 2023 Conference on Empirical Methods in Natural Language Processing},
  pages={3829--3846},
  year={2023}
}

@article{yu2024visrag,
  title={Visrag: Vision-based retrieval-augmented generation on multi-modality documents},
  author={Yu, Shi and Tang, Chaoyue and Xu, Bokai and Cui, Junbo and Ran, Junhao and Yan, Yukun and Liu, Zhenghao and Wang, Shuo and Han, Xu and Liu, Zhiyuan and others},
  journal={arXiv preprint arXiv:2410.10594},
  year={2024}
}

@inproceedings{karpukhin2020dense,
  title={Dense passage retrieval for open-domain question answering},
  author={Karpukhin, Vladimir and Oguz, Barlas and Min, Sewon and Lewis, Patrick and Wu, Ledell and Edunov, Sergey and Chen, Danqi and Yih, Wen-tau},
  booktitle={Proceedings of the 2020 conference on empirical methods in natural language processing (EMNLP)},
  pages={6769--6781},
  year={2020}
}

@article{zheng2026beyond,
  title={Beyond Imitation: Reinforcement Learning for Active Latent Planning},
  author={Zheng, Zhi and Lee, Wee Sun},
  journal={arXiv preprint arXiv:2601.21598},
  year={2026}
}

@book{robertson2009probabilistic,
  title={The probabilistic relevance framework: BM25 and beyond},
  author={Robertson, Stephen and Zaragoza, Hugo},
  volume={4},
  year={2009},
  publisher={Now Publishers Inc}
}

@article{izacard2023atlas,
  title={Atlas: Few-shot learning with retrieval augmented language models},
  author={Izacard, Gautier and Lewis, Patrick and Lomeli, Maria and Hosseini, Lucas and Petroni, Fabio and Schick, Timo and Dwivedi-Yu, Jane and Joulin, Armand and Riedel, Sebastian and Grave, Edouard},
  journal={Journal of Machine Learning Research},
  volume={24},
  number={251},
  pages={1--43},
  year={2023}
}

@inproceedings{guu2020retrieval,
  title={Retrieval augmented language model pre-training},
  author={Guu, Kelvin and Lee, Kenton and Tung, Zora and Pasupat, Panupong and Chang, Mingwei},
  booktitle={International conference on machine learning},
  pages={3929--3938},
  year={2020},
  organization={PMLR}
}

@article{liu2023visual,
  title={Visual instruction tuning},
  author={Liu, Haotian and Li, Chunyuan and Wu, Qingyang and Lee, Yong Jae},
  journal={Advances in neural information processing systems},
  volume={36},
  pages={34892--34916},
  year={2023}
}

@article{xu2025mem,
  title={A-mem: Agentic memory for llm agents},
  author={Xu, Wujiang and Liang, Zujie and Mei, Kai and Gao, Hang and Tan, Juntao and Zhang, Yongfeng},
  journal={arXiv preprint arXiv:2502.12110},
  year={2025}
}

@inproceedings{dong2024survey,
  title={A survey of llm-based agents: Theories, technologies, applications and suggestions},
  author={Dong, Xiaofei and Zhang, Xueqiang and Bu, Weixin and Zhang, Dan and Cao, Feng},
  booktitle={2024 3rd International Conference on Artificial Intelligence, Internet of Things and Cloud Computing Technology (AIoTC)},
  pages={407--413},
  year={2024},
  organization={IEEE}
}

@article{wu2025human,
  title={From human memory to ai memory: A survey on memory mechanisms in the era of llms},
  author={Wu, Yaxiong and Liang, Sheng and Zhang, Chen and Wang, Yichao and Zhang, Yongyue and Guo, Huifeng and Tang, Ruiming and Liu, Yong},
  journal={arXiv preprint arXiv:2504.15965},
  year={2025}
}

@article{jiang2026embedrl,
  title={Embed-RL: Reinforcement Learning for Reasoning-Driven Multimodal Embeddings},
  author={Jiang, Haonan and Wang, Yuji and Zhu, Yongjie and Lu, Xin and Qin, Wenyu and Wang, Meng and Wan, Pengfei and Tang, Yansong},
  journal={arXiv preprint arXiv:2602.13823},
  year={2026}
}

@article{lan2026umer1,
  title={UME-R1: Exploring Reasoning-Driven Generative Multimodal Embeddings},
  author={Lan, Zhibin and Niu, Liqiang and Meng, Fandong and Zhou, Jie and Su, Jinsong},
  journal={arXiv preprint arXiv:2511.00405},
  year={2026}
}

@article{he2026plume,
  title={PLUME: Latent Reasoning Based Universal Multimodal Embedding},
  author={He, Chenwei and Hao, Xiangzhao and Yang, Tianyu and Ma, Yuxiang and Jia, Yuheng and Wu, Lingxiang and Zhao, Chaoyang and Guo, Haiyun and Wang, Jinqiao},
  journal={arXiv preprint arXiv:2604.02073},
  year={2026}
}

@article{daull2023complex,
  title={Complex QA and language models hybrid architectures, Survey},
  author={Daull, Xavier and Bellot, Patrice and Bruno, Emmanuel and Martin, Vincent and Murisasco, Elisabeth},
  journal={arXiv preprint arXiv:2302.09051},
  year={2023}
}

@article{lin2025llm,
  title={Llm inference enhanced by external knowledge: A survey},
  author={Lin, Yu-Hsuan and Chen, Qian-Hui and Cheng, Yi-Jie and Zhang, Jia-Ren and Liu, Yi-Hung and Hsia, Liang-Yu and Chen, Yun-Nung},
  journal={arXiv preprint arXiv:2505.24377},
  year={2025}
}

@article{yue2025survey,
  title={A survey of large language model agents for question answering},
  author={Yue, Murong},
  journal={arXiv preprint arXiv:2503.19213},
  year={2025}
}

@article{lewis2020retrieval,
  title={Retrieval-augmented generation for knowledge-intensive nlp tasks},
  author={Lewis, Patrick and Perez, Ethan and Piktus, Aleksandra and Petroni, Fabio and Karpukhin, Vladimir and Goyal, Naman and K{\"u}ttler, Heinrich and Lewis, Mike and Yih, Wen-tau and Rockt{\"a}schel, Tim and others},
  journal={Advances in neural information processing systems},
  volume={33},
  pages={9459--9474},
  year={2020}
}

@article{park2025mobilerag,
  title={MobileRAG: A Fast, Memory-Efficient, and Energy-Efficient Method for On-Device RAG},
  author={Park, Taehwan and Lee, Geonho and Kim, Min-Soo},
  journal={arXiv preprint arXiv:2507.01079},
  year={2025}
}

@inproceedings{mutlu2025memory,
  title={Memory-centric computing: solving computing’s memory problem},
  author={Mutlu, Onur and Olgun, Ataberk and Y{\"u}ksel, {\.I}smail Emir},
  booktitle={2025 IEEE International Memory Workshop (IMW)},
  pages={1--4},
  year={2025},
  organization={IEEE}
}
}

% ============================================================

\newpage
% ============================================================
\appendix
% ============================================================

\thispagestyle{plain}
\section*{Appendix Contents}
\begin{enumerate}
    \item \hyperref[app:more_related]{\textbf{Related Work}} \dotfill \pageref{app:more_related}
    \begin{enumerate}
        \item \hyperref[app:related_rag]{RAG and Embedding Retrieval} \dotfill \pageref{app:related_rag}
        \item \hyperref[app:related_context]{Evidence Compression for Generation} \dotfill \pageref{app:related_context}
        \item \hyperref[app:related_interaction]{Retrieval-Compression Interaction} \dotfill \pageref{app:related_interaction}
        \item \hyperref[app:related_latent]{Latent-Space Modeling for LLMs} \dotfill \pageref{app:related_latent}
    \end{enumerate}
    \item \hyperref[app:implementation_details]{\textbf{Implementation Details}} \dotfill \pageref{app:implementation_details}
    \begin{enumerate}
        \item \hyperref[app:prompt_details]{Prompt Templates} \dotfill \pageref{app:prompt_details}
        \item \hyperref[app:detailed_pipeline]{Detailed Pipeline} \dotfill \pageref{app:detailed_pipeline}
        \item \hyperref[app:hyperparameter_settings]{Hyperparameter Settings} \dotfill \pageref{app:hyperparameter_settings}
        \item \hyperref[app:time_complexity]{Time Complexity Analysis} \dotfill \pageref{app:time_complexity}
        \item \hyperref[app:space_complexity]{Space Complexity Analysis} \dotfill \pageref{app:space_complexity}
    \end{enumerate}
    \item \hyperref[app:more_exp]{\textbf{Additional Experiments and Discussion}} \dotfill \pageref{app:more_exp}
    \begin{enumerate}
        \item \hyperref[app:ablation]{Token Count Ablation} \dotfill \pageref{app:ablation}
        \item \hyperref[app:text_super_ood]{Generalization Ability on More Domains} \dotfill \pageref{app:text_super_ood}
        \item \hyperref[app:ablation_recon]{Ablation on Core Settings} \dotfill \pageref{app:ablation_recon}
        \item \hyperref[app:ablation_qwen15]{Ablation on Stronger Text Compressors} \dotfill \pageref{app:ablation_qwen15}
        \item \hyperref[app:generator_transfer]{Direct Transfer to Similar Generator} \dotfill \pageref{app:generator_transfer}
        \item \hyperref[app:gemma_mm_results]{Multimodal Results with Gemma} \dotfill \pageref{app:gemma_mm_results}
        \item \hyperref[app:embedding_model]{Latent Tokens as Retrievers} \dotfill \pageref{app:embedding_model}
    \end{enumerate}
    \item \hyperref[app:examples]{\textbf{Case Study}} \dotfill \pageref{app:examples}
    \begin{enumerate}
        \item \hyperref[app:case_recon]{Reconstruction Quality of Latent Tokens} \dotfill \pageref{app:case_recon}
        \item \hyperref[app:examples_text]{Text-only Case Studies} \dotfill \pageref{app:examples_text}
        \item \hyperref[app:examples_mm]{More Multimodal QA Case Studies} \dotfill \pageref{app:examples_mm}
    \end{enumerate}
    \item \hyperref[app:assets]{\textbf{Baselines, Datasets, and Licenses}} \dotfill \pageref{app:assets}
\end{enumerate}

\newpage

\section{Related Work}
\label{app:more_related}

\begin{table}[h]
\centering
\small
\setlength{\tabcolsep}{3.5pt}
\renewcommand{\arraystretch}{1}
\caption{Capability comparison with representative related work. \textbf{Multimodal (Text \& Image)} discusses whether the method can adapt to multimodal documents or evidence. \textbf{Efficient generation} means whether this method aims at reducing the generator token consumption. \textbf{Unified generation \& retrieval} means that the same compressed representation is used both for retrieval and as the object consumed by the generator (\ding{51} making it unnecessary to work separately on compression and retrieval, xRAG ($\triangle$) using the same retrieved embeddings as generator input at inference, but relies on a separately trained bridge rather than unified retrieval-generation training). \textbf{No need to fine-tune generator LLM/VLM} means the memory item can be directly fed into the generator without requiring it to comprehend. (\ding{51} making it easier to deploy and avoid catastrophic forgetting)}
\label{tab:related_compare}
\resizebox{\textwidth}{!}{
\begin{tabular}{>{\raggedright\arraybackslash}m{5.0cm}>{\centering\arraybackslash}m{2cm}>{\centering\arraybackslash}m{2cm}>{\centering\arraybackslash}m{2.5cm}>{\centering\arraybackslash}m{3cm}}
\toprule
\textbf{Representative work} & \shortstack{\textbf{Multimodal}\\\textbf{(Text \& Image)}} & \shortstack{\textbf{Efficient}\\\textbf{generation}} & \shortstack{\textbf{Unified generation}\\\textbf{\& retrieval}} & \shortstack{\textbf{No need to fine-tune}\\\textbf{generator LLM/VLM}} \\
\midrule
\multicolumn{5}{l}{\textbf{(1) RAG baselines:} \textit{Raw-evidence RAG}} \\
BM25 / Dense RAG / Qwen3-Embedding \citep{robertson2009probabilistic,karpukhin2020dense,zhang2025qwen3} & \ding{55} & \ding{55} & \ding{55} & \ding{51} \\
VisRAG / Qwen3-VL-Embedding / Nemo Retriever \citep{yu2024visrag,li2026qwen3,xu2025llama} & \ding{51} & \ding{55} & \ding{55} & \ding{51} \\
\midrule
\multicolumn{5}{l}{\textbf{(2) Compression-based baselines:} \textit{Evidence compression for generation}} \\
LLMLingua / ICAE \citep{jiang2023llmlingua,ge2023context} & \ding{55} & \ding{51} & \ding{55} & \ding{51} \\
AutoCompressor \citep{chevalier2023adapting} & \ding{55} & \ding{51} & \ding{55} & \ding{55} \\
LCC / PISCO \citep{li2026latent,louis2025pisco} & \ding{55} & \ding{51} & \ding{55} & \ding{55} \\
\midrule
\multicolumn{5}{l}{\textbf{(1) \& (2) Interaction baselines:} \textit{Embedding-based retrieval then latent-context generation}} \\
xRAG \citep{cheng2024xrag} & \ding{55} & \ding{51} & $\triangle$ & \ding{51} \\
CLaRa \citep{he2025clara} & \ding{55} & \ding{51} & \ding{51} & \ding{55} \\
\midrule
\multicolumn{5}{l}{\textbf{Ours: }\textit{Compiling Latent Memory for retrieval \& generation}} \\
\textbf{Latent Memory} & \textbf{\ding{51}} & \textbf{\ding{51}} & \textbf{\ding{51}} & \textbf{\ding{51}} \\
\bottomrule
\end{tabular}
}
\end{table}

In this section, I will discuss works that are related to the proposed Latent Memory paradigm, as well as a broad range of works that have a similar latent-representation idea. Table \ref{tab:related_compare} presents a comprehensive collection of existing related works.

\subsection{RAG and Embedding Retrieval}
\label{app:related_rag}

\textbf{BM25-based sparse RAG.}
Retrieval-augmented generation selects a small subset of evidence and then lets a pretrained generator answer from the retrieved content \citep{lewis2020retrieval,izacard2023atlas}. BM25 is the classical sparse retrieval baseline, relying on lexical matching and term statistics rather than learned semantic representations \citep{robertson2009probabilistic}. It is simple and directly compatible with frozen LLMs, but the generator still consumes raw retrieved text, so the retrieval step does not reduce the per-evidence token cost of generation.

\textbf{Text-only Dense retrieval and embedding models.}
Dense retrieval learns a neural text-embedding space for query-evidence matching, as in DPR-style retrieval \citep{karpukhin2020dense}. As the most recent embedding model for dense retrieval, Qwen3-Embedding strengthens this retrieval front end with a modern text embedding and reranking model \citep{zhang2025qwen3}. These methods improve semantic retrieval over textual evidence, but the representation is still used primarily as a retrieval key; after retrieval, the generator receives raw text rather than the embedding itself. Moreover, these works cannot generalize well to multimodal settings, especially on images with details that cannot be accurately captioned.

\textbf{VisRAG and multimodal embedding retrieval.}
Multimodal RAG extends retrieval from text-only corpora to mixed text-image evidence. VisRAG is representative of raw-evidence multimodal RAG: it retrieves relevant visual documents but still hands visual inputs to the generator \citep{yu2024visrag}. As pretrained embedding model for multimodal RAG, Qwen3-VL-Embedding provides a unified vision-language embedding and reranking framework for text-image retrieval \citep{li2026qwen3}. Nemo Retriever is another strong text-image retrieval model used as a retrieval front end \citep{xu2025llama}.

Recent reasoning-aware multimodal embedding methods further refine what the retrieval vector should encode: Embed-RL optimizes multimodal embeddings through reinforcement learning signals \citep{jiang2026embedrl}, UME-R1 explores reasoning-driven generative multimodal embeddings \citep{lan2026umer1}, and PLUME uses latent reasoning for universal multimodal embedding \citep{he2026plume}. As summarized in Table~\ref{tab:related_compare}, these methods support multimodal retrieval and can be used with pretrained VLM generators, but they remain raw-evidence RAG: retrieved images or text are still passed to the generator in their native input form. Thus, a retrieved image can still expand into many visual tokens, keeping generation expensive even when retrieval is accurate.

\subsection{Evidence Compression for Generation}
\label{app:related_context}

\textbf{Discrete prompt compression.}
Evidence-compression methods reduce the context consumed by the generator after the evidence has already been chosen. LLMLingua prunes or rewrites discrete prompt tokens, preserving compatibility with pretrained LLMs while shortening the input \citep{jiang2023llmlingua}. This improves generator-side efficiency, but it does not build a retrievable memory: the shortened context is produced for the current prompt rather than stored as a corpus item for future retrieval.

\textbf{Latent or compact context compression.}
ICAE compresses context into learned memory-like latent states for language modeling \citep{ge2023context}. AutoCompressor trains models to summarize previous context into compact summary vectors that can condition later generation \citep{chevalier2023adapting}. LCC studies latent context compression for reducing long-context generation cost \citep{li2026latent}. PISCO similarly targets compact soft-context representations for efficient generation \citep{louis2025pisco}. These methods correspond to the second block of Table~\ref{tab:related_compare}: they focus on efficient generation, but they generally do not retrieve over the compressed evidence representation. So, because there is no retrieval system, these methods can exceed the generation context window when the irrelevant context is very long, resulting in meaningless output. Several methods also require the generator side to be trained or adapted to understand the compressed tokens, so compression and retrieval remain separate problems.

\subsection{Retrieval-Compression Interaction}
\label{app:related_interaction}

These are also works that combine the idea of retrieval and compression.

\textbf{xRAG.}
xRAG first presents the idea of using a unified representation for generation \& retrieval. However, as we mark it with $\triangle$ for unified generation and retrieval in Table~\ref{tab:related_compare}. xRAG relies on training a bridge based on pre-trained retrieval embeddings; noting that there are representations being important for generation but meaningless for retrieval (e.g., some details), the two-stage process makes it unable to find the optimal and unified generation \& retrieval representation.

\textbf{CLaRa.}
The most recent work, CLaRa, is closer to unified latent-context retrieval and generation because its latent context can be retrieved and then used for generation \citep{he2025clara}. Both CLaRa and xRAG have the same limitation: it is text-only. Moreover, CLaRa requires a large amount of pre-training effort, and the latent representation in CLaRa is processed to fine-tune LLM/VLM Latent Memory, which might lead to catastrophic forgetting.

\textbf{Latent Memory \textit{vesus} xRAG \& CLaRa.}

Compared to the two methods mentioned above, the Latent Memory paradigm \textbf{extends the text-only problem to multimodal scenarios}. The success of Latent Memory in multimodal scenarios \textbf{not only improves generation token efficiency but also reduces storage pressure}, which cannot be achieved by pure text-only Latent Memory. We provide a more detailed description in Appendix~\ref{app:space_complexity}, showing that saving text-based data does not significantly increase storage pressure compared to Latent Memory; on the contrary, high-dimensional vectors are often more difficult to store. However, for images, Latent Memory in LLaVA scenarios (4096-dimensional bf16 latent token) is more efficient than an uncompressed RGB image once the image is larger than roughly 53 $\times$ 53 pixels.

\subsection{Latent-Space Modeling for LLMs}
\label{app:related_latent}

\textbf{Latent reasoning and hidden-state computation.}
Latent Memory is inspired by a broader direction of latent-space representation \citep{zhu2025survey,chen2025reasoning,liu2025efficient,yu2026latent}. Previous work shows that continuous hidden states can carry useful information in a wide collection of applications, including latent chain-of-thought math reasoning \citep{hao2024training,shen2025codi,wei2025sim,zheng2026beyond,wang2025monet,yu2025vismem}, soft-thinking reasoning \citep{zhang2025soft,butt2025soft,wu2025llms,zheng2025soft}, and agentic communication in multi-turn / multi-agent \citep{fu2026latentmem,zhang2025memgen}. It is worth noting that although some works also use a similar name, Memory \citep{fu2026latentmem,zhang2025memgen,yu2025vismem}, we are the first to use latent tokens to save storage and generation token consumption in multimodal contextual QA. Besides, some studies seek a unified latent representation space for multimodal understanding and generation \citep{chen2025vl,sun2026vla,nam2026causal}. These studies motivate the idea that hidden-state vectors have the ability to carry unified representations for contextual memory.

\newpage

\section{Implementation Details}
\label{app:implementation_details}

\subsection{Prompt Templates}
\label{app:prompt_details}

Teacher and student use the same system-user-assistant scaffold. Across text-only and LLaVA pipelines, the instruction is: \textit{``You are a helpful assistant. Answer the question concisely (a few words or a short phrase) based on the provided context.''} The only difference is the evidence representation: the teacher sees raw evidence, while the student receives one latent token for each retrieved evidence item in the same retrieved order.

\begin{tcolorbox}[colback=gray!5,colframe=black!20,title=Text-only teacher prompt,boxrule=0.4pt]
\small
\texttt{[system] You are a helpful assistant. Answer the question concisely}\\
\texttt{(a few words or a short phrase) based on the provided context.}\\
\texttt{[user]}\\
\texttt{Context 1: <retrieved fact 1>}\\
\texttt{Context 2: <retrieved fact 2>}\\
\texttt{\dots}\\
\texttt{Context k: <retrieved fact k>}\\
\texttt{Question: <question>}\\
\texttt{[assistant]}\\
\texttt{Answer:}
\end{tcolorbox}

\begin{tcolorbox}[colback=gray!5,colframe=black!20,title=Text-only student prompt,boxrule=0.4pt]
\small
\texttt{[system] You are a helpful assistant. Answer the question concisely}\\
\texttt{(a few words or a short phrase) based on the provided context.}\\
\texttt{[user]}\\
\texttt{Latent context 1: [LATENT]}\\
\texttt{Latent context 2: [LATENT]}\\
\texttt{\dots}\\
\texttt{Latent context k: [LATENT]}\\
\texttt{Question: <question>}\\
\texttt{[assistant]}\\
\texttt{Answer:}
\end{tcolorbox}

\begin{tcolorbox}[colback=gray!5,colframe=black!20,title=Multimodal teacher prompt,boxrule=0.4pt]
\small
\texttt{[system] You are a helpful assistant. Answer the question concisely}\\
\texttt{(a few words or a short phrase) based on the provided context.}\\
\texttt{[user]}\\
\texttt{Context 1: <retrieved text fact 1>}\\
\texttt{Context 2: <image>}\\
\texttt{Title: <retrieved image title/caption 2>}\\
\texttt{\dots}\\
\texttt{Question: <question>}\\
\texttt{[assistant]}\\
\texttt{Answer:}
\end{tcolorbox}

\begin{tcolorbox}[colback=gray!5,colframe=black!20,title=Multimodal student prompt,boxrule=0.4pt]
\small
\texttt{[system] You are a helpful assistant. Answer the question concisely}\\
\texttt{(a few words or a short phrase) based on the provided context.}\\
\texttt{[user]}\\
\texttt{Latent context 1: [LATENT]}\\
\texttt{Latent context 2: [LATENT]}\\
\texttt{\dots}\\
\texttt{Latent context k: [LATENT]}\\
\texttt{Question: <question>}\\
\texttt{[assistant]}\\
\texttt{Answer:}
\end{tcolorbox}

\subsection{Detailed Pipeline}
\label{app:detailed_pipeline}

\paragraph{Retrieval projection MLP.}
The retrieval head maps the compressor hidden state into a 512-dimensional retrieval space. In all reported text-only and multimodal runs, it is implemented as
\[
\mathrm{retrieval\_proj}(\boldsymbol{z})=\ell_2\left(\mathrm{LayerNorm}(\mathrm{Linear}(\boldsymbol{z}))\right),
\]
where the linear layer maps $d_\theta \rightarrow d_r$ and $d_r=512$ in all reported runs. The $\ell_2$ normalization is applied before FAISS storage and query-time similarity search, so inner product retrieval is equivalent to cosine similarity.

\paragraph{Generator projection MLP.}
The generator-side projector converts a retrieved Latent Memory vector into the frozen generator hidden dimension. It is a two-layer MLP with LayerNorm:
\[
\mathrm{cross\_proj}(\boldsymbol{z})=\mathrm{LayerNorm}(W_2\,\mathrm{GELU}(W_1 \boldsymbol{z})).
\]
For the text-only setting, it maps the Llama-3.2-1B hidden size $d_\theta=2048$ to the Meta-Llama-3-8B generator hidden size $d_\phi=4096$, with intermediate dimension $(2048+4096)/2=3072$. For multimodal LLaVA, it maps the LLaVA-7B compressor hidden size $d_\theta=4096$ to the LLaVA-13B generator hidden size $d_\phi=5120$, with intermediate dimension $(4096+5120)/2=4608$.

\paragraph{Decoder and image-reconstruction MLPs.}
For textual reconstruction, the projected latent token is inserted into the decoding prompt from Appendix~\ref{app:prompt_details}, and the decoder $\pi$ is trained to autoregressively recover the original text evidence or image caption. The decoder projector maps compressor hidden states into the lightweight decoder hidden space with \texttt{Linear}$(d_\theta,d_\pi)\rightarrow\texttt{LayerNorm}$, where $d_\pi$ denotes the hidden dimension of the decoder $\pi$. For image evidence, we do not reconstruct raw pixels. Instead, the image embedding reconstruction head predicts the frozen CLIP CLS hidden state. This MLP is
\[
\mathrm{img\_embed\_decode\_proj}(\boldsymbol{z})=W_2\,\mathrm{LayerNorm}(\mathrm{GELU}(W_1 \boldsymbol{z})),
\]
where $W_1$ maps $d_\theta$ to $(d_\theta+d_v)/2$ and $W_2$ maps to $d_v=1024$ for the CLIP target used in our runs. For LLaVA-7B, this midpoint is $(4096+1024)/2=2560$.

\paragraph{Online question answering.}
At inference time, the query is encoded with the query adapter and projected into the same normalized retrieval space. FAISS returns the top-$k$ memory entries. Their full latent vectors are ordered by retrieval score, projected with \texttt{cross\_proj}, and inserted into the frozen generator through \texttt{inputs\_embeds}. Ordinary prompt tokens are embedded with the generator embedding layer; projected latent memories are spliced between the prompt prefix and suffix. No generator weights are updated during training or evaluation.

\paragraph{Trainable components.}
The text model uses separate LoRA adapters for \texttt{compress}, \texttt{query}, \texttt{decode}, and \texttt{query\_decode}. The multimodal model additionally uses \texttt{image\_decode}. Encoder-side adapters target \texttt{q\_proj}, \texttt{k\_proj}, \texttt{v\_proj}, and \texttt{o\_proj}. Decoder-side adapters also include \texttt{gate\_proj} and \texttt{up\_proj}. The trainable non-LoRA components are the \texttt{[MEM]} token, retrieval projector, generator projector, decoder projector, and image embedding reconstruction projector.

\paragraph{Training-Loss Distillation.} The teacher is the frozen generator prompted with raw positive evidence, while the student uses the same frozen generator prompted with projected latent memories. For text-side distillation, the student latent context is randomly augmented with $0$--$3$ sampled hard-negative latent memories during training, while the teacher prompt remains positive-only. This augmentation exposes the student generator to small retrieval noise without changing the teacher target distribution; the KL loss is still computed on the first 16 generated answer tokens.

\newpage
\subsection{Hyperparameter Settings}
\label{app:hyperparameter_settings}

\begin{table}[h]
\centering
\small
\caption{Main hyperparameters and LoRA configuration used in the reported runs.}
\label{tab:hyperparams}
\resizebox{\textwidth}{!}{
\begin{tabular}{p{4.2cm}p{5.2cm}p{5.2cm}}
\toprule
\textbf{Component} & \textbf{Text-only} & \textbf{Multimodal} \\
\midrule
Backbone / generator & Llama-3.2-1B-Instruct compressor $\rightarrow$ frozen Meta-Llama-3-8B-Instruct generator & LLaVA-1.5-7B $\rightarrow$ frozen LLaVA-1.5-13B \\
Compression / latent dimension $d_\theta$ & 2048 for Llama-3.2-1B; one latent token per evidence item in the main setting & 4096 for LLaVA-1.5-7B; one latent token per text or image evidence item \\
Retrieval dimension $d_r$ & 512-dimensional normalized retrieval vector for FAISS inner-product search & 512-dimensional normalized retrieval vector shared by text and image evidence \\
Generation dimension $d_\phi$ & 4096 for the frozen Meta-Llama-3-8B generator; \texttt{cross\_proj}: $2048\rightarrow3072\rightarrow4096$ & 5120 for frozen LLaVA-1.5-13B with \texttt{cross\_proj}: $4096\rightarrow4608\rightarrow5120$ \\
Text reconstruction decoder & Llama-3.2-1B-Instruct & Llama-3.2-1B-Instruct for reported WebQA runs \\
Decoder / reconstruction dimension $d_\pi$ & 2048 decoder hidden size; \texttt{decode\_proj}: $2048\rightarrow2048$ & 2048 LLaMA decoder hidden size for reported WebQA runs; LLaVA \texttt{decode\_proj}: $4096\rightarrow2048$; image embedding target $d_v{=}1024$ \\
Encoder-side LoRA targets & \texttt{q\_proj}, \texttt{k\_proj}, \texttt{v\_proj}, \texttt{o\_proj} & \texttt{q\_proj}, \texttt{k\_proj}, \texttt{v\_proj}, \texttt{o\_proj} \\
Decoder-side LoRA targets & \texttt{q\_proj}, \texttt{k\_proj}, \texttt{v\_proj}, \texttt{o\_proj}, \texttt{gate\_proj}, \texttt{up\_proj} & \texttt{q\_proj}, \texttt{k\_proj}, \texttt{v\_proj}, \texttt{o\_proj}, \texttt{gate\_proj}, \texttt{up\_proj} \\
LoRA rank / alpha / dropout & $r{=}64$, $\alpha{=}128$, 0.05 & $r{=}64$, $\alpha{=}128$, 0.05 \\
\midrule
Optimizer and LR & AdamW, peak LR $1{\times}10^{-4}$ & AdamW, peak LR $1{\times}10^{-4}$ \\
Training batching & batch size 8, gradient accumulation 4 & batch size 8, gradient accumulation 2 \\
Training epochs & 3 epochs (about 20 hours) & 2 epochs (about 30 hours) \\
\midrule
Text loss weights & $\lambda_{\text{recon}}{=}0.5$, $\lambda_{\text{contrast}}{=}0.2$, $\lambda_{\text{distill}}{=}1.0$ & $\lambda_{\text{recon}}{=}0.5$, $\lambda_{\text{contrast}}{=}0.2$, $\lambda_{\text{distill}}{=}1.0$ \\
Image loss weights & -- & $\lambda_{\text{image-contrast}}{=}0.2$, $\lambda_{\text{image-distill}}{=}2.0$ \\
Embedding reconstruction & -- & $\lambda_{\text{embed-recon}}{=}5.0$ \\
Hard negatives per sample & up to 8 text negatives & up to 4 text + 4 image negatives \\
Distillation supervision & first 16 answer tokens & first 16 answer tokens \\
\bottomrule
\end{tabular}}
\end{table}

The hyperparameter of loss is selected basically based on the scale.

Additional implementation choices are shared across settings. During training, retrieval scores are computed in memory without building a FAISS index. During offline evaluation, we build an FAISS inner-product index over normalized retrieval vectors. Negative reconstruction is enabled for text evidence and image captions.

Query reconstruction is disabled in the main runs. The token count considers from ``Context 1:`` to the end, so introducing one more latent token may consume more than one generator token budget (usually 7 to 10 based on LLM/VLM tokenizer). For WebQA token accounting, raw retrieved images are counted after the generator's native visual frontend; Latent Memory bypasses this raw visual expansion and inserts one projected latent token per retrieved item.

\newpage
\subsection{Time Complexity Analysis}
\label{app:time_complexity}

This section compares the dominant time complexity under two deployment settings: the evidence index or memory bank has already been compiled, or it must be compiled for the current corpus. Let $\mathcal{C}=\{\boldsymbol{x}_i\}_{i=1}^{N}$ be the evidence pool and let $\bar{L}$ be the average number of raw text or visual tokens per evidence item. We denote the retrieval depth by $k$, the number of latent tokens per evidence item by $T$, the question length by $|\boldsymbol{Q}|$, the frozen generator hidden size by $d_\phi$, the retrieval embedding model hidden size by $d_e$, and the Latent Memory compressor hidden size by $d_\theta$. The table omits retrieval/search overhead and keeps only the dominant evidence-encoding and generator-prefill terms.

\begin{table}[H]
\centering
\small
\caption{Dominant time complexity comparison under precompiled and non-precompiled deployment. Retrieval/search overhead is omitted; the key term is the frozen generator prefill length.}
\label{tab:time_complexity}
\setlength{\tabcolsep}{10pt}
\resizebox{\textwidth}{!}{
\begin{tabular}{lll}
\toprule[0.5mm]
\textbf{Method} & \textbf{With precompiled index / memory} & \textbf{Without precompilation} \\
\midrule
Full Context &
$\mathcal{O}\!\left((|\boldsymbol{Q}|+N\bar{L})^2 d_\phi\right)$ &
$\mathcal{O}\!\left((|\boldsymbol{Q}|+N\bar{L})^2 d_\phi\right)$ \\
\midrule
Raw-evidence RAG &
$\mathcal{O}\!\left((|\boldsymbol{Q}|+k\bar{L})^2 d_\phi\right)$ &
$\mathcal{O}\!\left(N\bar{L}^2 d_e+(|\boldsymbol{Q}|+k\bar{L})^2 d_\phi\right)$ \\
\midrule
Latent Memory &
$\mathcal{O}\!\left((|\boldsymbol{Q}|+kT)^2 d_\phi\right)$ &
$\mathcal{O}\!\left(N\bar{L}^2 d_\theta+(|\boldsymbol{Q}|+kT)^2 d_\phi\right)$ \\
\bottomrule[0.5mm]
\end{tabular}}
\end{table}

These formulas follow directly from Transformer prefill complexity: forwarding a sequence of length $S$ through the frozen generator costs $\mathcal{O}(S^2d_\phi)$. Full Context uses $S=|\boldsymbol{Q}|+N\bar{L}$, RAG uses $S=|\boldsymbol{Q}|+k\bar{L}$ after selecting $k$ raw evidence items, and Latent Memory uses $S=|\boldsymbol{Q}|+kT$ because each retrieved evidence item is represented by $T$ latent tokens.
\begin{itemize}[leftmargin=*]
    \item Compared with Full Context, Latent Memory avoids sending the whole evidence pool to the generator. Full Context has a generator-prefill term $\mathcal{O}((|\boldsymbol{Q}|+N\bar{L})^2d_\phi)$, which grows quadratically with the number of evidence items $N$ when $\bar{L}$ is fixed. With precompilation, Latent Memory instead uses $\mathcal{O}((|\boldsymbol{Q}|+kT)^2d_\phi)$, which no longer depends on $N$ in the generator. Without precompilation, Latent Memory adds the evidence compilation term $\mathcal{O}(N\bar{L}^2d_\theta)$, which is linear in $N$ and can be highly parallelized across evidence items. As shown in the figure below, we plot the relationship of time consumption with the number of evidence items. Latent Memory shows linear complexity and leads to significantly less time complexity, even with compilation.

    \begin{figure}[H]
        \centering
        \includegraphics[width=\linewidth]{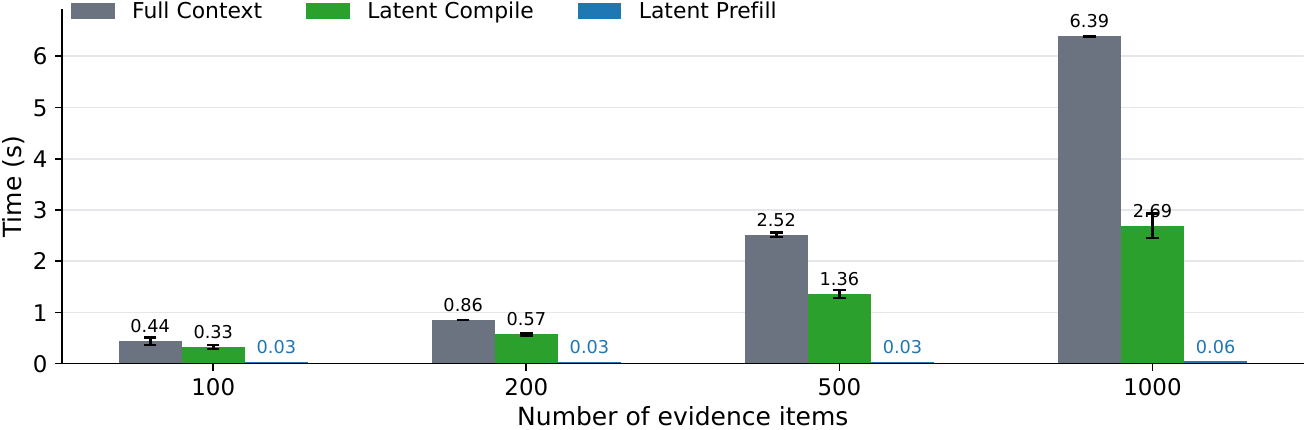}
        \caption{Time analysis on HotpotQA, full-context vs Latent Memory. Compile means pre-compile, and prefill means the prefill process before answer-generation. We run 30 times for variance.}
    \end{figure}

    \item Compared with raw-evidence RAG, Latent Memory has a similar non-precompiled setup structure: RAG embeds all evidence items, while Latent Memory compiles them into latent memories. The key difference is again the generator input after retrieval. RAG still sends $k$ raw evidence items to the generator, giving $|\boldsymbol{Q}|+k\bar{L}$, whereas Latent Memory sends $kT$ latent tokens, giving $|\boldsymbol{Q}|+kT$. Since $T\ll\bar{L}$ and usually $k\ll N$, Latent Memory has the smallest online generator-prefill complexity among the three methods. Moreover, fewer tokens lead to less probability of out-of-memory, and a \textbf{square-root less space complexity}.
\end{itemize}

\newpage
\subsection{Storage Complexity Analysis}
\label{app:space_complexity}

Table~\ref{tab:storage_complexity_webqa} reports an average per-evidence storage comparison on WebQA. The text side uses the official WebQA text snippets in \texttt{WebQA\_train\_val.json}. The image side uses the official WebQA image files after extraction, which occupy about 75GB in total (the compressed package is about 50GB), averaged over the 350{,}777 unique image IDs appearing in the official train/validation annotations. We count only the stored evidence representation and do not include small metadata fields.

\begin{table}[h]
\centering
\small
\caption{Average per-item storage on WebQA. Raw text and raw image storage follow the official WebQA annotations and extracted image files. Latent text uses the text-only 1-token memory size ($2048$ bf16 values), while latent image uses the LLaVA 1-token memory size ($4096$ bf16 values).}
\label{tab:storage_complexity_webqa}
\setlength{\tabcolsep}{6pt}
\resizebox{0.92\textwidth}{!}{
\begin{tabular}{lccc}
\toprule
\textbf{Evidence type} & \textbf{Stored representation} & \textbf{Avg. storage / item} & \textbf{Storage effect} \\
\midrule
Text & Official raw text snippet & 0.23 KB & Reference \\
Text & Latent Memory token & 4.00 KB & \textcolor{red}{17.6$\times$ larger} \\
\midrule
Image & Official extracted image file & 209 KB & Reference \\
Image & Latent Memory token & 8.00 KB & \textcolor{ourgreen}{26.1$\times$ smaller} \\
\bottomrule
\end{tabular}}
\end{table}

This comparison clarifies where the storage advantage comes from. On the pure language side, a short raw snippet is often smaller than a high-dimensional latent vector, so text-only Latent Memory should mainly be understood as reducing generator-side token usage rather than persistent corpus storage. On the image side, however, the raw evidence item is much larger; replacing each extracted WebQA image with a single LLaVA latent token gives a clear storage reduction while also avoiding raw visual-token expansion during generation.

The same conclusion also follows from a simple analytical threshold. A one-token LLaVA Latent Memory stores $4096$ bf16 values, i.e.,
\begin{equation}
    S_{\mathrm{latent/image}} = 2\times4096 = 8192\ \mathrm{bytes}.
\end{equation}
For an uncompressed RGB image with height $H$ and width $W$, raw storage is
\begin{equation}
    S_{\mathrm{image}} = 3HW\ \mathrm{bytes}.
\end{equation}
Thus, Latent Memory is smaller whenever $3HW>8192$, or $H=W>\sqrt{8192/3}\approx52.3$ for square images. In other words, a 4096-dimensional bf16 latent token is more storage efficient than an uncompressed RGB image once the image is larger than roughly $53\times53$ pixels. If a separate 512-dimensional fp32 retrieval key is also stored, the per-item latent storage becomes $8192+4\times512=10240$ bytes, shifting the square-image threshold only to about $59\times59$ pixels.

The online activation footprint follows the same pattern. Raw multimodal RAG must instantiate visual embeddings for each retrieved image inside the VLM, giving an evidence activation size of $\mathcal{O}(k\bar{L}d_\phi)$. Latent Memory instead instantiates only projected latent tokens, giving $\mathcal{O}(kT d_\phi)$. Thus the online generation path reduces both token consumption and activation memory when $\bar{L}\gg T$, which is typical for image evidence after visual-token expansion.

\newpage
\section{Additional Experiments and Discussion}
\label{app:more_exp}

The experiments in the main text answer the three questions as follows:

\begin{itemize}[leftmargin=*]
    \item 1-token Latent Memory can show a token-performance trade-off in multimodal settings, replacing the raw retrieved context.
    \item Replacing raw evidence with 8-token Latent Memory results in a better trade-off, surpassing the best RAG baseline on out-of-domain performance.
\end{itemize}

The appendix then studies where this behavior comes from and how broadly it holds. We organize the additional evidence around the following questions:

\begin{itemize}[leftmargin=*]
    \item \textbf{RQ-A1: How much token each latent evidence are needed for text-only and multimodal QA?} Appendix~\ref{app:ablation} varies the number of latent tokens per evidence item.
    \item \textbf{RQ-A2: Does the Latent Memory perform well across evidence domains, compressors, and generators?} Appendices~\ref{app:text_super_ood}, \ref{app:generator_transfer}, \ref{app:ablation_qwen15}, and \ref{app:gemma_mm_results} test broader text and image QA benchmarks, compressors, and generator settings.
    \item \textbf{RQ-A3: Which training objectives support the unified representation space?} Appendix~\ref{app:ablation_recon} ablates reconstruction, negative reconstruction, query reconstruction, and augmentation settings.
    \item \textbf{RQ-A5: How effective are compression and retrieval, respectively?} Appendix~\ref{app:embedding_model} excludes R@k improvements, how effective is compression? We considered a variant using latent tokens to provide the generator with raw evidence after retrieval to observe compression effectiveness.
\end{itemize}

\newpage
\subsection{Token Count Ablation}
\label{app:ablation}

This section varies the number of latent tokens allocated to each evidence item. The default model uses one token; we also evaluate 2-, 4-, and 8-token variants while keeping the evidence unit and retrieval granularity unchanged. For an evidence item $\boldsymbol{x}_i$, the multi-token variant emits
\begin{equation}
    \boldsymbol{Z}_i = [\boldsymbol{z}_{i,1}, \boldsymbol{z}_{i,2}, \ldots, \boldsymbol{z}_{i,T}], \qquad \boldsymbol{z}_{i,t} \in \mathbb{R}^{d_\theta}.
\end{equation}
Retrieval still uses one key per evidence item by pooling these tokens before the retrieval projection:
\begin{equation}
    \bar{\boldsymbol{z}}_i = \frac{1}{T}\sum_{t=1}^{T} \boldsymbol{z}_{i,t}, \qquad
    \widetilde{\boldsymbol{z}}_i = \mathrm{LayerNorm}(\boldsymbol{W}_r \bar{\boldsymbol{z}}_i).
\end{equation}
Thus, the ablation tests whether extra latent capacity improves generation enough to justify the larger token budget.

\begin{figure}[H]
\centering
\includegraphics[width=0.8\linewidth]{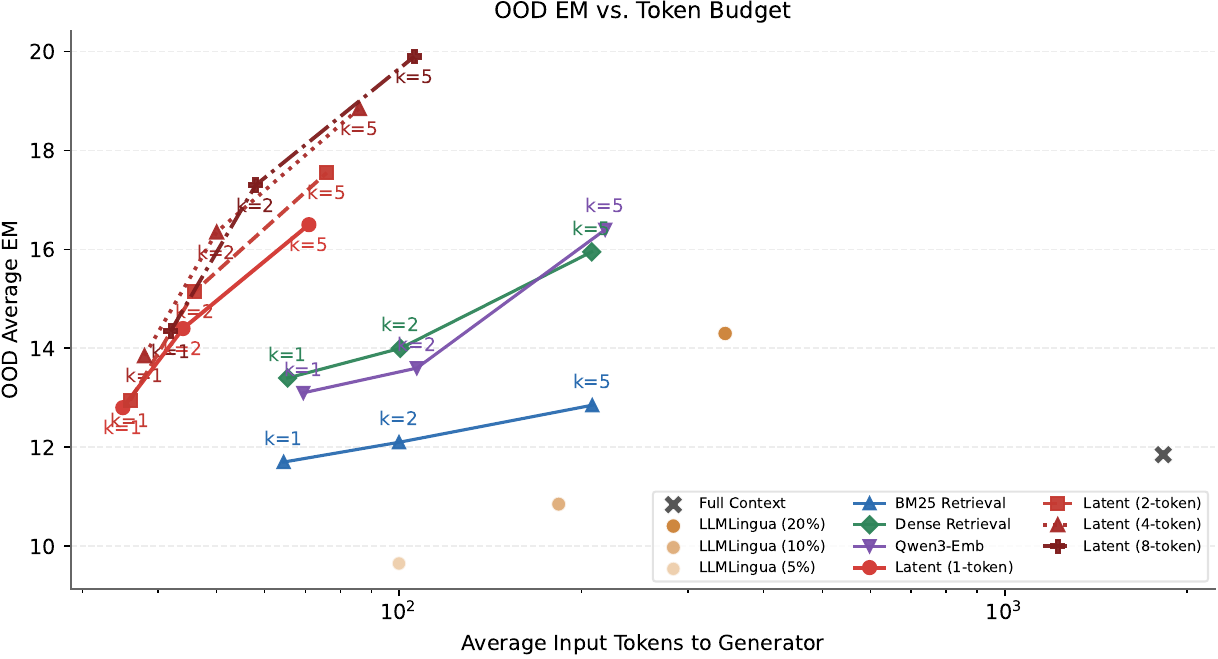}
\caption{OOD average text-QA EM as a function of average generator tokens. The average is computed over 2WikiMultihopQA and MuSiQue.}
\label{fig:ablation_em_token}
\end{figure}

\begin{figure}[H]
\centering
\includegraphics[width=0.8\linewidth]{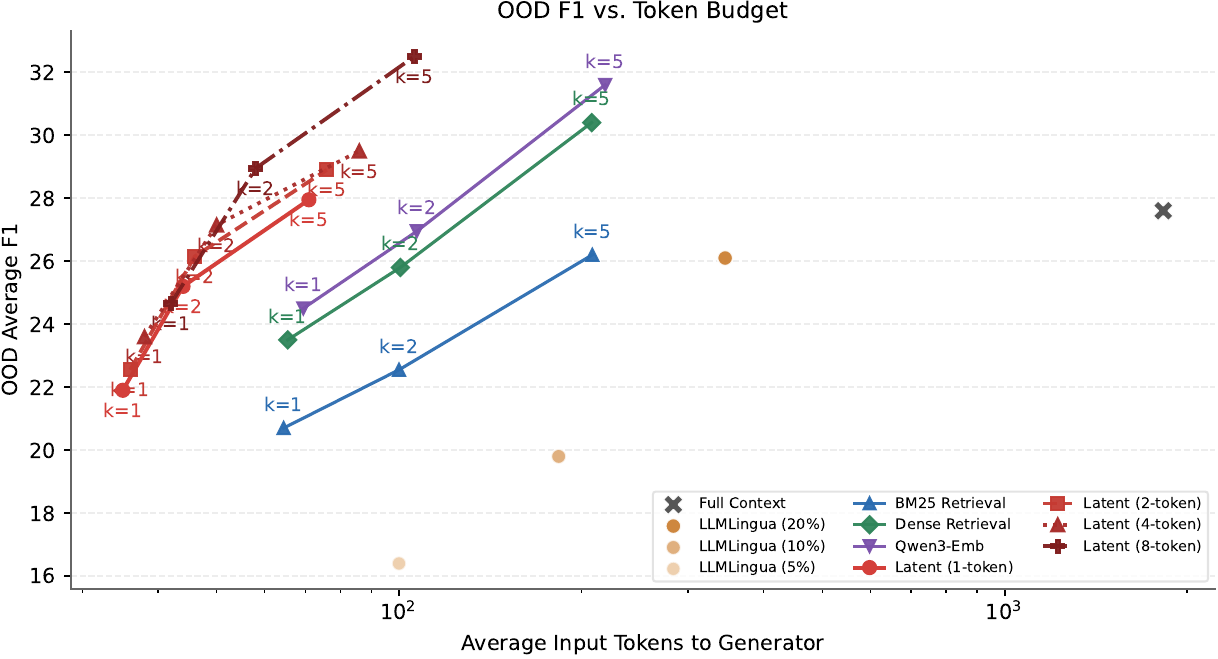}
\caption{OOD average text-QA F1 as a function of average generator tokens. The average is computed over 2WikiMultihopQA and MuSiQue.}
\label{fig:ablation_f1_token}
\end{figure}

\begin{figure}[H]
\centering
\includegraphics[width=0.8\linewidth]{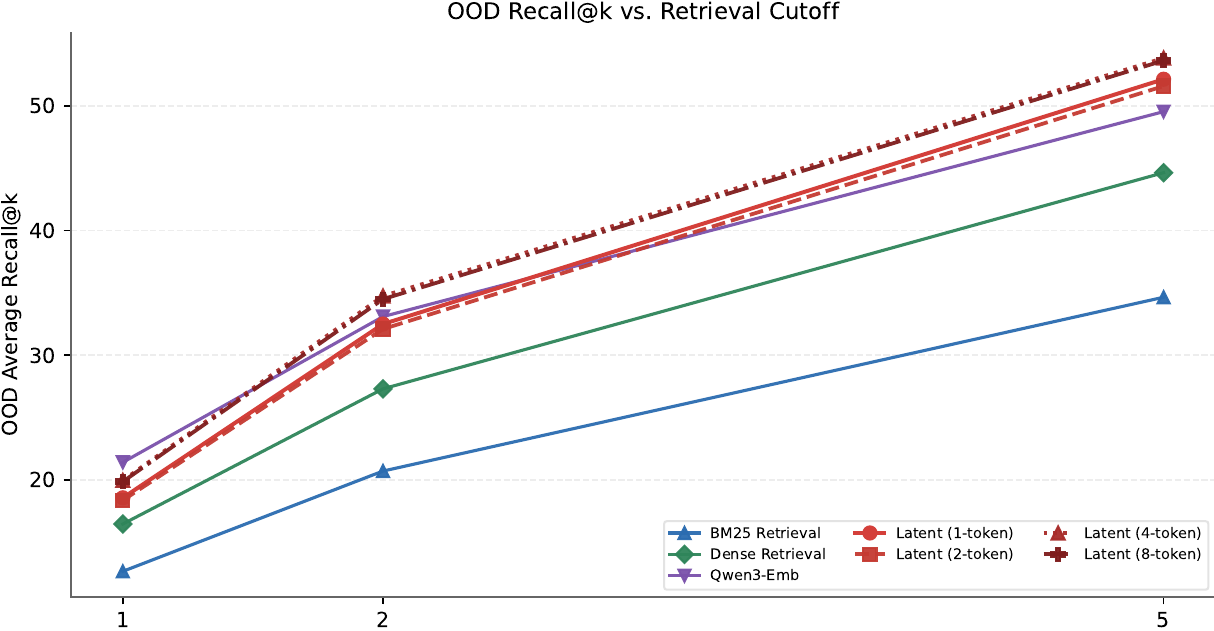}
\caption{OOD average Recall@$k$ as a function of $k$. The average is computed over 2WikiMultihopQA and MuSiQue.}
\label{fig:ablation_rk}
\end{figure}

Table~\ref{tab:ablation} reports the text-QA results, and Figures~\ref{fig:ablation_em_token}--\ref{fig:ablation_rk} summarize the accuracy, token, and recall trends.

\paragraph{Analysis.}
Three conclusions are consistent across text-only setting (Table~\ref{tab:ablation} and Figures~\ref{fig:ablation_em_token}--\ref{fig:ablation_rk}).
\begin{enumerate}
\item Increasing the latent-token budget consistently improves EM and F1, and the larger-token variants surpass the main retrieval baselines in answer quality.
\item By contrast, increasing the latent-token budget does not lead to a comparable improvement in Recall@$k$ at fixed $k$.
\item These quality gains are achieved while preserving the overall trade-off: larger latent budgets do increase generator tokens, but they remain substantially more efficient than Full Context and still compare favorably with the raw retrieval baselines.
\end{enumerate}

% The multimodal ablation in Table~\ref{tab:mm_token_ablation} shows a more nuanced pattern. On the WebQA text split, increasing the latent budget gives a small generation-side gain, especially at larger $k$, which is consistent with the text-only results: more latent tokens provide more capacity for carrying factual content into the generator. On the WebQA image split, however, using more latent tokens does not improve the final answer quality; at $k{=}5$, image F1 drops from 69.4 to 66.1 when moving from 1-token to 8-token Latent Memory. This suggests that the image-grounded setting is not primarily bottlenecked by the number of latent tokens per retrieved item. Instead, the key benefit comes from retrieving the right visual evidence and presenting it in a compact form that the VLM can use. Additional latent capacity is therefore most useful for text evidence, while the image side receives little benefit from simply allocating more latent tokens.

\begin{table}[H]
\centering
\small
\caption{Token-count ablation on text QA over HotpotQA, 2WikiMultihopQA, and MuSiQue. The Average columns report the out-of-domain average over 2WikiMultihopQA and MuSiQue. The 1-token setting is the main Latent Memory configuration; 2/4/8-token variants allocate more latent tokens per evidence item. \#Tok reports the task-relevant generator budget.}
\label{tab:ablation}
\resizebox{\textwidth}{!}{
\begin{tabular}{lcccc |cccc cccc| cccc}
\toprule
& \multicolumn{4}{c}{\textbf{In-Domain}} & \multicolumn{12}{c}{\textbf{Out-of-Domain}} \\
\cmidrule(lr){2-5}\cmidrule(lr){6-17}
Dataset & \multicolumn{4}{c}{\textbf{HotpotQA}} & \multicolumn{4}{c}{\textbf{2WikiMultihopQA}} & \multicolumn{4}{c}{\textbf{MuSiQue}} & \multicolumn{4}{c}{\textbf{Average}} \\
\cmidrule(lr){2-5} \cmidrule(lr){6-9} \cmidrule(lr){10-13} \cmidrule(lr){14-17}
Method & EM & F1 & R@$k$ & \#Tok & EM & F1 & R@$k$ & \#Tok & EM & F1 & R@$k$ & \#Tok & EM & F1 & R@$k$ & \#Tok \\
\midrule
Full Context & 42.0 & 57.5 & -- & 1462 & 17.7 & 39.7 & -- & 1074 & 6.0 & 15.5 & -- & 2580 & 11.9 & 27.6 & -- & 1827 \\
LLMLingua (20\%) & 31.8 & 44.8 & -- & 283 & 17.0 & 30.4 & -- & 199 & 11.6 & 21.8 & -- & 492 & 14.3 & 26.1 & -- & 346 \\
LLMLingua (10\%) & 25.0 & 36.1 & -- & 154 & 14.8 & 24.7 & -- & 108 & 6.9 & 14.9 & -- & 259 & 10.9 & 19.8 & -- & 184 \\
LLMLingua (5\%) & 20.9 & 30.2 & -- & 87 & 15.0 & 22.2 & -- & 63 & 4.3 & 10.6 & -- & 137 & 9.7 & 16.4 & -- & 100 \\
\midrule
\multicolumn{17}{l}{\textit{Retrieval Augmented Generation Methods}} \\
BM25 Retrieval ($k{=}1$) & 32.5 & 44.7 & 30.2 & 68 & 17.2 & 27.2 & 18.3 & 60 & 6.2 & 14.2 & 7.0 & 69 & 11.7 & 20.7 & 12.7 & 65 \\
BM25 Retrieval ($k{=}2$) & 36.8 & 49.9 & 45.5 & 106 & 16.2 & 28.4 & 29.2 & 94 & 8.0 & 16.7 & 12.2 & 106 & 12.1 & 22.6 & 20.7 & 100 \\
BM25 Retrieval ($k{=}5$) & 41.3 & 55.3 & 65.5 & 224 & 16.2 & 32.8 & 46.9 & 196 & 9.5 & 19.6 & 22.4 & 221 & 12.9 & 26.2 & 34.7 & 209 \\
Dense Retrieval ($k{=}1$) & 29.4 & 41.0 & 27.9 & 67 & 19.0 & 29.9 & 23.3 & 61 & 7.8 & 17.1 & 9.6 & 70 & 13.4 & 23.5 & 16.5 & 66 \\
Dense Retrieval ($k{=}2$) & 32.6 & 45.3 & 42.2 & 104 & 18.1 & 31.7 & 37.6 & 95 & 9.9 & 19.9 & 17.0 & 106 & 14.0 & 25.8 & 27.3 & 101 \\
Dense Retrieval ($k{=}5$) & 37.0 & 50.7 & 62.4 & 214 & 19.1 & 37.4 & 58.0 & 198 & 12.8 & 23.4 & 31.3 & 218 & 16.0 & 30.4 & 44.7 & 208 \\
Qwen3-Emb-0.6B ($k{=}1$) & 30.9 & 43.6 & 33.1 & 70 & 18.1 & 30.9 & 32.5 & 66 & 8.1 & 18.1 & 10.3 & 73 & 13.1 & 24.5 & 21.4 & 70 \\
Qwen3-Emb-0.6B ($k{=}2$) & 35.6 & 49.0 & 50.0 & 109 & 17.4 & 33.5 & 47.7 & 102 & 9.8 & 20.4 & 18.5 & 112 & 13.6 & 27.0 & 33.1 & 107 \\
Qwen3-Emb-0.6B ($k{=}5$) & 40.0 & 54.5 & 70.1 & 224 & 19.1 & 38.6 & 64.3 & 208 & 13.7 & 24.6 & 34.8 & 230 & 16.4 & 31.6 & 49.6 & 219 \\
\midrule
\multicolumn{17}{l}{\textit{Retrieval Augmented Generation for Latent Tokens in the Latent Memory (Ours)}} \\
1-token ($k{=}1$) & 27.4 & 39.4 & 34.6 & 36 & 19.8 & 29.2 & 28.4 & 33 & 5.8 & 14.6 & 8.7 & 37 & 12.8 & 21.9 & 18.6 & 35 \\
1-token ($k{=}2$) & 31.6 & 45.2 & 62.6 & 45 & 21.5 & 33.5 & 49.5 & 42 & 7.3 & 16.9 & 15.5 & 46 & 14.4 & 25.2 & 32.5 & 44 \\
1-token ($k{=}5$) & 34.8 & 48.9 & 87.1 & 72 & 24.3 & 36.7 & 74.2 & 69 & 8.7 & 19.2 & 30.1 & 73 & 16.5 & 28.0 & 52.2 & 71 \\
2-token ($k{=}1$) & 27.3 & 39.5 & 34.6 & 37 & 20.2 & 29.9 & 28.0 & 34 & 5.7 & 15.2 & 8.7 & 38 & 13.0 & 22.6 & 18.4 & 36 \\
2-token ($k{=}2$) & 31.4 & 45.2 & 62.6 & 47 & 23.0 & 34.7 & 49.0 & 44 & 7.3 & 17.6 & 15.2 & 48 & 15.2 & 26.2 & 32.1 & 46 \\
2-token ($k{=}5$) & 35.2 & 49.3 & 87.1 & 77 & 26.6 & 38.0 & 73.8 & 74 & 8.5 & 19.8 & 29.4 & 78 & 17.6 & 28.9 & 51.6 & 76 \\
4-token ($k{=}1$) & 28.6 & 40.4 & 35.2 & 39 & 21.9 & 31.4 & 30.9 & 36 & 5.8 & 15.8 & 9.0 & 40 & 13.9 & 23.6 & 20.0 & 38 \\
4-token ($k{=}2$) & 33.0 & 46.9 & 63.2 & 51 & 25.1 & 36.2 & 53.5 & 48 & 7.6 & 18.1 & 16.0 & 52 & 16.4 & 27.2 & 34.8 & 50 \\
4-token ($k{=}5$) & 35.8 & 49.9 & 88.0 & 87 & 28.9 & 39.1 & 76.7 & 84 & 8.8 & 19.9 & 31.0 & 88 & 18.9 & 29.5 & 53.9 & 86 \\
8-token ($k{=}1$) & 31.6 & 44.7 & 35.1 & 43 & 21.2 & 31.6 & 30.5 & 40 & 7.5 & 17.7 & 9.2 & 44 & 14.4 & 24.7 & 19.9 & 42 \\
8-token ($k{=}2$) & 38.9 & 53.2 & 63.4 & 59 & 24.0 & 36.4 & 52.8 & 56 & 10.6 & 21.5 & 16.2 & 60 & 17.3 & 29.0 & 34.5 & 58 \\
8-token ($k{=}5$) & 42.8 & 58.5 & 88.0 & 107 & 26.8 & 40.5 & 76.2 & 104 & 13.0 & 24.5 & 31.1 & 108 & 19.9 & 32.5 & 53.7 & 106 \\
\bottomrule
\end{tabular}}
\end{table}

\newpage

\newpage

\newpage
\subsection{Generalization Ability on More Domains}
\label{app:text_super_ood}
The experiments in the main body of this paper demonstrate that Latent Memory remains effective on the OOD dataset, but these results can be considered generalizations to similar corpora. In this section, we consider whether this effectiveness can be extended to more different situations and datasets that are very different from HotpotQA. Table~\ref{tab:text_super_ood_results} reports generalization to four additional text QA benchmarks: NQ (open-domain factoid), TriviaQA (trivia-style factoid), Qasper (scientific document QA), and WICE (claim verification). Unlike the three datasets in the main text, these datasets do not have a retrieval label design, so we did not calculate R@k. No fine-tuning is performed on these datasets; the same checkpoint trained on HotpotQA is evaluated directly. This tests whether the compressed latent representations generalize across domains and task types.

\begin{table}[h]
\centering
\small
\caption{Generalization ability on more text domains under the same frozen \textit{Meta-Llama-3-8B-Instruct} generator. No fine-tuning is performed on these datasets. \#Tok reports the task-relevant generator budget. Unlike the three datasets in the main text, these datasets do not have a retrieval label design, so we did not calculate R@k.}
\label{tab:text_super_ood_results}
\resizebox{\textwidth}{!}{
\begin{tabular}{lccc ccc ccc ccc|ccc}
\toprule
& \multicolumn{3}{c}{\textbf{NQ}} & \multicolumn{3}{c}{\textbf{TriviaQA}} & \multicolumn{3}{c}{\textbf{Qasper}} & \multicolumn{3}{c}{\textbf{WiCE}} & \multicolumn{3}{c}{\textbf{Average}} \\
\cmidrule(lr){2-4}\cmidrule(lr){5-7}\cmidrule(lr){8-10}\cmidrule(lr){11-13}\cmidrule(lr){14-16}
Method & EM & F1 & \#Tok & EM & F1 & \#Tok & EM & F1 & \#Tok & EM & F1 & \#Tok & EM & F1 & \#Tok \\
\midrule
Full Context & 0.0 & 1.1 & 23588 & 18.9 & 24.7 & 10828 & 2.5 & 16.5 & 4701 & 51.1 & 51.1 & 2311 & 18.1 & 23.3 & 10357 \\
\midrule
\multicolumn{16}{l}{\textit{Raw-evidence retrieval}} \\
BM25 ($k{=}1$) & 25.3 & 36.6 & 54 & 71.0 & 77.7 & 66 & 6.0 & 17.1 & 50 & 43.6 & 43.6 & 135 & 36.5 & 43.8 & 77 \\
BM25 ($k{=}2$) & 26.7 & 38.6 & 90 & 70.7 & 77.8 & 106 & 8.0 & 20.5 & 82 & 44.1 & 44.1 & 210 & 37.4 & 45.3 & 122 \\
BM25 ($k{=}5$) & 28.1 & 40.4 & 197 & 71.0 & 78.3 & 226 & 9.5 & 23.6 & 187 & 51.7 & 51.7 & 417 & 40.1 & 48.5 & 257 \\
Dense ($k{=}1$) & 28.2 & 40.5 & 58 & 70.2 & 77.1 & 57 & 5.5 & 16.6 & 46 & 39.9 & 39.9 & 116 & 36.0 & 43.5 & 69 \\
Dense ($k{=}2$) & 30.6 & 43.5 & 96 & 71.2 & 78.3 & 85 & 8.5 & 21.1 & 76 & 43.6 & 43.6 & 174 & 38.5 & 46.6 & 108 \\
Dense ($k{=}5$) & \underline{33.1} & \underline{46.5} & 210 & 71.7 & 79.2 & 171 & \textbf{11.5} & \underline{26.8} & 168 & 46.4 & 46.4 & 351 & \underline{40.7} & \underline{49.7} & 225 \\
Qwen3-Emb-0.6B ($k{=}1$) & 30.8 & 43.0 & 60 & 71.9 & 79.0 & 70 & 8.5 & 19.8 & 47 & 42.2 & 42.2 & 126 & 38.4 & 46.0 & 76 \\
Qwen3-Emb-0.6B ($k{=}2$) & 32.0 & 45.1 & 99 & \underline{72.1} & \underline{79.8} & 110 & \underline{10.0} & 26.6 & 80 & 42.5 & 42.5 & 201 & 39.1 & 48.5 & 122 \\
Qwen3-Emb-0.6B ($k{=}5$) & \textbf{34.5} & \textbf{48.2} & 218 & \textbf{72.8} & \textbf{80.9} & 221 & 8.5 & \textbf{30.2} & 171 & 47.5 & 47.5 & 389 & \textbf{40.8} & \textbf{51.7} & 250 \\
\midrule
\multicolumn{16}{l}{\textit{Latent Memory}} \\
Latent Memory ($k{=}1$) & 22.2 & 33.7 & 26 & 66.5 & 72.9 & 34 & 3.0 & 10.9 & 25 & 45.5 & 45.5 & 59 & 34.3 & 40.8 & 36 \\
Latent Memory ($k{=}2$) & 22.7 & 34.6 & 35 & 66.4 & 73.0 & 44 & 4.5 & 12.2 & 34 & 52.5 & 52.5 & 68 & 36.5 & 43.1 & 45 \\
Latent Memory ($k{=}5$) & 23.4 & 35.0 & 62 & 65.1 & 71.7 & 70 & 6.0 & 12.3 & 61 & 55.9 & 55.9 & 94 & 37.6 & 43.7 & 72 \\
8-token Latent Memory ($k{=}1$) & 25.5 & 37.2 & 33 & 68.9 & 75.4 & 42 & 5.0 & 14.5 & 32 & 52.8 & 52.8 & 66 & 38.1 & 45.0 & 43 \\
8-token Latent Memory ($k{=}2$) & 24.9 & 36.8 & 49 & 68.1 & 74.8 & 58 & 5.0 & 14.3 & 48 & \underline{57.0} & \underline{57.0} & 82 & 38.7 & 45.7 & 59 \\
8-token Latent Memory ($k{=}5$) & 26.7 & 38.5 & 97 & 67.7 & 74.4 & 106 & 5.0 & 13.2 & 96 & \textbf{60.3} & \textbf{60.3} & 129 & 39.9 & 46.6 & 107 \\
\bottomrule
\end{tabular}}
\end{table}

On these datasets, the RAG method based on Qwen-3-Embedding performs best, but we note that the 1-token and 8-token versions of Latent Memory exhibit good trade-offs. The 8-token Latent Memory method achieves similar performance to BM25 while requiring 2.5 times fewer tokens.

\paragraph{Multimodal generalization.}\label{app:multimodal_super_ood}
Table~\ref{tab:slideqa_results} further evaluates multimodal generalization on a 20-image dataset SlideQA (Consider getting detailed answers from 1-2 correct slides.) with the same frozen LLaVA-1.5-13B generator. Latent Memory improves retrieval coverage and competitive EM value with far fewer generator tokens, but Nemo remains stronger on EM/F1.

\begin{table}[H]
\centering
\small
\caption{Generalization ability on the multimodal SlideQA domain under a frozen \textit{LLaVA-1.5-13B} generator. \textbf{Bold} marks the best result in each metric column, and underlining marks the second best one. \#Tok reports the generator-side input budget.}
\label{tab:slideqa_results}
\setlength{\tabcolsep}{6pt}
\begin{tabular}{lcccc}
\toprule
\multicolumn{5}{c}{\textbf{Dataset: SlideQA \quad Generation VLM (fixed): LLaVA-1.5-13B}} \\
\midrule
\textbf{Method} & \textbf{EM} & \textbf{F1} & \textbf{R@$k$} & \textbf{\#Tok} \\
\midrule
\multicolumn{5}{l}{\textit{Full-context baseline}} \\
Full Context & 0.0 & 0.0 & -- & 11871 \\
\midrule
\multicolumn{5}{l}{\textit{Raw-evidence multimodal retrieval}} \\
Nemo ($k{=}1$) & \textbf{17.1} & \textbf{26.2} & 7.2 & 621 \\
Nemo ($k{=}2$) & 8.3 & \underline{17.7} & 14.1 & 1212 \\
Nemo ($k{=}5$) & 4.7 & 13.2 & 30.5 & 2887 \\
\midrule
\multicolumn{5}{l}{\textit{Latent Memory}} \\
Latent Memory ($k{=}1$) & \underline{11.1} & 15.7 & 21.5 & 37 \\
Latent Memory ($k{=}2$) & 7.0 & 12.4 & \underline{33.6} & 47 \\
Latent Memory ($k{=}5$) & 5.5 & 10.3 & \textbf{56.0} & 77 \\
\bottomrule
\end{tabular}
\end{table}

\newpage
\subsection{Core Training Ablation}
\label{app:ablation_recon}

Table~\ref{tab:core_settings} ablates the reconstruction and distillation-negative designs used to train the text-only 1B compressor. The retrieval and generation settings are fixed; only the training targets or distillation context augmentation change. The variants are:
\begin{itemize}[leftmargin=*]
    \item \textbf{Ours-1B}: Full model that reconstructs both positive evidence and sampled negative evidence, with query reconstruction disabled.
    \item \textbf{w/o Constructing}: Evidence reconstruction is removed entirely, so neither positive nor negative evidence is reconstructed.
    \item \textbf{w/o Constructing Negative Examples}: Only positive evidence is reconstructed; sampled negative evidence is not used as a reconstruction target.
    \item \textbf{w/ Query Reconstruction}: Adds query reconstruction to the default evidence-reconstruction setup.
    \item \textbf{w/ Distillation Negative Augmentation}: During distillation, the student latent context is augmented with 0-3 randomly sampled irrelevant negative memories, while the teacher still sees only positive.
\end{itemize}

\begin{table}[H]
\centering
\small
\caption{Generator-side reconstruction ablation on text QA. All variants use the same frozen \textit{Meta-Llama-3-8B-Instruct} generator and matched latent-token budget.}
\label{tab:core_settings}
\setlength{\tabcolsep}{5pt}
\resizebox{\textwidth}{!}{
\begin{tabular}{lccc ccc ccc| ccc}
\toprule
& \multicolumn{3}{c}{\textbf{HotpotQA}} & \multicolumn{3}{c}{\textbf{2WikiMultihopQA}} & \multicolumn{3}{c}{\textbf{MuSiQue}} & \multicolumn{3}{c}{\textbf{Average}} \\
\cmidrule(lr){2-4}\cmidrule(lr){5-7}\cmidrule(lr){8-10}\cmidrule(lr){11-13}
\textbf{Variant} & \textbf{EM} & \textbf{F1} & \textbf{R@$k$} & \textbf{EM} & \textbf{F1} & \textbf{R@$k$} & \textbf{EM} & \textbf{F1} & \textbf{R@$k$} & \textbf{EM} & \textbf{F1} & \textbf{R@$k$} \\
\midrule
\multicolumn{13}{l}{\textit{Default model: reconstruct positive and negative evidence}} \\
Ours-1B ($k{=}1$) & 27.4 & 39.4 & 34.6 & 19.8 & 29.2 & 28.4 & 5.8 & 14.6 & 8.7 & 17.7 & 27.7 & 23.9 \\
Ours-1B ($k{=}2$) & 31.6 & 45.2 & 62.6 & 21.5 & 33.5 & 49.5 & 7.3 & 16.9 & 15.5 & 20.1 & 31.9 & 42.6 \\
Ours-1B ($k{=}5$) & \textbf{34.8} & \textbf{48.9} & \textbf{87.1} & \underline{24.3} & \textbf{36.7} & \textbf{74.2} & \underline{8.7} & \textbf{19.2} & \underline{30.1} & \textbf{22.6} & \textbf{34.9} & \textbf{63.8} \\
\midrule
\multicolumn{13}{l}{\textit{Reconstruction Loss}} \\
w/o Constructing ($k{=}1$) & 23.4 & 37.8 & 33.4 & 20.3 & 28.8 & 29.8 & 5.9 & 15.1 & 8.3 & 16.5 & 27.2 & 23.8 \\
w/o Constructing ($k{=}2$) & 27.6 & 42.7 & 62.7 & 24.2 & 32.4 & 48.4 & 7.9 & 17.8 & 14.4 & 19.9 & 31.0 & 41.8 \\
w/o Constructing ($k{=}5$) & 30.6 & 45.2 & 85.2 & 22.7 & 33.8 & 73.3 & \textbf{9.4} & \textbf{19.2} & \textbf{30.4} & 20.9 & 32.7 & \underline{62.9} \\
w/o Constructing Negative ($k{=}1$) & 24.3 & 39.3 & 34.1 & 19.0 & 30.1 & 28.0 & 5.7 & 14.8 & 7.6 & 16.3 & 28.1 & 23.2 \\
w/o Constructing Negative ($k{=}2$) & 29.1 & 44.9 & 58.0 & 19.8 & 33.9 & 47.0 & 6.3 & 17.2 & 13.8 & 18.4 & 32.0 & 39.6 \\
w/o Constructing Negative ($k{=}5$) & \underline{31.7} & \underline{46.2} & \underline{86.2} & 22.3 & \underline{36.1} & 72.2 & 8.3 & \underline{18.9} & 28.7 & 20.8 & \underline{33.8} & 62.3 \\
w/ Query Reconstruction ($k{=}1$) & 22.3 & 35.9 & 34.2 & 19.3 & 28.4 & 25.8 & 4.8 & 13.6 & 7.8 & 15.5 & 26.0 & 22.6 \\
w/ Query Reconstruction ($k{=}2$) & 26.5 & 41.7 & 60.7 & 22.6 & 31.8 & 45.4 & 6.3 & 15.4 & 13.7 & 18.5 & 29.6 & 39.9 \\
w/ Query Reconstruction ($k{=}5$) & 30.1 & 45.7 & 84.9 & \textbf{26.3} & 35.4 & 72.6 & 7.7 & 18.0 & 26.3 & \underline{21.3} & 33.0 & 61.3 \\
\midrule
\multicolumn{13}{l}{\textit{Distillation Loss}} \\
w/o Negative Augmentation ($k{=}1$) & 26.2 & 38.2 & 33.4 & 21.0 & 30.9 & 27.6 & 5.5 & 14.6 & 8.1 & 17.6 & 27.9 & 23.0 \\
w/o Negative Augmentation ($k{=}2$) & 31.3 & 44.4 & 59.0 & 22.8 & 34.8 & 48.4 & 6.6 & 16.2 & 13.8 & 20.2 & 31.8 & 40.4 \\
w/o Negative Augmentation ($k{=}5$) & 31.6 & 44.8 & 83.0 & \underline{24.3} & 35.8 & \underline{73.6} & 7.2 & 17.5 & 25.5 & 21.0 & 32.7 & 60.7 \\
\bottomrule
\end{tabular}}
\end{table}

\paragraph{Analysis.}
\begin{enumerate}[leftmargin=*]
\item \textbf{Ours-1B} is the strongest default overall, but the main takeaway is that the reconstruction objective stabilizes both retrieval and generation.
\item \textbf{Removing evidence reconstruction entirely} (\textit{w/o Constructing}) consistently hurts EM/F1 and also lowers Recall@$k$ on average. This joint degradation supports the view that the same latent token is serving retrieval and generation in a unified representation space.
\item \textbf{Only reconstructing positive evidence} (\textit{w/o Constructing Negative Examples}) is less stable and degrades retrieval more clearly. This supports the role of negative evidence as a geometric anchor for the latent memory space.
\item \textbf{Adding query reconstruction} (\textit{w/ Query Reconstruction}) is harmful in this setting. It forces the query adapter to preserve surface reconstruction information, which conflicts with its role as a retrieval query representation.
\item \textbf{Adding 0--3 irrelevant negative memories during distillation} (\textit{w/ Distillation Negative Augmentation}) does not recover the full model, especially when $k=5$.
\end{enumerate}

Across the datasets, removing reconstruction weakens both retrieval and generation rather than trading one for the other. Negative evidence is useful when it is reconstructed as part of the memory representation, but query reconstruction and direct irrelevant-memory augmentation perturb the intended roles of the query encoder and generator-side conditioning target.

\newpage
\subsection{Stronger Text Compressors}
\label{app:ablation_qwen15}

This section asks whether a larger or a different-family compressor improves one-token Latent Memory's quality. The generator (frozen Meta-Llama-3-8B-Instruct), token budget, and training recipe are held fixed; only the compression backbone changes. Table~\ref{tab:encoder_ablation} compares LLaMA-1B (the default), LLaMA-3B, and Qwen-1.5B, averaged over HotpotQA, 2WikiMultihopQA, and MuSiQue.

\begin{table}[H]
\centering
\small
\caption{Text-compressor ablation on text QA. The frozen generator and training recipe are fixed; only the compression backbone changes.}
\label{tab:encoder_ablation}
\renewcommand\arraystretch{1.05}
\setlength{\tabcolsep}{5pt}
\resizebox{\textwidth}{!}{
\begin{tabular}{lccc ccc ccc |ccc}
\toprule
& \multicolumn{3}{c}{\textbf{HotpotQA}} & \multicolumn{3}{c}{\textbf{2WikiMultihopQA}} & \multicolumn{3}{c}{\textbf{MuSiQue}} & \multicolumn{3}{c}{\textbf{Average}} \\
\cmidrule(lr){2-4}\cmidrule(lr){5-7}\cmidrule(lr){8-10}\cmidrule(lr){11-13}
\textbf{Encoder} & \textbf{EM} & \textbf{F1} & \textbf{R@$k$} & \textbf{EM} & \textbf{F1} & \textbf{R@$k$} & \textbf{EM} & \textbf{F1} & \textbf{R@$k$} & \textbf{EM} & \textbf{F1} & \textbf{R@$k$} \\
\midrule
\multicolumn{13}{l}{\textit{LLaMA-family compressors}} \\
LLaMA-1B ($k{=}1$) & 27.4 & 39.4 & 34.6 & 19.8 & 29.2 & 28.4 & 5.8 & 14.6 & 8.7 & 17.7 & 27.7 & 23.9 \\
LLaMA-1B ($k{=}2$) & 31.6 & 45.2 & 62.6 & 21.5 & 33.5 & 49.5 & 7.3 & 16.9 & 15.5 & 20.1 & 31.9 & 42.6 \\
LLaMA-1B ($k{=}5$) & \underline{34.8} & \underline{48.9} & \underline{87.1} & 24.3 & 36.7 & 74.2 & 8.7 & \underline{19.2} & \textbf{30.1} & 22.6 & 34.9 & 63.8 \\
\midrule
LLaMA-3B ($k{=}1$) & 29.6 & 42.5 & 35.7 & 20.1 & 30.5 & 29.1 & 6.3 & 15.8 & 9.1 & 18.7 & 29.6 & 24.6 \\
LLaMA-3B ($k{=}2$) & 34.4 & 48.9 & 64.2 & 23.0 & 35.9 & 51.9 & 8.9 & 19.4 & 16.0 & 22.1 & 34.7 & 44.0 \\
LLaMA-3B ($k{=}5$) & \textbf{35.4} & \textbf{49.7} & \textbf{88.6} & \underline{25.1} & \underline{37.3} & \textbf{77.6} & \textbf{9.9} & \textbf{19.9} & \textbf{30.1} & \textbf{23.5} & \textbf{35.6} & \textbf{65.4} \\
\midrule
\multicolumn{13}{l}{\textit{Qwen-family compressor}} \\
Qwen-1.5B ($k{=}1$) & 25.5 & 37.2 & 34.8 & 20.6 & 30.3 & 30.0 & 5.0 & 13.2 & 8.8 & 17.0 & 26.9 & 24.5 \\
Qwen-1.5B ($k{=}2$) & 29.2 & 41.8 & 62.1 & 24.0 & 35.4 & 52.1 & 6.0 & 14.6 & 15.5 & 19.8 & 30.6 & 43.2 \\
Qwen-1.5B ($k{=}5$) & 30.7 & 43.4 & 86.4 & \textbf{27.4} & \textbf{37.4} & \underline{76.5} & 6.2 & 16.3 & \textbf{30.1} & 21.4 & 32.3 & \underline{64.3} \\
\bottomrule
\end{tabular}}
\end{table}

\paragraph{Analysis.}
\begin{enumerate}
\item \textit{LLaMA-3B} is consistently the strongest encoder in this comparison, improving both EM/F1 over \textit{LLaMA-1B} and \textit{Qwen-1.5B} at all reported $k$ values.
\item The gap is more pronounced in EM/F1 than in Recall@$k$, which suggests that encoder choice influences downstream answer quality more strongly than retrieval coverage alone.
\item Since the token budget is matched across all three variants, the advantage of \textit{LLaMA-3B} is best interpreted as a representational benefit rather than an efficiency effect.
\end{enumerate}

The dataset-level breakdown further shows that the stronger compressor helps both in-domain and OOD evaluation. On HotpotQA, LLaMA-3B improves the default LLaMA-1B model from 34.8/48.9 EM/F1 to 35.4/49.7 at $k{=}5$, while maintaining similar high Recall@$k$. On 2WikiMultihopQA, the improvement is more visible: LLaMA-3B reaches 25.1 EM and 37.3 F1 at $k{=}5$, compared with 24.3/36.7 for LLaMA-1B, and also improves Recall@$k$ from 74.2 to 77.6. MuSiQue remains the most difficult OOD dataset; LLaMA-3B improves EM/F1 modestly, but Recall@$k$ is nearly unchanged at $k{=}5$, indicating that the encoder upgrade mainly improves the quality of the latent representation consumed by the generator rather than solving all retrieval coverage limitations. Qwen-1.5B obtains competitive Recall@$k$ on 2Wiki and MuSiQue, but its lower EM/F1 suggests that high retrieval scores do not necessarily imply better latent-conditioned generation.

\newpage
\subsection{Direct Transfer to Similar Generator}
\label{app:generator_transfer}

This experiment studies whether Latent Memory trained with the \textit{Meta-Llama-3-8B-Instruct} generator can be directly reused by another similar LLaMA generator. We keep the compressor, latent memory bank, retrieval procedure, and projection interface fixed, and replace only the frozen answer generator with \textit{LLaMA-3.1-8B-Instruct}. No additional compressor training or latent-token adaptation is performed. Table~\ref{tab:generator_ablation} reports the full text-QA results under this transferred generator.

\begin{table}[H]
\centering
\small
\caption{Direct generator transfer on text QA. Latent tokens are trained under \textit{Meta-Llama-3-8B-Instruct} and evaluated directly with a frozen \textit{LLaMA-3.1-8B-Instruct} generator. \#Tok reports the generator-side input budget.}
\label{tab:generator_ablation}
\setlength{\tabcolsep}{3.5pt}
\resizebox{\textwidth}{!}{
\begin{tabular}{lcccc cccc cccc |cccc}
\toprule
\multicolumn{17}{c}{\textbf{Generation LLM (fixed): LLaMA-3.1-8B-Instruct}} \\
\midrule
& \multicolumn{4}{c}{\textbf{HotpotQA}} & \multicolumn{4}{c}{\textbf{2WikiMultihopQA}} & \multicolumn{4}{c}{\textbf{MuSiQue}} & \multicolumn{4}{c}{\textbf{Average}} \\
\cmidrule(lr){2-5}\cmidrule(lr){6-9}\cmidrule(lr){10-13}\cmidrule(lr){14-17}
\textbf{Method} & \textbf{EM} & \textbf{F1} & \textbf{R@$k$} & \textbf{\#Tok} & \textbf{EM} & \textbf{F1} & \textbf{R@$k$} & \textbf{\#Tok} & \textbf{EM} & \textbf{F1} & \textbf{R@$k$} & \textbf{\#Tok} & \textbf{EM} & \textbf{F1} & \textbf{R@$k$} & \textbf{\#Tok} \\
\midrule
\multicolumn{17}{l}{\textit{Full-context reference}} \\
Full Context & \textbf{50.2} & \textbf{65.9} & -- & 1462 & \textbf{35.2} & \textbf{48.7} & -- & 1074 & \textbf{18.8} & \textbf{30.7} & -- & 2580 & \textbf{34.7} & \textbf{48.4} & -- & 1706 \\
\midrule
\multicolumn{17}{l}{\textit{Sparse raw-evidence RAG baseline}} \\
BM25 Retrieval ($k{=}1$) & 29.6 & 43.0 & 30.2 & 68 & 13.5 & 23.2 & 18.3 & 60 & 6.1 & 13.3 & 7.0 & 69 & 16.4 & 26.5 & 18.5 & 66 \\
BM25 Retrieval ($k{=}2$) & 34.6 & 48.4 & 45.5 & 106 & 16.7 & 25.4 & 29.2 & 94 & 7.8 & 15.2 & 12.2 & 106 & 19.7 & 29.7 & 29.0 & 102 \\
BM25 Retrieval ($k{=}5$) & 41.2 & 55.4 & 65.5 & 224 & 24.4 & 32.0 & 46.9 & 196 & 10.6 & 19.2 & 22.4 & 221 & 25.4 & 35.5 & 44.9 & 214 \\
\midrule
\multicolumn{17}{l}{\textit{Dense raw-evidence RAG baselines}} \\
Dense Retrieval ($k{=}1$) & 26.5 & 39.3 & 27.9 & 67 & 15.0 & 25.5 & 23.3 & 61 & 6.6 & 15.1 & 9.6 & 70 & 16.0 & 26.6 & 20.3 & 66 \\
Dense Retrieval ($k{=}2$) & 30.6 & 44.1 & 42.2 & 104 & 18.9 & 29.0 & 37.6 & 95 & 8.3 & 17.5 & 17.0 & 106 & 19.3 & 30.2 & 32.3 & 102 \\
Dense Retrieval ($k{=}5$) & 36.9 & 50.9 & 62.4 & 214 & 27.8 & 37.0 & 58.0 & 198 & 12.6 & 22.7 & 31.3 & 218 & 25.7 & 36.8 & 50.6 & 210 \\
Qwen3-Emb ($k{=}1$) & 30.9 & 43.6 & 33.1 & 70 & 18.1 & 30.9 & 32.5 & 66 & 8.1 & 18.2 & 10.3 & 73 & 19.0 & 30.9 & 25.3 & 70 \\
Qwen3-Emb ($k{=}2$) & 35.6 & 49.0 & 50.0 & 109 & 17.4 & 33.5 & 47.7 & 102 & 9.8 & 20.4 & 18.5 & 112 & 21.0 & 34.3 & 38.8 & 108 \\
Qwen3-Emb ($k{=}5$) & 40.0 & 54.5 & 70.1 & 224 & 19.1 & 38.6 & 64.3 & 208 & 13.7 & \underline{24.7} & \textbf{34.8} & 230 & 24.3 & 39.3 & 56.4 & 221 \\
\midrule
\multicolumn{17}{l}{\textit{Direct transfer of Latent Memory}} \\
1-token Latent Memory ($k{=}1$) & 23.7 & 36.8 & 34.6 & 36 & 21.4 & 29.0 & 28.3 & 33 & 5.1 & 13.3 & 8.7 & 37 & 16.7 & 26.4 & 23.9 & 36 \\
1-token Latent Memory ($k{=}2$) & 29.6 & 43.6 & 62.7 & 45 & 26.9 & 34.9 & 49.5 & 42 & 6.2 & 15.7 & 15.5 & 46 & 20.9 & 31.4 & 42.6 & 45 \\
1-token Latent Memory ($k{=}5$) & 34.8 & 48.3 & 87.0 & 72 & 33.0 & 39.9 & 74.2 & 69 & 8.1 & 18.4 & 30.1 & 73 & 25.3 & 35.5 & 63.8 & 72 \\
2-token Latent Memory ($k{=}1$) & 24.5 & 36.8 & 34.6 & 37 & 22.6 & 30.0 & 28.0 & 34 & 5.2 & 13.2 & 8.7 & 38 & 17.4 & 26.7 & 23.7 & 36 \\
2-token Latent Memory ($k{=}2$) & 29.9 & 43.6 & 62.5 & 47 & 27.3 & 34.9 & 49.1 & 44 & 7.4 & 16.9 & 15.3 & 48 & 21.5 & 31.8 & 42.3 & 46 \\
2-token Latent Memory ($k{=}5$) & 34.6 & 47.8 & 87.0 & 77 & 31.0 & 38.6 & 73.8 & 74 & 9.3 & 19.9 & 29.4 & 78 & 25.0 & 35.5 & 63.4 & 76 \\
4-token Latent Memory ($k{=}1$) & 25.5 & 38.3 & 35.2 & 39 & 23.4 & 30.8 & 30.9 & 36 & 5.4 & 14.0 & 9.2 & 40 & 18.1 & 27.7 & 25.1 & 38 \\
4-token Latent Memory ($k{=}2$) & 32.7 & 45.8 & 63.2 & 51 & 30.2 & 36.9 & 53.5 & 48 & 8.1 & 18.1 & 16.2 & 52 & 23.7 & 33.6 & 44.3 & 50 \\
4-token Latent Memory ($k{=}5$) & 36.4 & 49.4 & \textbf{88.0} & 87 & \underline{33.7} & 40.0 & \textbf{76.7} & 84 & 9.3 & 19.6 & \underline{31.0} & 88 & 26.5 & 36.3 & \textbf{65.2} & 86 \\
8-token Latent Memory ($k{=}1$) & 29.4 & 43.1 & 35.1 & 80 & 24.2 & 32.4 & 30.5 & 97 & 7.4 & 16.4 & 9.2 & 101 & 20.3 & 30.6 & 24.9 & 93 \\
8-token Latent Memory ($k{=}2$) & 38.2 & 52.7 & 63.4 & 96 & 29.5 & 38.1 & 52.7 & 113 & 11.1 & 21.1 & 16.3 & 117 & 26.2 & 37.3 & 44.1 & 109 \\
8-token Latent Memory ($k{=}5$) & \underline{44.2} & \underline{58.7} & \textbf{88.0} & 144 & 33.5 & \underline{42.0} & \underline{76.2} & 161 & \underline{14.3} & \underline{24.7} & \underline{31.1} & 165 & \underline{30.7} & \underline{41.8} & \underline{65.1} & 157 \\
\bottomrule
\end{tabular}}
\end{table}

\paragraph{Analysis.}
The transferred latent tokens remain usable with the new generator without re-training. The one-token variant preserves the same token-efficiency pattern as in the main experiments, while larger latent-token budgets provide a clear capacity gain. In particular, the 8-token variant improves the average EM/F1 to 30.7/41.8 at $k{=}5$, exceeding the strongest raw-evidence embedding baseline at the same $k$ while still using fewer generator tokens. This suggests that Latent Memory is not tightly bound to a single frozen LLaMA generator: once the latent tokens learn to act as compact evidence, another compatible LLaMA instruction model can consume them directly.

\newpage
\subsection{Multimodal Results with Gemma}
\label{app:gemma_mm_results}

Table~\ref{tab:gemma_mm_results} reports WebQA results when the generator is switched to a frozen \textit{Gemma-3-12B-Instruct}. This model-swap setting keeps the retrieval pool and evidence candidates fixed, and compares four groups: full-context prompting, text-only raw-evidence retrievers, multimodal raw-evidence retrievers, and direct generation from Latent Memory. We fine-tune \textit{Gemma-3-4B-PT} as the compressor and use \textit{LLaMA-3.2-1B-Instruct} as the reconstruction decoder.

\begin{table}[H]
\centering
\small
\caption{Multimodal QA (WebQA) with unified retrieval and generation using a frozen \textit{Gemma-3-12B-Instruct} generator.}
\label{tab:gemma_mm_results}
\resizebox{\textwidth}{!}{
\begin{tabular}{lcccc cccc |cccc}
\toprule
\multicolumn{13}{c}{\textbf{Generation VLM (fixed): Gemma-3-12B-Instruct}} \\
\midrule
Method & \multicolumn{4}{c}{\textbf{WebQA-Image}} & \multicolumn{4}{c}{\textbf{WebQA-Text}} & \multicolumn{4}{c}{\textbf{Avg}} \\
\cmidrule(lr){2-5}\cmidrule(lr){6-9}\cmidrule(lr){10-13}
 & F1 & Acc & R@$k$ & \#Tok & F1 & Acc & R@$k$ & \#Tok & F1 & Acc & R@$k$ & \#Tok \\
\midrule
\multicolumn{13}{l}{\textit{Full-context reference}} \\
Full Context & \textbf{13.8} & \underline{18.0} & -- & 5856 & \textbf{54.9} & \textbf{54.9} & -- & 4337 & \textbf{34.3} & \textbf{36.4} & -- & 5097 \\
\midrule
\multicolumn{13}{l}{\textit{Raw-evidence text retrieval baselines}} \\
BM25 Retrieval ($k{=}1$) & 8.2 & 9.2 & 21.8 & 167 & 38.6 & 40.4 & 31.8 & 106 & 23.4 & 24.8 & 26.8 & 137 \\
BM25 Retrieval ($k{=}2$) & 8.3 & 9.5 & 28.3 & 274 & 43.6 & 44.8 & 51.0 & 183 & 26.0 & 27.2 & 39.6 & 228 \\
BM25 Retrieval ($k{=}5$) & 8.8 & 10.5 & 37.9 & 559 & 49.5 & 50.3 & 73.0 & 415 & 29.2 & 30.4 & 55.5 & 487 \\
Dense Retrieval ($k{=}1$) & 10.6 & 15.8 & 39.7 & 241 & 36.2 & 37.0 & 25.5 & 158 & 23.4 & 26.4 & 32.6 & 200 \\
Dense Retrieval ($k{=}2$) & 10.3 & 14.6 & 53.1 & 411 & 39.0 & 40.0 & 40.1 & 303 & 24.6 & 27.3 & 46.6 & 357 \\
Dense Retrieval ($k{=}5$) & 10.6 & 14.7 & 67.1 & 848 & 45.2 & 45.9 & 62.0 & 750 & 27.9 & 30.2 & 64.5 & 799 \\
\midrule
\multicolumn{13}{l}{\textit{Raw-evidence multimodal retrieval baselines}} \\
Qwen3-VL-8B ($k{=}1$) & 8.6 & 11.6 & 24.0 & 164 & 40.7 & 42.5 & 40.8 & 108 & 24.7 & 27.1 & 32.4 & 136 \\
Qwen3-VL-8B ($k{=}2$) & 9.1 & 11.1 & 32.6 & 270 & 47.6 & 48.7 & 66.3 & 187 & 28.4 & 29.9 & 49.5 & 228 \\
Qwen3-VL-8B ($k{=}5$) & 9.5 & 12.4 & 49.6 & 570 & \underline{52.3} & \underline{53.5} & \textbf{87.5} & 442 & 30.9 & 32.9 & 68.5 & 506 \\
Nemo-Emb ($k{=}1$) & 11.3 & 17.8 & 48.8 & 255 & 41.5 & 43.3 & 40.5 & 107 & 26.4 & 30.5 & 44.6 & 181 \\
Nemo-Emb ($k{=}2$) & 11.1 & 17.0 & 64.7 & 444 & 48.9 & 50.1 & 66.6 & 186 & 30.0 & 33.5 & 65.7 & 315 \\
Nemo-Emb ($k{=}5$) & 11.2 & 17.3 & \underline{81.5} & 948 & 52.1 & 53.0 & \underline{87.1} & 450 & \underline{31.7} & \underline{35.1} & \textbf{84.3} & 698 \\
\midrule
\multicolumn{13}{l}{\textit{Ours: direct generation from Latent Memory}} \\
Latent Memory ($k{=}1$) & \underline{11.7} & \textbf{18.7} & 52.5 & 38 & 31.6 & 31.9 & 23.7 & 39 & 21.6 & 25.3 & 38.1 & 38 \\
Latent Memory ($k{=}2$) & 11.3 & 18.0 & 70.1 & 47 & 31.9 & 32.3 & 41.3 & 48 & 21.6 & 25.1 & 55.7 & 48 \\
Latent Memory ($k{=}5$) & 11.3 & 17.5 & \textbf{89.0} & 74 & 31.3 & 31.7 & 70.5 & 75 & 21.3 & 24.6 & \underline{79.8} & 74 \\
\bottomrule
\end{tabular}}
\end{table}

\paragraph{Analysis.}
With Gemma-3-12B-Instruct, Latent Memory remains the most token-efficient option. Demonstrating the best WebQA-Image Accuracy and the second best F1 with 10$\times$ less tokens.

\subsection{Latent Tokens as Retrievers}
\label{app:embedding_model}

Latent Memory's latent token plays two roles at once: it is a retrieval key used to rank evidence and a generation input consumed by the frozen generator. This section isolates these two roles with a hybrid \textbf{as-RAG} mode. In this mode, latent tokens are used only for retrieval; after the top-$k$ evidence items are selected, generation is performed from the retrieved raw text or images rather than from latent tokens. We report text-only and multimodal variants separately because the raw evidence format differs across settings.

\begin{itemize}[leftmargin=*]
    \item \textbf{Ours-1B}: full Latent Memory---latent tokens used for both retrieval \emph{and} generation.
    \item \textbf{Ours-1B-RAG}: latent tokens used for retrieval only; generator sees raw retrieved text sentences.
    \item \textbf{Latent-Token-based RAG}: multimodal variant of the above---latent tokens rank the unified text-image pool; generator sees raw retrieved text and images.
\end{itemize}

Comparing these variants reveals how much of Latent Memory's gain comes from better retrieval (latent key quality) versus better generation (latent token quality as generator input).

\begin{table}[t]
\centering
\small
\caption{Text-only-as-RAG results. Latent retrieval selects raw text evidence, and generation is performed from the retrieved raw text. \#Tok excludes fixed prompt scaffolding.}
\label{tab:text_as_rag}
\resizebox{\textwidth}{!}{
\begin{tabular}{lcc|cccc|cccc}
\toprule
 & \multicolumn{2}{c}{HotpotQA} & \multicolumn{2}{c}{2WikiMultihopQA} & \multicolumn{2}{c}{MuSiQue} & \multicolumn{4}{c}{Avg} \\
\cmidrule(lr){2-3}\cmidrule(lr){4-5}\cmidrule(lr){6-7}\cmidrule(lr){8-11}
Methods & EM & F1 & EM & F1 & EM & F1 & EM & F1 & R@$k$ & \#Tok  \\
\midrule
\multicolumn{11}{l}{\textit{Full-context reference}} \\
Full Context & 42.0 & \underline{57.8} & 17.7 & \underline{39.2} & 6.0 & 17.1 & 21.9 & 38.0 & -- & 1706 \\
\midrule
\multicolumn{11}{l}{\textit{Raw-evidence RAG baseline}} \\
Qwen3-Emb ($k{=}1$) & 30.9 & 43.6 & 18.1 & 30.9 & 8.1 & 18.2 & 19.0 & 30.9 & 25.3 & 70 \\
Qwen3-Emb ($k{=}2$) & 35.6 & 49.0 & 17.4 & 33.5 & 9.8 & 20.4 & 21.0 & 34.3 & 38.8 & 108 \\
Qwen3-Emb ($k{=}5$) & 40.0 & 54.5 & 19.1 & 38.6 & \underline{13.7} & \underline{24.6} & 24.3 & \underline{39.2} & \underline{56.4} & 221 \\
\midrule
\multicolumn{11}{l}{\textit{Direct generation from Latent Memory}} \\
Latent Memory ($k{=}1$) & 27.4 & 39.4 & 19.8 & 29.2 & 5.8 & 14.6 & 17.7 & 27.7 & 23.9 & 36 \\
Latent Memory ($k{=}2$) & 31.6 & 45.2 & 21.5 & 33.5 & 7.3 & 16.9 & 20.1 & 31.9 & 42.6 & 45 \\
Latent Memory ($k{=}5$) & 34.8 & 48.9 & \textbf{24.3} & 36.7 & 8.7 & 19.2 & 22.6 & 34.9 & \textbf{63.8} & 72 \\
\midrule
\multicolumn{11}{l}{\textit{Latent retrieval as raw-evidence RAG}} \\
Latent Memory-as-RAG ($k{=}1$) & 37.0 & 49.8 & 18.6 & 30.7 & 10.6 & 19.9 & 22.1 & 33.5 & 23.9 & 75 \\
Latent Memory-as-RAG ($k{=}2$) & \underline{44.0} & \underline{58.9} & 18.4 & 35.4 & 13.2 & 22.8 & \underline{25.2} & 39.0 & 42.6 & 125 \\
Latent Memory-as-RAG ($k{=}5$) & \textbf{49.9} & \textbf{66.2} & \underline{21.9} & \textbf{42.8} & \textbf{18.0} & \textbf{30.2} & \textbf{29.9} & \textbf{46.4} & \textbf{63.8} & 259 \\
\bottomrule
\end{tabular}}
\end{table}

\begin{table}[t]
\centering
\small
\caption{Multimodal-as-RAG results. Latent retrieval ranks the unified text-image pool, but generation is performed from retrieved raw text and raw images.}
\label{tab:mm_as_rag}
\resizebox{\textwidth}{!}{
\begin{tabular}{lcccc cccc |cccc}
\toprule
Method & \multicolumn{4}{c}{\textbf{WebQA-Image}} & \multicolumn{4}{c}{\textbf{WebQA-Text}} & \multicolumn{4}{c}{\textbf{Avg}} \\
\cmidrule(lr){2-5}\cmidrule(lr){6-9}\cmidrule(lr){10-13}
 & F1 & Acc & R@$k$ & \#Tok & F1 & Acc & R@$k$ & \#Tok & F1 & Acc & R@$k$ & \#Tok \\
\midrule
\multicolumn{13}{l}{\textit{Raw-evidence multimodal retrieval baselines}} \\
Qwen3-VL-8B ($k{=}1$) & 15.1 & 22.0 & 24.1 & 284 & 40.8 & 43.5 & 40.7 & 131 & 27.9 & 32.8 & 32.4 & 208 \\
Qwen3-VL-8B ($k{=}2$) & 20.1 & 24.5 & 32.9 & 465 & 46.1 & 50.2 & 66.3 & 235 & 33.1 & 37.3 & 49.6 & 350 \\
Qwen3-VL-8B ($k{=}5$) & 34.1 & 30.3 & 49.7 & 957 & \underline{47.8} & \underline{57.2} & \textbf{87.2} & 612 & 41.0 & 43.8 & 68.5 & 784 \\
Nemo-Emb ($k{=}1$) & 19.9 & 25.2 & 48.7 & 508 & 41.1 & 44.0 & 40.4 & 130 & 30.5 & 34.6 & 44.6 & 319 \\
Nemo-Emb ($k{=}2$) & 32.9 & 30.6 & 64.8 & 892 & 46.8 & 50.9 & 66.6 & 233 & 39.9 & 40.8 & 65.7 & 563 \\
Nemo-Emb ($k{=}5$) & 53.0 & 37.2 & \underline{81.4} & 1885 & \textbf{48.6} & \textbf{57.9} & \underline{87.1} & 629 & \textbf{50.8} & \textbf{47.6} & \textbf{84.2} & 1257 \\
\midrule
\multicolumn{13}{l}{\textit{Direct generation from Latent Memory}} \\
Latent Memory ($k{=}1$) & 32.0 & 28.7 & 56.6 & 42 & 28.8 & 30.8 & 24.0 & 44 & 30.4 & 29.7 & 40.3 & 43 \\
Latent Memory ($k{=}2$) & \underline{56.5} & \underline{39.5} & 74.7 & 52 & 30.0 & 32.8 & 41.1 & 54 & 43.2 & 36.2 & 57.9 & 53 \\
Latent Memory ($k{=}5$) & \textbf{69.4} & \textbf{44.2} & \textbf{91.2} & 82 & 30.7 & 34.3 & 70.5 & 84 & \underline{50.0} & 39.2 & \underline{80.8} & 83 \\
\midrule
\multicolumn{13}{l}{\textit{Latent retrieval as raw-evidence multimodal RAG}} \\
Latent Memory-as-RAG ($k{=}1$) & 21.4 & 25.8 & 56.6 & 638 & 34.9 & 37.6 & 24.0 & 122 & 28.2 & 31.7 & 40.3 & 380 \\
Latent Memory-as-RAG ($k{=}2$) & 38.3 & 33.3 & 74.7 & 1233 & 40.3 & 43.9 & 41.1 & 207 & 39.3 & 38.6 & 57.9 & 720 \\
Latent Memory-as-RAG ($k{=}5$) & 54.7 & 37.0 & \textbf{91.2} & 2970 & 45.1 & 51.2 & 70.5 & 461 & 49.9 & \underline{44.1} & \underline{80.8} & 1715 \\
\bottomrule
\end{tabular}}
\end{table}

\paragraph{Analysis.}
\begin{enumerate}[leftmargin=*]
\item In the plain text setting, the "Latent Memory as RAG" method based on raw evidence effectively isolates retrieval and compression. We observed that with Latent Memory k=1, it outperforms Latent-Memory-as-RAG (k=5) using the same number of tokens. This demonstrates that even disregarding the advantage of retrieval, the compression portion of Latent Memory still exhibits a trade-off, and the 8-token Latent Memory results show a better trade-off.

\item In multimodal WebQA, the conclusion is different. The "Latent Memory as RAG" method based on raw evidence outperforms in the text portion, but it lags behind using latent tokens in the image portion. This indicates that the performance improvement on the image side is related to obtaining a more efficient representation, which helps alleviate the context processing pressure on large models.
\item Furthermore, the improvement in recall@k itself may provide inspiration for the training of subsequent embedding models. This is an embedding design based on providing raw context, demonstrating the effectiveness of unified retrieval and representation space.
\end{enumerate}

\newpage

\section{Case Study}
\label{app:examples}

The case studies are not additional leaderboard evidence; they diagnose how the compact representation behaves on individual examples. We use them to answer three qualitative questions:

\begin{itemize}[leftmargin=*]
    \item \textbf{RQ-C1: What does reconstruction reveal about the latent token?} Appendix~\ref{app:case_recon} checks whether reconstruction preserves key evidence rather than copying the full sentence verbatim.
    \item \textbf{RQ-C2: Can text-only latent retrieval recover complete multi-hop chains?} Appendix~\ref{app:examples_text} shows examples where the retrieved latent memories cover all required supporting facts.
    \item \textbf{RQ-C3: How does unified text-image retrieval behave on concrete multimodal questions?} Appendix~\ref{app:examples_mm} shows additional WebQA cases using the same unified text-image memory pool as the main experiments.
\end{itemize}

\subsection{Reconstruction Quality of Latent Tokens}
\label{app:case_recon}

\begin{figure}[H]
\centering
\includegraphics[width=0.80\linewidth]{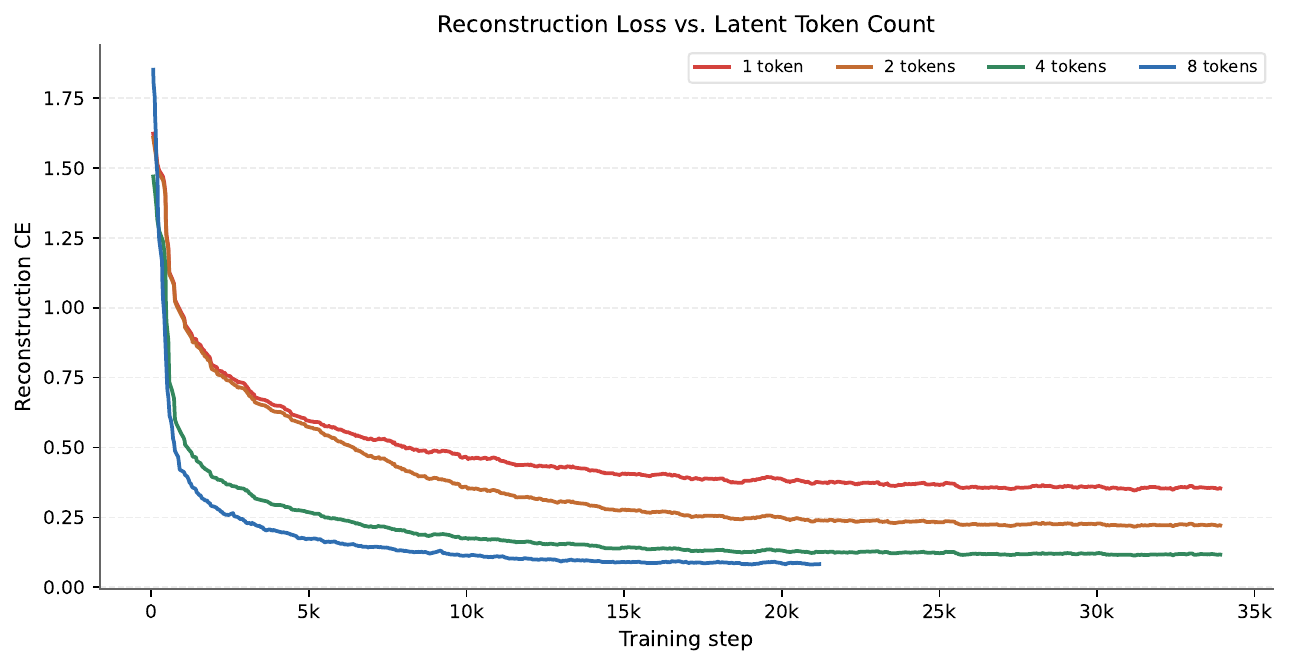}
\caption{Training reconstruction loss for 1-, 2-, 4-, and 8-token Latent Memory variants. More latent tokens consistently reduce the reconstruction CE, indicating that additional latent capacity preserves more evidence information during compression.}
\label{fig:reconstruction_loss_token_curve}
\end{figure}

\begin{table}[H]
\centering
\small
\caption{Qualitative reconstruction examples on HotpotQA. CE is measured against the original evidence sentence under teacher forcing.}
\label{tab:reconstruction_quality_examples}
\setlength{\tabcolsep}{4pt}
\renewcommand{\arraystretch}{1.08}
\resizebox{\linewidth}{!}{
\begin{tabular}{p{0.27\linewidth}p{0.31\linewidth}c p{0.31\linewidth}c}
\toprule
\textbf{Original evidence} & \textbf{1-token reconstruction} & \textbf{CE} & \textbf{8-token reconstruction} & \textbf{CE} \\
\midrule
\textbf{Shirley Temple:} Shirley Temple Black (April 23, 1928--February 10, 2014) was an American actress, singer, dancer, businesswoman, and diplomat who was Hollywood's number one box-office draw as a child actress from 1935 to 1938.
&
actress greater greater greater greater greater \ldots
&
0.226
&
Shirley Temple (April 23, 1928--February 10, 2014) was an actress, singer, dancewoman, and business actress who was Hollywood's number one child \ldots
&
0.062 \\
\midrule
\textbf{2014 S/S:} 2014 S/S is the debut album of South Korean group WINNER.
&
S:\#:\#:\#:\#:\#:\#:\#:\#:\#:\# \ldots
&
0.486
&
2014 is the debut album of South Korean group WINNER \ldots
&
0.016 \\
\bottomrule
\end{tabular}}
\end{table}

The table above provides examples for latent reconstruction. 1-tokens are okay for low CE but cannot fully reconstruct the evidence; improving the number of tokens per latent evidence is helpful for reconstruction as the reconstruction loss in Figure \ref{fig:reconstruction_loss_token_curve} will drop accordingly.

The ability to reconstruct the text not only provides a stronger and more faithful representation of the original text but also enhances the interpretability of latent memory.

\newpage
\subsection{Text-only Case Studies}
\label{app:examples_text}
\noindent Table~\ref{tab:text_case_study_selected} presents two additional text-only cases. Both require multi-hop composition rather than simple lexical matching: the first links a screenwriter to a Nicolas Cage film through two supporting facts, while the second links a person to a company and then to the company's headquarters. In both cases, the required evidence chain is ranked at the top, and the final answer is recovered from a compact retrieved set.

\begin{table}[H]
\centering
\small
\caption{Two text-only case studies. Both examples achieve exact-match generation, and the retrieved evidence chain is fully covered by the top-ranked latent retrieval results.}
\label{tab:text_case_study_selected}
\setlength{\tabcolsep}{4pt}
\renewcommand{\arraystretch}{1.08}
\begin{tabular}{p{0.16\textwidth}p{0.74\textwidth}}
\toprule
\textbf{Case 1} & \textbf{Question:} What screenwriter with credits for ``Evolution'' co-wrote a film starring Nicolas Cage and T\'ea Leoni? \\
\midrule
Generation &
\textbf{Gold answer:} \textit{David Weissman}. \\
&
\textbf{Predicted answer:} \textit{David Weissman}. \\
&
\textbf{EM:} 1.0 \\
&
\textbf{Retrieved positives:} 3/3. \\
Latent Memory &
\begin{tabular}[t]{p{0.035\textwidth}p{0.045\textwidth}p{0.075\textwidth}p{0.43\textwidth}p{0.06\textwidth}}
\toprule
Rank & Pos? (gold) & Retrieval score & Corresponding text of retrieved latent token & CE loss \\
\midrule
1 & Y & 0.507 & David Weissman: His film credits include ``The Family Man'' (2000), ``Evolution'' (2001), and ``When in Rome'' (2010). & 0.260 \\
2 & Y & 0.447 & The Family Man: The Family Man is a 2000 American romantic comedy-drama film ... starring Nicolas Cage and T\'ea Leoni. & 0.089 \\
3 & Y & 0.394 & David Weissman: David Weissman is a screenwriter and director. & 0.010 \\
\bottomrule
\end{tabular} \\
\midrule
\textbf{Case 2} & \textbf{Question:} Where is the company that Sachin Warrier worked for as a software engineer headquartered? \\
\midrule
Generation &
\textbf{Gold answer:} \textit{Mumbai}. \\
&
\textbf{Predicted answer:} \textit{Mumbai}. \\
&
\textbf{EM:} 1.0 \\
&
\textbf{Retrieved positives:} 2/2. \\
Latent Memory &
\begin{tabular}[t]{p{0.035\textwidth}p{0.045\textwidth}p{0.075\textwidth}p{0.43\textwidth}p{0.06\textwidth}}
\toprule
Rank & Pos? (gold) & Retrieval score & Corresponding text of retrieved latent token & CE loss \\
\midrule
1 & Y & 0.623 & Tata Consultancy Services: Tata Consultancy Services Limited (TCS) is an Indian multinational information technology service company headquartered in Mumbai, Maharashtra. & 0.342 \\
2 & Y & 0.557 & Sachin Warrier: He was working as a software engineer in Tata Consultancy Services in Kochi. & 0.083 \\
\bottomrule
\end{tabular} \\
\bottomrule
\end{tabular}
\end{table}

\newpage
\subsection{More Multimodal QA Case Studies}
\label{app:examples_mm}
\begin{figure}[H]
\centering\vspace{-3pt}
\includegraphics[width=\textwidth]{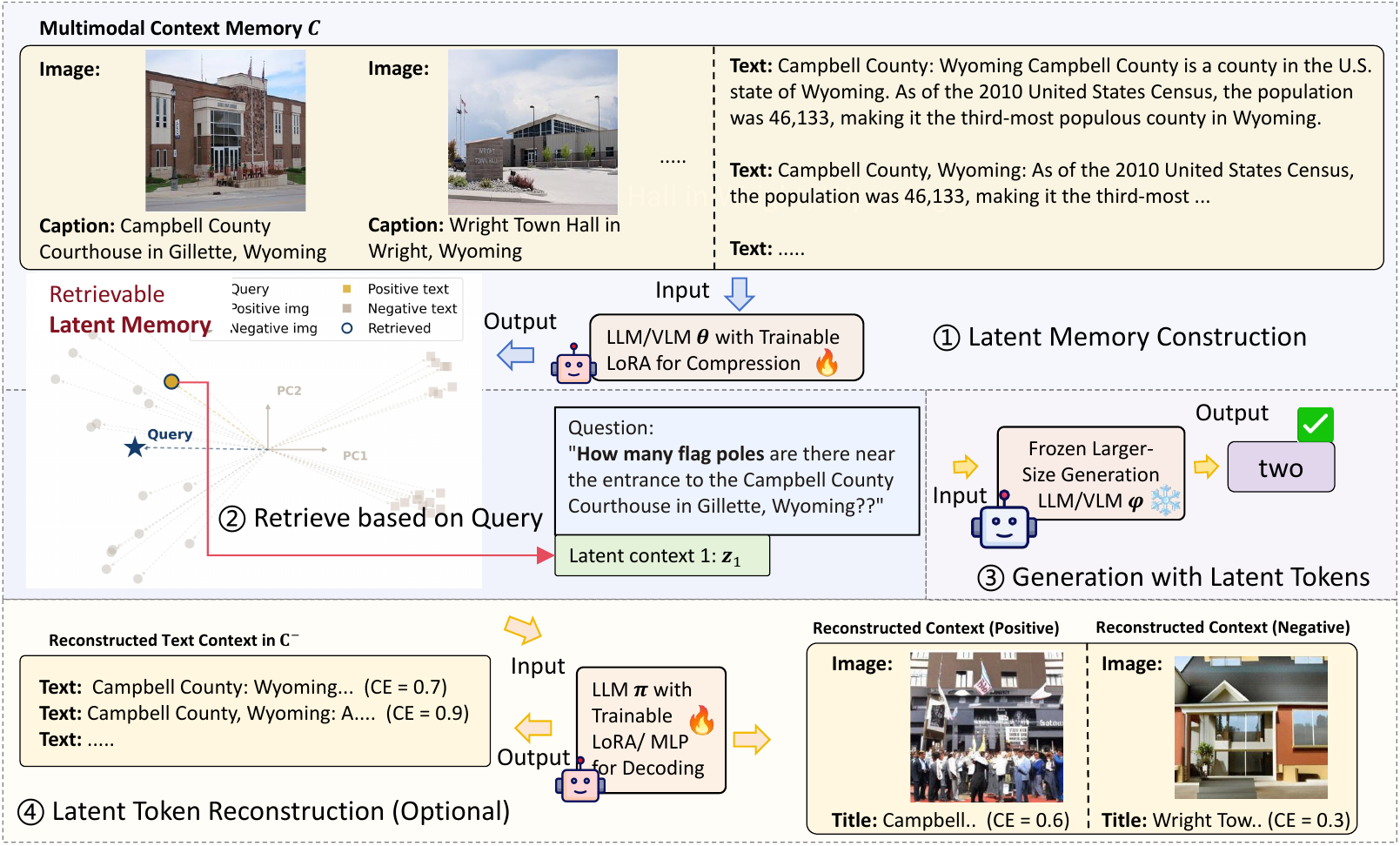}\vspace{5pt}
\includegraphics[width=\textwidth]{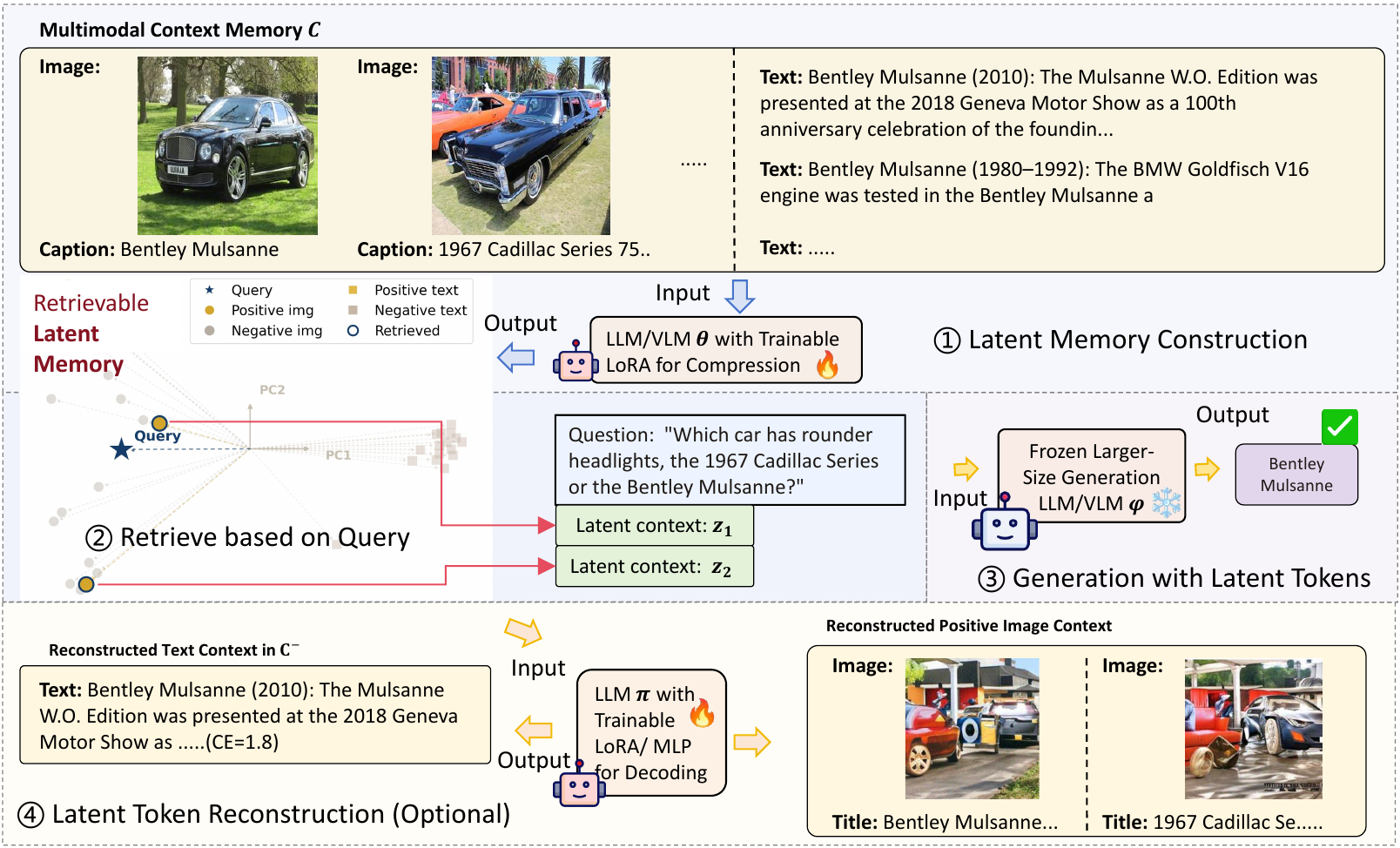}
\caption{Additional multimodal WebQA case studies with LLaVA-1.5-13B. Each example uses the same Latent Memory retrieval-and-generation pipeline as the main experiments.}
\label{fig:case_study_more}\vspace{-2pt}
\end{figure}

Figure~\ref{fig:case_study_more} provides additional WebQA examples beyond the main-text case study, showing the counting and comparison ability in multimodal reasoning. These examples illustrate how Latent Memory retrieves from a unified text-image candidate pool and then conditions the frozen LLaVA-1.5-13B generator on the retrieved latent tokens. We include both successful and challenging multimodal cases to show the qualitative behavior behind the aggregate results: image-grounded examples highlight whether visual evidence can be preserved after compression, while text-grounded examples show how textual facts are selected and used under the same retrieval interface.

\newpage

\section{Baselines, Datasets, and Licenses}
\label{app:assets}

\paragraph{Datasets.}
Table~\ref{tab:dataset_coverage} summarizes all datasets used in the paper. For the main text-only setting, Latent Memory is trained on the HotpotQA training split and evaluated on HotpotQA validation, 2WikiMultihopQA, and MuSiQue without additional task-specific fine-tuning for the out-of-domain datasets. The generalization experiments further evaluate the same text checkpoint on NQ, TriviaQA, Qasper, and WICE to cover open-domain factoid QA, scientific-document QA, and claim verification.

For multimodal QA, WebQA is used for both training and evaluation under its public split; evaluation is reported separately for image-grounded and text-grounded questions, while retrieval is performed over the unified text-image candidate pool. SlideQA is a pure visual dataset covering slides and detail capture. We hope to find more multimodal datasets similar to WebQA that involve retrieval and reasoning, but unfortunately, this is almost the only one. The rest of the datasets often contain data outside the Latent Memory scope, such as tables and videos.

As shown in Table \ref{tab:text_case_study_selected} and case study Figures, all evidence is processed in ``Title: Sentence`` form for the Text-only setting, ``Title: Evidence`` and ``Caption: Image`` for the multimodal setting.

\begin{table}[h]
\centering
\small
\caption{Coverage of the datasets used in this work.}
\label{tab:dataset_coverage}
\setlength{\tabcolsep}{4pt}
\resizebox{\textwidth}{!}{
\begin{tabular}{p{2.2cm}p{4.2cm}p{3.0cm}p{5.4cm}}
\toprule
\textbf{Dataset} & \textbf{Knowledge / Task} & \textbf{Domain} & \textbf{Primary Source} \\
\midrule
HotpotQA \citep{yang2018hotpotqa} & Multi-hop entity-centric QA with supporting-fact grounding & Encyclopedic factual knowledge & Wikipedia articles and supporting sentences \\
2WikiMultihopQA \citep{ho20202wikimultihop} & Cross-page factual reasoning and relation composition & Encyclopedic factual knowledge & Wikipedia-based evidence chains \\
MuSiQue \citep{trivedi2022musique} & More compositional multi-hop QA over multiple supporting paragraphs & Broad factual knowledge & Multi-paragraph textual QA collections \\
WebQA \citep{chang2022webqa} & Unified text-image QA with visually grounded evidence & Web knowledge, mixed textual and visual evidence & Web text and web images \\
SlideQA & Multimodal slide-domain QA with text and visual evidence & Slide / presentation understanding & Slide pages, associated visual regions, and textual metadata \\
\midrule
NQ \citep{kwiatkowski2019natural} & Open-domain factoid QA, often short-answer retrieval & Open-domain factual knowledge & Search queries paired with Wikipedia evidence \\
TriviaQA \citep{joshi2017triviaqa} & Trivia-style factoid QA with strong entity coverage & Open-domain factual knowledge & Trivia questions with web / Wikipedia evidence \\
Qasper \citep{dasigi2021qasper} & Document-grounded QA over scientific papers & Scientific document understanding & Research papers with caption, section, and paragraph structure \\
WICE \citep{kamoi2023wice} & Evidence-centered claim verification and explanation & Fact verification / evidence reasoning & Claim-evidence collections built from textual evidence sources \\
\bottomrule
\end{tabular}}
\end{table}

\paragraph{Baseline settings.}
For text-only QA, the main table uses the same fixed \textbf{Meta-Llama-3-8B-Instruct} generation model and the same question prompt format. \textbf{Full Context} concatenates all available context sentences in the sample and sends them directly to the generator. \textbf{LLMLingua} \citep{jiang2023llmlingua} compresses the raw text context first, and the compressed text is then inserted into the same generator prompt. \textbf{BM25 Retrieval} \citep{robertson2009probabilistic} ranks candidate sentences with sparse lexical matching and feeds the top-$k$ raw sentences to the generator. \textbf{Dense Retrieval} \citep{karpukhin2020dense} ranks the same candidate sentence pool with \texttt{all-MiniLM-L6-v2} sentence embeddings and also feeds top-$k$ raw sentences. \textbf{Qwen3-Embedding} \citep{zhang2025qwen3} uses the Qwen3 text embedding model as a stronger dense retriever, with generation still performed from retrieved raw text. Since a series of embedding models, such as the Qwen-3-Embedding model, are usually pre-trained for our in-domain scenario (Wikipedia), fine-tuning on our chosen in-domain environment could disrupt the balance; so we believe that direct comparison with Latent Memory for retrieval is relatively fair.

Because xRAG \citep{cheng2024xrag} and CLaRa \citep{he2025clara} are built around Mistral-style latent-context generation, we only compare to their pretrained models on the \textbf{Mistral-7B-Instruct} setting in Table~\ref{tab:mistral_text_results}. For xRAG, we use the pretrained \texttt{Hannibal046/xrag-7b} generator-side checkpoint together with the pretrained \texttt{Salesforce/SFR-Embedding-Mistral} retrieval encoder. For CLaRa, we use the released pretrained CLaRa checkpoints at the reported 16$\times$ compression settings (which are more suitable for our sentence-level evidence). Our Mistral-setting Latent Memory uses \textbf{LLaMA-3.2-1B-Instruct} as both the compressor/encoder and the reconstruction decoder, while the answer generator is frozen Mistral-7B-Instruct.

For multimodal QA, all baselines retrieve from the same unified WebQA candidate pool \citep{chang2022webqa} containing both text facts and images. We report two frozen generator families: \textbf{LLaVA-1.5-13B} in the main WebQA setting and \textbf{Gemma-3-12B-Instruct} in Appendix~\ref{app:gemma_mm_results}. In the implementation, the LLaVA setting is run through \texttt{scripts/baselines\_llava.py}, while the Gemma setting is run through \texttt{scripts/baselines\_qwen.py}, whose generator wrapper also supports \texttt{model\_type=gemma3}. For Gemma, the same system instruction is prepended inside the user message because the Gemma chat template used by the processor does not take a separate system role in our code. \textbf{Full Context} passes all candidate text and images to the generator, which is often expensive because images expand through the model's visual frontend. \textbf{BM25 Retrieval} \citep{robertson2009probabilistic} performs sparse retrieval over textual fields; if an image candidate is retrieved through its title/caption metadata, the raw image is still provided to the VLM for generation. \textbf{Dense Retrieval} \citep{karpukhin2020dense} uses textual surrogates for retrieval: text facts are encoded directly, while images are represented by their titles/captions for scoring; after retrieval, the original raw image is sent to the generator. \textbf{Qwen3-VL-Embedding} \citep{li2026qwen3}, and \textbf{Nemo Retriever} \citep{xu2025llama} are multimodal retrieval baselines that rank text-image candidates with pretrained vision-language retrieval models, but the generator still consumes the retrieved raw text or image rather than the retrieval embedding. This keeps the retrieval candidate pool unified while making the raw-evidence baselines comparable to Latent Memory.

For all retrieval baselines, $k$ denotes the number of retrieved evidence items and is matched to the corresponding Latent Memory setting when reported. Token counts always measure the generator-side input budget after constructing the final prompt. Thus, text retrieval baselines count the retrieved text tokens, multimodal baselines count the effective tokens consumed after visual processing, and Latent Memory counts the projected Latent Memory tokens inserted into the frozen generator.

\paragraph{Licenses.}
Table~\ref{tab:assets} summarises the main datasets, pretrained models, and baselines used in this work. We report the license or usage terms stated on the official upstream release pages that we could verify directly. When an official public release does not state a license, we mark it as not stated rather than inferring one. For benchmarks that redistribute web pages or images, downstream use must additionally comply with the rights of the original content providers.

\begin{table}[h]
\centering
\small
\caption{Summary of datasets, baselines, and upstream license provenance.}
\label{tab:assets}
\setlength{\tabcolsep}{4pt}
\resizebox{\textwidth}{!}{
\begin{tabular}{p{3.0cm}p{1.6cm}p{4.2cm}p{5.8cm}}
\toprule
\textbf{Item} & \textbf{Type} & \textbf{License / Terms} & \textbf{Source} \\
\midrule
HotpotQA \citep{yang2018hotpotqa} & Dataset & CC BY-SA 4.0 & Hugging Face: \texttt{hotpotqa/hotpot\_qa} \\
2WikiMultihopQA \citep{ho20202wikimultihop} & Dataset & Apache License 2.0 & Hugging Face: \texttt{xanhho/2WikiMultihopQA} \\
MuSiQue \citep{trivedi2022musique} & Dataset & Available Online & Hugging Face dataset cards: \texttt{bdsaglam/musique} \\
WebQA \citep{chang2022webqa} & Dataset & Available Online & WebQA official projects \\
\midrule
NQ \citep{kwiatkowski2019natural,jiang2024longrag} & Dataset & GitHub: \texttt{TIGER-Lab/LongRAG} \\
TriviaQA \citep{joshi2017triviaqa} & Dataset & Apache License 2.0 in the Hugging Face release used by our data pipeline & Hugging Face: \texttt{mandarjoshi/trivia\_qa} \\
Qasper \citep{dasigi2021qasper} & Dataset & CC BY 4.0 & Hugging Face: \texttt{allenai/qasper} \\
WICE \citep{kamoi2023wice} & Dataset & Available Online & WiCE project page by the authors and mirrored dataset pages \\
SlideQA & Dataset & Available Online & Hugging Face: \texttt{NTT-hil-insight/SlideVQA} \\
\midrule
BM25 Retrieval \citep{robertson2009probabilistic} & Baseline & Available Online & Standard sparse retrieval baseline \\
Dense Retrieval \citep{karpukhin2020dense} & Baseline & Apache License 2.0 & Hugging Face: \texttt{sentence-transformers/all-MiniLM-L6-v2} \\
Qwen3-Embedding \citep{zhang2025qwen3} & Baseline & Apache License 2.0 & Hugging Face: \texttt{Qwen/Qwen3-Embedding-0.6B} \\
Qwen3-VL-Embedding \citep{li2026qwen3} & Baseline & Apache License 2.0 & Hugging Face: \texttt{Qwen/Qwen3-VL-Embedding-8B} \\
Nemo Retriever \citep{xu2025llama} & Baseline & Apache License 2.0 & Hugging Face: \texttt{nvidia/llama-nemotron-embed-vl-1b-v2} \\
LLMLingua \citep{jiang2023llmlingua} & Baseline & Apache License 2.0 & Microsoft LLMLingua repository \\
xRAG \citep{cheng2024xrag} & Baseline & Available Online & Pretrained checkpoint: \texttt{Hannibal046/xrag-7b}; retrieval encoder: \texttt{Salesforce/SFR-Embedding-Mistral} \\
CLaRa \citep{he2025clara} & Baseline & Not stated & Released pretrained CLaRa checkpoints for 16$\times$ compression settings \\
\midrule
Llama-3.2-1B-Instruct \citep{touvron2023llama} & Model & Llama 3.2 Community License & Hugging Face / Meta model card: \texttt{meta-llama/Llama-3.2-1B-Instruct} \\
Meta-Llama-3-8B-Instruct \citep{touvron2023llama} & Model & Meta Llama 3 Community License & Hugging Face / Meta model card: \texttt{meta-llama/Meta-Llama-3-8B-Instruct} \\
Mistral-7B-Instruct \citep{jiang2023mistral} & Model & Apache License 2.0 & Hugging Face / Mistral AI model release: \texttt{mistralai/Mistral-7B-Instruct-v0.3} \\
LLaVA-1.5-7B \citep{liu2023visual} & Model & Llama 2 Community License & Hugging Face: \texttt{liuhaotian/llava-v1.5-7b}; LLaVA website \\
LLaVA-1.5-13B \citep{liu2023visual} & Model & Llama 2 Community License & Hugging Face: \texttt{liuhaotian/llava-v1.5-13b}; LLaVA website \\
Gemma-3-4B-PT \citep{Kamath2025Gemma3T} & Model & Gemma Terms of Use / Gemma license & Hugging Face: \texttt{google/gemma-3-4b-pt}; Google AI for Developers Gemma Terms \\
Gemma-3-12B-IT \citep{Kamath2025Gemma3T} & Model & Gemma Terms of Use / Gemma license & Hugging Face: \texttt{google/gemma-3-12b-it}; Google AI for Developers Gemma Terms \\
CLIP ViT-L/14-336 \citep{radford2021learning} & Model & MIT for OpenAI CLIP code/release & GitHub: \texttt{openai/CLIP}; Hugging Face mirror: \texttt{openai/clip-vit-large-patch14-336} \\
\bottomrule
\end{tabular}
}
\end{table}
\newpage

\end{document}